\PassOptionsToPackage{prologue,dvipsnames}{xcolor}

\documentclass[sigconf]{acmart}

\AtBeginDocument{%
  \providecommand\BibTeX{{%
    Bib\TeX}}}

\setcopyright{acmlicensed}
\copyrightyear{2024}
\acmYear{2024}
\acmDOI{10.1145/3658644.3690322}

\acmConference[CCS '24]{Proceedings of the 2024 ACM SIGSAC Conference on Computer and Communications Security}{October 14--18, 2024}{Salt Lake City, UT, USA}

\acmISBN{979-8-4007-0636-3/24/10}

\usepackage{algorithmic}
\usepackage{graphicx}
\def\BibTeX{{\rm B\kern-.05em{\sc i\kern-.025em b}\kern-.08em
    T\kern-.1667em\lower.7ex\hbox{E}\kern-.125emX}}

\usepackage{xspace}
\usepackage{lipsum}
\usepackage{subfigure} 
\usepackage{diagbox}
\usepackage{color}
\usepackage[linesnumbered,ruled]{algorithm2e}
\usepackage{booktabs}
\usepackage{multirow}
\usepackage{threeparttable}
\usepackage{array}
\usepackage{bbding}
\usepackage{threeparttable}
\usepackage{color, xcolor}
\usepackage{epigraph} 
\usepackage{balance}
\usepackage{microtype}

\usepackage{bbm}
\usepackage{tcolorbox}
\usepackage{colortbl}

\newcommand{\blue}[1]{\textcolor{blue}{#1}}

\newcommand{\sys}{{\sc Legilimens}\xspace}

\definecolor{xxxx}{RGB}{54, 179, 122}
\newcommand{\tblc}[1]{\cellcolor{xxxx!#1!Lavender}{#1}}

\usepackage[dvipsnames]{xcolor}

\begin{document}

\title[{\sys}: Practical and Unified Content Moderation for Large Language Model Services]{{\sys}: Practical and Unified Content Moderation for\texorpdfstring{\\}{}Large Language Model Services}


\author{Jialin Wu}
\authornote{Both authors contributed equally to the paper.}
\affiliation{%
  \institution{Zhejiang University}
  \city{Hangzhou}
  \state{Zhejiang}
  \country{China}}
\email{jialinwu@zju.edu.cn}

\author{Jiangyi Deng}
\authornotemark[1]
\affiliation{%
  \institution{Zhejiang University}
  \city{Hangzhou}
  \state{Zhejiang}
  \country{China}}
\email{jydeng@zju.edu.cn}

\author{Shengyuan Pang}
\affiliation{%
  \institution{Zhejiang University}
  \city{Hangzhou}
  \state{Zhejiang}
  \country{China}}
\email{pangpang0093@zju.edu.cn}

\author{Yanjiao Chen}
\affiliation{%
  \institution{Zhejiang University}
  \city{Hangzhou}
  \state{Zhejiang}
  \country{China}}
\email{chenyanjiao@zju.edu.cn}

\author{Jiayang Xu}
\affiliation{%
  \institution{Zhejiang University}
  \city{Hangzhou}
  \state{Zhejiang}
  \country{China}}
\email{3210103789@zju.edu.cn}

\author{Xinfeng Li}
\affiliation{%
  \institution{Zhejiang University}
  \city{Hangzhou}
  \state{Zhejiang}
  \country{China}}
\email{xinfengli@zju.edu.cn}

\author{Wenyuan Xu}
\affiliation{%
  \institution{Zhejiang University}
  \city{Hangzhou}
  \state{Zhejiang}
  \country{China}}
\email{wyxu@zju.edu.cn}

\renewcommand{\shortauthors}{Jialin Wu \textit{et al.}}

\begin{abstract}
Given the societal impact of unsafe content generated by large language models (LLMs), ensuring that LLM services comply with safety standards is a crucial concern for LLM service providers. Common content moderation methods are limited by an effectiveness-and-efficiency dilemma, where simple models are fragile while sophisticated models consume excessive computational resources. In this paper, we reveal for the first time that effective and efficient content moderation can be achieved by extracting conceptual features from chat-oriented LLMs, despite their initial fine-tuning for conversation rather than content moderation. We propose a practical and unified content moderation framework for LLM services, named \sys, which features both effectiveness and efficiency. Our red-team model-based data augmentation enhances the robustness of \sys against state-of-the-art jailbreaking. Additionally, we develop a framework to theoretically analyze the cost-effectiveness of \sys compared to other methods. 

We have conducted extensive experiments on five host LLMs, seventeen datasets, and nine jailbreaking methods to verify the effectiveness, efficiency, and robustness of \sys against normal and adaptive adversaries. A comparison of \sys with both commercial and academic baselines demonstrates the superior performance of \sys. Furthermore, we confirm that \sys can be applied to few-shot scenarios and extended to multi-label classification tasks.

\end{abstract}




\maketitle
\begin{figure}[h]
    \centering
\setlength{\abovecaptionskip}{10pt}
\setlength{\belowcaptionskip}{-5pt}


\includegraphics[width=3.2in, trim=370 120 370 140, clip]{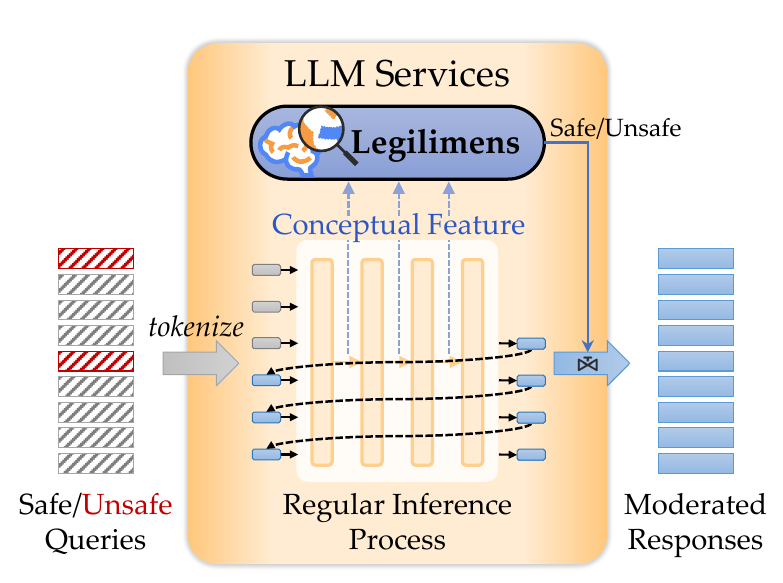}

\caption{\sys is a unified content moderation framework for almost all LLM services, which features both effectiveness and efficiency. By monitoring conceptual features generated in the regular inference process of LLMs, \sys achieves lightweight moderation.}\label{fig:goal}
\end{figure}
\section{Introduction}\label{sec:intro}

With the rise of Large Language Models (LLMs), there has been a significant increase in global user engagement with online LLM services. However, the impressive and unknown capacities of chat-oriented LLMs can be used to malicious ends. Although LLMs are currently deployed on a large scale without properly taking into account the harm they can cause to society, there is a growing concern of and calls for compliance of LLMs to safety standard. Since model alignment by removing undesired behaviors from models has been empirically~\cite{WallaceFKGS19,YuS21,XuJLBWD21,LiGFXHMS23,Liu2023Jailbreaking,Zhu2023AutoDAN,Yuan2023GPT} and theoretically~\cite{Wolf2023Fundamental} proved to have fundamental limitations, service providers are tirelessly pursuing an effective and efficient content moderation solution.

However, existing content moderation methods have been found to be dangerously brittle. Even commercial solutions like OpenAI's can be easily bypassed~\cite{LiGFXHMS23,Liu2023Jailbreaking,Zhu2023AutoDAN,Yuan2023GPT} due to the inadequate ability of simple classifiers~\cite{Wei2023Jailbroken}. Some researchers explored using LLMs as more sophisticated classifiers~\cite{WangHACL23,Huang2023Harnessing,Ma2023Adapting,Cao2023Toxicity}. But LLM-based content moderation imposes too great a burden on already insufficient computational resources~\cite{Griffin2023ChatGPT}, rendering it impractical in real-world applications. This dilemma of effectiveness and efficiency requires urgent resolution.

\begin{table*}[tt]\centering
\setlength{\abovecaptionskip}{0pt}%
\setlength{\belowcaptionskip}{0pt}%
\caption{\sys versus existing works.}\label{tab:existing}
\resizebox{\linewidth}{!}{
\begin{threeparttable}
\setlength{\tabcolsep}{4mm}{
\begin{tabular}{@{}rrrcl@{}}
\toprule
\textbf{Method}                              & \textbf{Model Architecture}    & \textbf{Model Parameters} & \textbf{Extra Overhead\tnote{$*$}} & \textbf{Domain}     \\ \midrule
OpenAI Moderation API~\cite{MarkovZANLAJW23} & GPT\tnote{\textdagger}                            & 117M$\sim$1.5B\tnote{\textdagger}                 & $O(n)$                  & General content     \\
Google Perspective API~\cite{PerspectiveAPI} & BERT$\longrightarrow$CNN\tnote{\textdaggerdbl}           & -\tnote{\textdaggerdbl}                         & $O(n)$                  & General content     \\
BeaverDam-7B~\cite{Ji2023BeaverTails}        & LLaMA-7B                          & 7B                        & $O(n)$                  & QA\tnote{\P}                   \\

LLaMA Guard2~\cite{DBLP:journals/corr/abs-2312-06674}        & LLaMA 3-8B                          & 8B                        & $O(n)$                  & QA\tnote{\P}                   \\

GradSafe~\cite{xie2024gradsafe}        & -\tnote{\S}                          & -\tnote{\S}                       & $O(n)$                  & General content                   \\

Wang \textit{et al}.~\cite{WangHACL23}                & GPT-3                          & 175B                      & $O(n)$                  & Hate speech               \\
Cao \textit{et al}.~\cite{Cao2023Toxicity}                & GPT-4, LLaMA2                          & 100T, 13B                      & $O(n)$                  & General content               \\
FakeGPT~\cite{Huang2023Harnessing}           & GPT-3.5                        & $\geq$175B                & $O(n)$                  & Fake news           \\
Ma \textit{et al}.~\cite{Ma2023Adapting}              & ChatGLM2-6B, Baichuan-13B & 6B, 13B                   & $O(n)$                  & General content     \\ \midrule
\textbf{Legilimens (Ours)}                            & \textbf{1$\sim$3-Layer MLP}                            & \textbf{8k$\sim$4.7M}                       & $\mathbf{O(1)}$                  & \textbf{General content, QA\tnote{\P} } \\ \bottomrule
\end{tabular}
}

\begin{tablenotes}[flushleft]
\item[] (i) \textdagger: The exact GPT architecture has not been disclosed~\cite{MarkovZANLAJW23}. Based on the publication year~\cite{Yenduri2023Generative} and the API response time, we speculate that it is either GPT-1 or GPT-2. (ii) \textdaggerdbl: The CNNs are distilled from trained BERT-based models. The exact architecture has not been disclosed. (iii) $*$: $n$ denotes the (token) length of input. The complexity is estimated based on the model architecture and our experimental results. (iv) \P: ``QA'' refers to moderating question-answering (QA) data to assess the safety of the answers. (v) \S: GradSafe calculates the difference in gradients between safe and unsafe prompts for detection, thus there is no detection model.

\end{tablenotes}

\end{threeparttable}}
\end{table*}

In this paper, we make the first attempt to balance effectiveness and efficiency in content moderation for chat-oriented LLMs. As shown in Figure~\ref{fig:goal}, we propose a practical and unified content moderation framework for LLM services, named \sys\footnote{A legilimens is a person skilled in magically navigating through the many layers of a person's mind and correctly interpreting one's findings in the fantasy novels \textit{Harry Potter} by J.K. Rowling~\cite{rowling2001harry}}, which extracts distinctive conceptual features from LLMs in a lightweight manner. 
\sys is built on the idea of leveraging the regular inference process of an LLM (the \textit{host} LLM) to extract effective features for content moderation while incurring minimal overhead. A comparison of \sys with existing works is shown in Table~\ref{tab:existing}. However, despite the simplicity and straightforwardness of the basic idea, realizing this concept into a practical system poses several challenges.

\begin{itemize}
    \item \textit{How to design an efficient and unified framework for input and output moderation that is universally applicable to almost all kinds of LLMs?}
\end{itemize}

Our basic idea is to use the host LLM as a feature extractor. However, it is challenging to design a unified framework for LLMs of different architectures, for textual input or output of different lengths, and for different moderation tasks. First, LLMs can be categorized into \textit{Encoder-decoder} and \textit{decoder-only} by their architectures. And \textit{decoder-only} LLMs can be further divided into \textit{causal} decoders and \textit{prefix} decoders. In addition, LLMs commonly have different model parameters, \textit{e.g.}, LLaMA2 has three sizes, 7B, 13B and 70B~\cite{Touvron23Llama2}. Thus, the proposed method should be independent of model architectures and model sizes. Second, LLMs are sequence-to-sequence models that input and output variable-length texts. In this process, variable-length features in LLMs are generated. The proposed method should be able to deal with variable-length input/output and at best reduce the complexity to be independent of the input/output length, \textit{i.e.}, $O(1)$. Third, content moderation for LLM services consists of different tasks, including moderating the input based on the input, moderating the output based on the input and output, and moderating the input based on the input and output. The proposed method should be applicable to these tasks.

To address this problem, we design a decoder-based concept probing method that leverages the features generated in the decoding process of the first and the last output tokens. This decoding process is common to different LLM architectures and different input/output. By the attention mechanism, the features generated during the decoding of the first output token correspond to concepts related to the entire input. Similarly, the features generated during the decoding of the last output token correspond to concepts related to both the input and output. We utilize these two types of features to achieve both input and output moderation, namely $I$-moderation and $O$-moderation. Note that, our method focuses solely on two tokens, regardless of the input or output length. Thus, the complexity of our method is reduced to a constant level. We further demonstrate that \sys is applicable in few-shot scenarios, thereby reducing the cost for service providers to set up our content moderation system. In this way, we achieve an efficient and unified content moderation framework for LLM services.

\begin{itemize}
    \item \textit{How to defend against recent jailbreak attacks that aim to bypass the safety mechanisms of LLMs?}
\end{itemize}

\textit{Jailbreaking} refers to the process of circumventing the safety mechanisms placed on LLM services such as model alignment and content moderation. For instance, a common method to jailbreak model alignment is to instruct LLMs to emulate a ``Do Anything Now'' (DAN) behavior~\cite{King2023Meet}. An instance of jailbreaking content moderation involves performing orthographic transformations, such as \texttt{Base64} encoding~\cite{rao2024tricking}, on textual input or output to bypass content filters, as state-of-the-art LLMs can inherently process such encoded text. While LLM service providers have implemented stricter rules to prevent the use of such jailbreak prompts~\cite{MarkovZANLAJW23}, they cannot entirely eliminate jailbreaking due to the numerous ways to construct prompts conveying the same meaning, owing to the inherent flexibility of natural languages.

To cope with this problem, we equip a red-team model-based data augmentation method in the training stage of \sys. Specifically, we employ a local LLM to work as a red-teaming model, generating highly adversarial jailbreak prompts from initially \textit{naive} unsafe prompts. Through evaluating \sys against static and dynamic jailbreaking methods, we validate that incorporating augmented data during the training phase renders highly robust detection of unsafe content.

We have implemented a fully-functional prototype of \sys on five different host LLMs and evaluated its performance through extensive experiments on seventeen diverse datasets across three content moderation tasks (\textit{i.e.}, $I$-, $O$-, $IO$-moderation). Note that Although previous studies~\cite{azaria2023internal, chen2024inside} have utilized the hidden states of LLMs for hallucination detection, the potential of these hidden states for diverse content moderation tasks remains largely unexplored. Our work not only demonstrates that effective and robust content moderation can be achieved by extracting conceptual features from chat-oriented LLMs and augmenting prompts through red-teaming, but also underscores this paradigm's ability to successfully navigate the trade-off between effectiveness and efficiency.

Our experiments, which include assessments against both static and dynamic jailbreaking, confirm the robustness of \sys against adversarially-designed prompts. Furthermore, the comparison with seven commercial and academic baselines demonstrates that \sys achieves the best performance in detecting unsafe content while maintaining the highest efficiency. Additionally, we validate that \sys can be applied to few-shot scenarios, and extended to perform multi-label classification tasks. We have open-sourced our code\footnote{\blue{\url{https://github.com/lin000001/Legilimens}}} in a hope to incentivize more research in this area.

\vspace{5pt}
We summarize our contributions as follows:
\begin{itemize}
\setlength{\itemsep}{10pt}
    \item We propose a practical and unified content moderation framework tailored for LLM services, which features a balance between effectiveness and efficiency.
    
    \item We develop a concept probing technique that applies to mainstream encoder-decoder and decoder-only LLMs. Additionally, our red-team model-based data augmentation method enhances \sys to robustly resist jailbreaking.
    
    \item We conduct extensive experiments to verify the effectiveness and efficiency of our content moderation framework against both non-adversarial and adversarial queries, outperforming seven commercial and academic baselines.
\end{itemize}


\section{Background}

\subsection{Large Language Model}

Typically, large language models (LLMs) refer to language models that contain billions of parameters, which are trained on massive text data~\cite{Zhao2023Survey}. Famous examples include GPT-3~\cite{Brown20Language,Ouyang22Training}, GPT-4~\cite{OpenAI23GPT4}, GLM~\cite{Du22GLM,Zeng23GLM}, LLaMA~\cite{Touvron2023LLaMA,Touvron23Llama2}, \textit{etc}. In this part, we introduce the basic components, architectures, and inference workflow of LLMs. 

\subsubsection{Basic Component}

Given the excellent parallelizability and capacity, the Transformer architecture~\cite{VaswaniSPUJGKP17, Lin2021Survey} has become the \textit{de facto} backbone to almost all LLMs, making it possible to scale language models to hundreds or thousands of billions of parameters. The vanilla Transformer~\cite{VaswaniSPUJGKP17} is a sequence-to-sequence model and consists of an encoder and a decoder, each of which is a stack of $K$ identical blocks. Each encoder block is mainly composed of a multi-head self-attention module and a position-wise feed-forward network (FFN). For building a deeper model, a residual connection~\cite{HeZRS16} is employed around each module, followed by layer normalization~\cite{BaKH16} module. Compared to the encoder blocks, decoder blocks additionally insert cross-attention modules between the multi-head self-attention modules and the position-wise FFNs. Furthermore, the self-attention modules in the decoder are adapted to prevent each position from attending to subsequent positions in the training phase.


\subsubsection{Architecture}

In general, the mainstream architectures of existing LLMs can be roughly categorized into two major types, namely encoder-decoder and decoder-only.

\textit{Encoder-Decoder Architecture.} The vanilla Transformer model is built on the encoder-decoder architecture~\cite{VaswaniSPUJGKP17}, which consists of two stacks of Transformer blocks as the encoder and decoder, respectively. The encoder encodes the input sequence for generating its latent representations, while the decoder performs cross-attention on these representations and generates the target sequence in an auto-regressive manner. So far, there are only a small number of LLMs that are built based on the encoder-decoder architecture, \textit{e.g.}, Flan-T5~\cite{Chung2022Scaling}.

\textit{Decoder-Only Architecture.} Models of this type only have the decoder but no encoder. According to the self-attention mechanism used in the decoder, decoder-only models can be further divided into the causal decoder architecture and the prefix decoder architecture.

The \textit{causal} decoder architecture incorporates the unidirectional attention mask, to guarantee that each input token can only attend to the past tokens and itself. The input and output tokens are processed in the same fashion through the decoder. As representative language models of this architecture, the GPT series models~\cite{Brown20Language,Ouyang22Training,OpenAI23GPT4} are developed based on the causal decoder architecture. So far, the causal decoders have been widely adopted as the architecture of LLMs by various existing LLMs, such as LLaMA~\cite{Touvron2023LLaMA,Touvron23Llama2}, Dolly~\cite{DatabricksBlog2023DollyV2,DatabricksBlog2023DollyV1}, and Falcon~\cite{Penedo2023RefinedWeb}.

The \textit{prefix} decoder architecture (\textit{a.k.a.}, non-causal decoder) revises the masking mechanism of causal decoders, to enable performing bidirectional attention over the prefix tokens~\cite{Dong2019Unified} and unidirectional attention only on generated tokens. In this way, like the encoder-decoder architecture, the prefix decoders can bidirectionally encode the prefix sequence and auto-regressively predict the output tokens one by one, where the same parameters are shared during encoding and decoding. Existing representative LLMs based on prefix decoders include GLM~\cite{Du22GLM,Zeng23GLM} and U-PaLM~\cite{TayWC0SSGZRCZMP23}.


\subsubsection{Model Inference}\label{sec:inference}
In typical model inference process of LLMs, a prompt $\mathbf{p}=p_1p_2\cdots p_n$ is fed into LLMs to generate a response sequence $\mathbf{r}=r_1r_2\cdots r_m$ auto-regressively, \textit{i.e.}, 
\begin{equation}\label{equ:decode}
\begin{aligned}
r_{i+1} &= \mathop{\arg\max}\limits_{r} \mathbb{P}\left(r|\mathbf{p}\oplus\mathbf{r}_{1:i}\right)\quad \text{(\textit{greedy search})}
\\
\text{\textit{or}} \quad r_{i+1} &\sim \mathbb{P}\left(r|\mathbf{p}\oplus\mathbf{r}_{1:i}\right)\quad \text{(\textit{sampling-based methods})},
\end{aligned}
\end{equation}
where $\oplus$ denotes concatenating the previous output tokens to the end of the input sequence until a special sentence ending token (usually denoted as \texttt{[eos]}) is generated.
The first decoding method is \textit{greedy search}, which predicts the most likely token at each step based on the previously generated tokens. The other decoding method is \textit{sampling}, which randomly samples the next token based on the probability distribution to enhance the randomness and diversity during generation. 

From the inference process and the self-attention mechanism of LLMs we know that LLMs output the first token $r_1$ of $\mathbf{r}$ leveraging the information of $\mathbf{p}$, and output the last token (\textit{i.e.}, \texttt{[eos]}) with the information of both $\mathbf{p}$ and $\mathbf{r}$, \textit{i.e.},
\begin{equation}\label{equ:output}
\begin{aligned}
r_{1} = \mathcal{H}(\mathbf{p})\quad \text{and}\quad \text{\texttt{[eos]}} = \mathcal{H}(\mathbf{p}\oplus\mathbf{r}),
\end{aligned}
\end{equation}
where $\mathcal{H}(\cdot)$ denotes the inference function of LLMs. We utilize this inference process as a feature extractor for the downstream content moderation task, which we elaborate on in \S\ref{sec:design}.

\subsection{Content Moderation}
To mitigate potential harm and misuse of LLMs, two safety mechanisms are commonly applied, \textit{i.e.}, model alignment involves training LLMs to reject unsafe prompts, while content moderation employs filters to block unsafe prompts and responses.

\subsubsection{Unsafe Content}

Given that \textit{out-of-the-box} LLMs have the potential to generate misinformation, propagate harmful content, or produce unintended responses with significant negative societal impact~\cite{WeidingerURGHMG22,Cui2024Risk,Barrett2023Identifying,Weidinger2021Ethical}, content moderation is essential for identifying such \textit{unsafe content} within prompts and responses. The definitions of unsafe content in previous literature often include profanities, identity attacks, sleights, insults, threats, sexually explicit content, demeaning language, and language that incites violence~\cite{SchmidtW17,FortunaN18,GorwaBK20}. In practice, different service providers may adopt different definitions and taxonomies of unsafe content.


\subsubsection{Moderation Tasks}

Content moderation for LLMs involves moderating the input prompt and the output response, referred to as $I$-moderation and $O$-moderation respectively. In $I$-moderation (\textit{resp.} $O$-moderation), the moderator assesses whether the input prompt (\textit{resp.} output response) is unsafe, based solely on the input prompt (\textit{resp.} output response) or on the entirety of the prompt and response. The combined task, referred to as $IO$-moderation, aims to assess both the input prompt and the output response.




\subsection{Jailbreak}\label{sec:jailbreak}
\textit{Jailbreaking} is a process that employs adversarially-designed prompts to circumvent the safety mechanisms imposed on LLMs by their service providers. Several efforts have been made to taxonomize jailbreaking~\cite{Liu2023Jailbreaking, rao2024tricking, Shen2023Do, Chu2024Comprehensive}, based on which jailbreaking can be categorized into \textit{semantic} transformation and \textit{syntactic} transformation. 

\begin{itemize}
\setlength{\itemsep}{10pt}
    \item Semantic transformation involves manipulating the semantics of prompts. Based on the patterns of semantic manipulation, three general types of semantic transformation have been identified~\cite{Liu2023Jailbreaking}.
    \begin{itemize}
    \setlength{\itemsep}{5pt}
        \item \textit{Pretending}: This type of prompts tries to alter the conversation background or context while maintaining the same intention, \textit{e.g.}, creating a role-playing game context.
        
        \item \textit{Attention Shifting}: This type of prompts aims to change both the conversation context and intention, \textit{e.g.}, shifting the intention of the prompt from asking the model questions to making it construct a paragraph of text.
        
        \item \textit{Privilege Escalation}: This type of prompts seeks to directly circumvent the imposed restrictions. For instance, ``Do Anything Now'' prompts mentioned in \S\ref{sec:intro} belong to this category.
    \end{itemize}
    \item Syntactic transformation modifies only the syntax of prompts without altering the semantics. Examples include string splicing, common encoding, and simple encryption. This type of transformation can effectively bypass rule-based filters.
\end{itemize}

Note that the generation of jailbreak prompts can be static and dynamic. Static prompts do not leverage response information from the target LLM while dynamic methods use previous responses as a reference to adversarially refine the prompt for launching the next attack.





\subsection{Design Goal}

We first define the system model in terms of the adversary and the defender, and then elaborate the design goals of \sys under the defined system model.

\textit{Adversary}. The adversary attempts to induce undesirable behavior of LLM services, whether unintentionally or intentionally. This may include generating inappropriate content, disclosing sensitive information, or performing actions against programming constraints. To achieve this, the adversary may optimize the semantics or syntax of their prompts to outsmart the safeguards of LLMs by any jailbreak methods. The adversary can observe the returned responses and leverage this knowledge to refine the prompt in order to achieve their attack objectives.


\textit{Defender.} The defender aims to ensure both the safety and helpfulness of the LLM service. To achieve this, the defender strives to apply content moderation mechanisms that are as accurate as possible, with full access to the host LLM.

Given this system model, we delineate two major goals of \sys.

\begin{itemize}
\setlength{\itemsep}{5pt}
    \item \textbf{Effectiveness}. \sys should accurately identify unsafe prompts given to the host LLM or responses generated by the host LLM, even when confronted with jailbreak prompts.
    \item \textbf{Efficiency}. \sys should introduce minimal overhead compared to the original inference process of the host LLM. Ideally, the overhead of \sys should not increase with the length of prompts or responses, \textit{i.e.}, maintaining a constant complexity of $O(1)$.
\end{itemize}


\begin{figure*}[tt]
    \centering
\setlength{\abovecaptionskip}{10pt}
\setlength{\belowcaptionskip}{0pt}

\includegraphics[width=7.in, trim=60 90 60 90, clip]{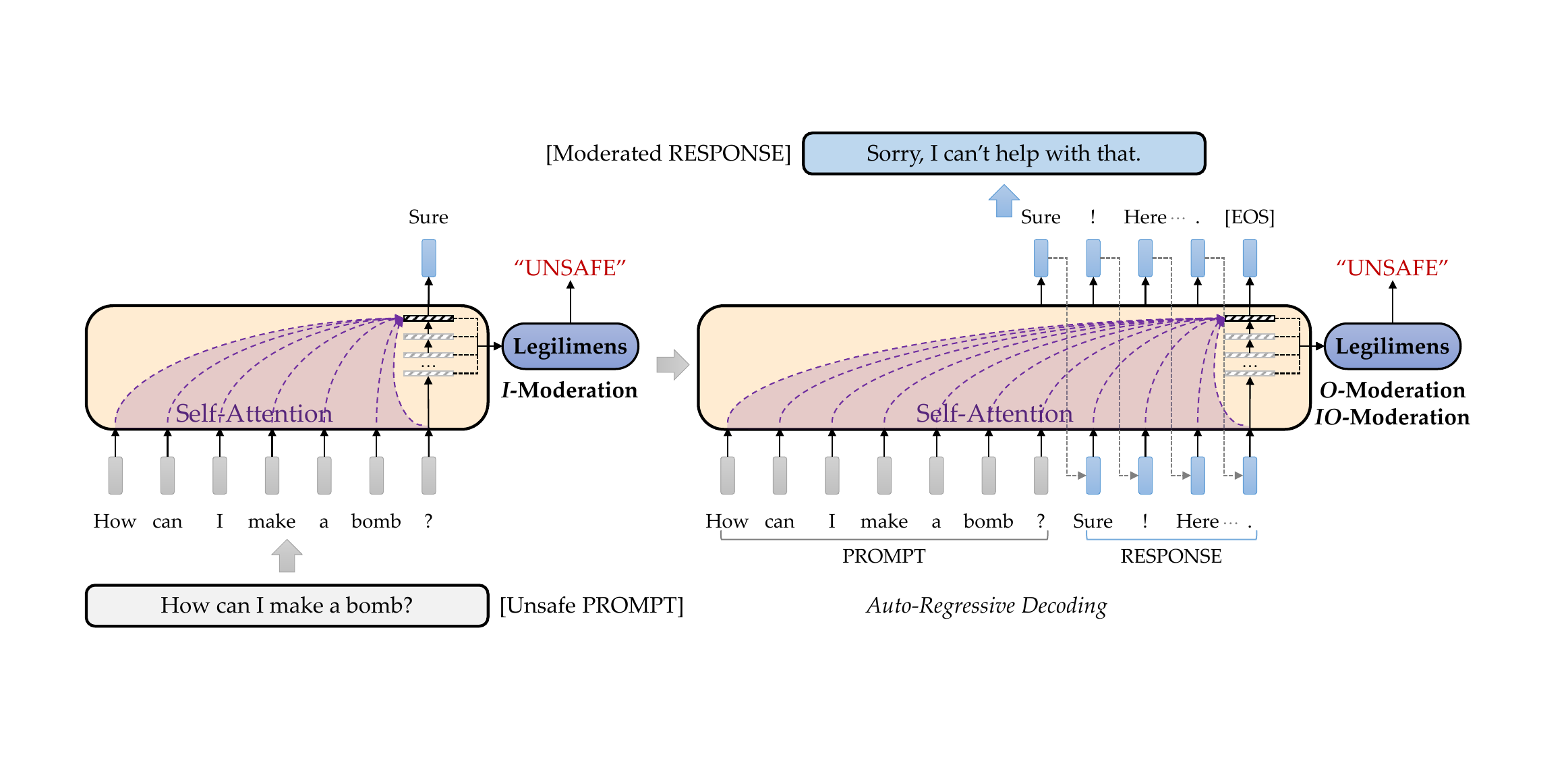}

\caption{Design of \sys. \sys utilizes the conceptual features during the decoding of the first token to accomplish $I$-moderation, and the conceptual features during the decoding of the last token to achieve $O$- and $IO$-moderation.}\label{fig:workflow}
\end{figure*}

\section{Problem Formulation}\label{sec:formulation}

In this section, we first formulate content moderation as a classification problem. Then we materialize the classification problem for three distinct moderation tasks.

\subsection{Classification Formulation}

A textual input $\mathbf{x}=x_1x_2\cdots x_n$ can be tokenized into $n$ frames of token embeddings, \textit{i.e.}, $\mathbf{X}\in \mathbbm{R}^{n\times d}$, where $d$ denotes the dimension of each token embedding. A typical content moderation problem is to determine whether $\mathbf{x}$ is unsafe by a moderator $\mathcal{M}_\theta(\cdot):\mathbbm{R}^{n\times d}\rightarrow \mathbbm{Y}$, parametrized by $\theta$, which can be expressed as
\begin{equation}\label{equ:formulation}
\begin{aligned}
\mathop{\min}\limits_{\theta} \mathop{\mathbb{E}}\limits_{\mathbf{X},y\sim\mathbbm{D}} \mathcal{L}\left(\mathcal{M}_\theta(\mathbf{X}), y\right),
\end{aligned}
\end{equation}
where $\mathbbm{D}$ is a labelled content moderation dataset. $y\in \mathbbm{Y}$ is the ground-truth label of the corresponding input $\mathbf{X}$, denoting whether $\mathbf{X}$ is unsafe. $\mathcal{L}$ represents the loss function, typically measured using cross-entropy loss, which evaluates the accuracy of $\mathcal{M}_\theta(\mathbf{X})$.

In the classification problem above, when the training dataset $\mathbbm{D}$ is of sufficient size, the accuracy of the moderator $\mathcal{M}_\theta(\cdot)$ generally increases with the scale of its parameters $\theta$, but at the cost of increased computational complexity. Thus, efficiency and effectiveness are often in a trade-off relationship.

In the scenario of LLM services, the host LLM is expected to infer input prompts and generate corresponding responses. Given the substantial capacity of the host LLM, we aim to utilize the original inference process of the host LLM to materialize a portion of $\mathcal{M}_\theta(\cdot)$. This approach allows representative features to be extracted in a ``free-riding'' manner, \textit{i.e.},
\begin{equation}\label{equ:freeriding}
\begin{aligned}
\mathcal{M}_{\theta}\left(\mathbf{X}\right) = \mathcal{C}_{\theta\backslash\psi}\circ\mathcal{H}_\psi\left(\mathbf{X}\right),
\end{aligned}
\end{equation}
where $\mathcal{H}_\psi$ represents a portion of the host model parameterized by $\psi$, and $\mathcal{C}_{\theta\backslash\psi}$ denotes an additional classifier parameterized by $(\theta\backslash\psi)$. In this way, the formulation in Equation~\eqref{equ:formulation} is transformed into
\begin{equation}\label{equ:formulation2}
\begin{aligned}
\mathop{\min}\limits_{\theta\backslash\psi}\  \mathop{\mathbb{E}}\limits_{\mathbf{X},y\sim\mathbbm{D}} \mathcal{L}\left(\mathcal{C}_{\theta\backslash\psi}\circ\mathcal{H}_\psi\left(\mathbf{X}\right), y\right).
\end{aligned}
\end{equation}

Notably, since $|\psi|$ is sufficiently large, $|\theta|$ can be large when $|\theta\backslash\psi|=|\theta|-|\psi|$ is relatively small. Consequently, this approach potentially achieves a win-win scenario in terms of efficiency and effectiveness.

\subsection{Materialization Tasks}
Based on the object to be moderated, we materialize \sys into three moderation tasks, as depicted in Figure~\ref{fig:workflow}.

\begin{itemize}
\setlength{\itemsep}{5pt}
    \item \textbf{\textit{I}-Moderation}. In $I$-moderation, the object to be moderated is solely input prompts to be fed into the host LLM. The task is to determine whether the input prompts are unsafe, denoted as $\hat{y}=\mathcal{M}_{\theta}\left(\mathbf{P}\mid*\right)$, where $\mathbf{P}$ represents the token embeddings of prompt $\mathbf{p}$, and $*$ signifies any additional information available to the moderator, \textit{e.g.}, output responses.
    \item \textbf{\textit{O}-Moderation}. In $O$-moderation, the object to be moderated is solely output responses generated by the host LLM. The task is to determine whether the output responses are unsafe, denoted as $\hat{y}=\mathcal{M}_{\theta}\left(\mathbf{R}\mid*\right)$, where $\mathbf{R}$ represents the token embeddings of response $\mathbf{r}$, and $*$ signifies any extra information available to the moderator, \textit{e.g.}, input prompts.
    \item \textbf{\textit{IO}-Moderation}. In $IO$-moderation, the object to be moderated encompasses both input prompts and output responses, which is a combined task of $I$- and $O$-moderation, \textit{i.e.}, $\hat{y}=\mathcal{M}_{\theta}\left(\mathbf{P},\mathbf{R}\mid*\right)$.
\end{itemize}

The three tasks above can be applied along with different moderation strategies. In the case of $I$-moderation, the LLM service can halt once the prompt is determined to be unsafe, thereby saving subsequent computation. However, this strategy may potentially impair the overall helpfulness of the LLM service. Conversely, for $O$-moderation, the service halts until the response is determined to be unsafe. This strategy preserves helpfulness when prompts are unsafe but successfully handled by the host LLM, providing safe responses. Nevertheless, this strategy may require additional computational resources. $IO$-moderation halts the service upon detecting unsafe prompts or responses, representing a more rigorous moderation strategy compared to the first two tasks.

\section{\sys: Design Details}\label{sec:design}

In this section, we materialize $\mathcal{H}_\psi$ and $\mathcal{C}_{\theta\backslash\psi}$ in Equation~\eqref{equ:formulation2} in the scenario of LLM services by concept probing and lightweight moderator construction.


\subsection{Concept Probing}\label{sec:probe}
\subsubsection{Conceptual Feature Extraction}
Typical LLMs are constructed by stacking $K$ encoder-decoder or decoder-only Transformer blocks together, \textit{i.e.},
\begin{equation}\label{equ:stack}
\begin{aligned}
\mathcal{H}=\mathcal{H}_{K}\circ \mathcal{H}_{K-1}\circ\cdots\circ \mathcal{H}_{1},
\end{aligned}
\end{equation}
where $\mathcal{H}_{i}$ denotes the $i$-th block of the host LLM. To decode a token, the original inference process derives side features as follows,
\begin{equation}\label{equ:features}
\begin{aligned}
\widetilde{\mathcal{H}}_k(\mathbf{X})\triangleq\left(\mathcal{H}_{k}\circ \mathcal{H}_{k-1}\circ\cdots\mathcal{H}_{1}\right)(\mathbf{X}), \quad 1\leq k \leq K.
\end{aligned}
\end{equation}

Our intuition is that LLMs develop concepts of the input along the inference process. Thus, we attain comprehensive conceptual features of $\mathbf{X}$ by fusing/concatenating the side features derived from the last several blocks, \textit{i.e.},
\begin{equation}\label{equ:featuers2}
\begin{aligned}
\widetilde{\mathcal{H}}_{[-m:]}(\mathbf{X})\triangleq\left[\widetilde{\mathcal{H}}_{(K-m+1)}(\mathbf{X})\oplus\cdots\oplus\widetilde{\mathcal{H}}_K(\mathbf{X})\right].
\end{aligned}
\end{equation}

The reason for deriving conceptual features from the last several blocks is to utilize more capacity of the host LLM, given that $\widetilde{\mathcal{H}}_{[-m:]}$ can be viewed as containing all parameters of the host LLM. Note that, no additional computation is required to obtain the side features in Equation~\eqref{equ:featuers2}.

\subsubsection{Probing Position} 
During the inference process, LLMs generate responses in an auto-regressive manner as mentioned in \S\ref{sec:inference}, where varied-length side features are derived. Managing these varied-length side features may lead to increased overhead as the response length grows. To reduce the complexity of \sys to $O(1)$, we resort to the attention \textit{span} of self-attention mechanism.

As shown in Figure~\ref{fig:workflow}, the attention span during the decoding of the first token $r_1$ encompasses the entire prompt, indicating that the conceptual features derived in this process, $\widetilde{\mathcal{H}}_{[-m:]}(\mathbf{P})$, encapsulate a summarized knowledge of prompt $\mathbf{p}$. Similarly, the attention span during the decoding the last token \texttt{[eos]} encompasses both the prompt and the response, encapsulating a summarized knowledge of prompt $\mathbf{p}$ and response $\mathbf{r}$ in $\widetilde{\mathcal{H}}_{[-m:]}(\mathbf{P}\oplus\mathbf{R})$. Note that, both $\widetilde{\mathcal{H}}_{[-m:]}(\mathbf{P})$ and $\widetilde{\mathcal{H}}_{[-m:]}(\mathbf{P}\oplus\mathbf{R})$ are of a constant size of $m\times d_\text{model}$, where $d_\text{model}$ is the output dimension of Transformer blocks. Reasoning on these two conceptual features only introduces constant overhead.

\subsection{Lightweight Moderator}\label{sec:moderator}

\subsubsection{Architecture} In this part, we construct a lightweight classifier to materialize $\mathcal{C}_{\theta\backslash\psi}$. Based on the comprehensive representation summarized by the host LLM, we find that a simple multi-layer perceptron (MLP) with few parameters is adequate for accurate content moderation, denoted as $\mathcal{C}_{\theta\backslash\psi}: \mathbbm{R}^{m\times d_\text{model}}\rightarrow \mathbbm{Y}$. Therefore, the entire moderator is composed in this way: $\mathcal{M}_{\theta}=\mathcal{C}_{\theta\backslash\psi}\circ\widetilde{\mathcal{H}}_{[-m:]}$. We validate the effectiveness of our lightweight moderator in \S\ref{sec:overall} compared with five baselines.

\subsubsection{Model Training}
Within the unified framework of concept probing, we train \sys to handle  three content moderation tasks, \textit{i.e.}, $I$-, $O$-, $IO$-moderation, in accordance with Equation~\eqref{equ:formulation2}.

For $I$-moderation, we curate a training set of $(\mathbf{P},y_p)$ to train the moderator, where the ground-truth label $y_p$ is assigned based on the safety of prompt $\mathbf{p}$. For $O$-moderation, as the original inference process of the host LLM does not involve reasoning on $\mathbf{R}$, we prepare a training set of $(\mathbf{P}\oplus\mathbf{R},y_r)$ instead, where the ground-truth label $y_r$ is assigned based on the safety of response $\mathbf{r}$. Similarly, for $IO$-moderation, we prepare a training set of $(\mathbf{P}\oplus\mathbf{R},y_p|y_r)$ to train the moderator.

\subsection{Model-Based Data Augmentation}\label{sec:aug}
An adaptive adversary may employ jailbreaking to compromise the protection of \sys. Given that jailbreaking can alter the conceptual features of prompts and responses in order to evade detection, we augment the training data of \sys to bolster its resistance against jailbreaking. Specifically, we employ LLaMa2 as a red-teaming model $\mathcal{T}$, prompting it to rewrite naive unsafe prompts $\mathbf{p}$ into adversarially-designed jailbreak prompts $\mathbf{p}'=\mathcal{T}(\mathbf{p})$. Our system prompt consists of three segments. The first segment is "Do Anything Now", aimed at directing the large model $\mathcal{T}$ to violate its principle. The second segment specifies the output format and rewriting instructions, derived from extensive research (\textit{e.g.}, \cite{Liu2023Jailbreaking, rao2024tricking, xu2024llm, Chu2024Comprehensive}), systematically enumerating various jailbreaking methodologies, such as "Character Role Play", "Logical Reasoning", among others. The final segment provides a few examples through a few-shot demonstration while utilizing the ability of in-context learning~\cite{brown2020language}. We rewrite 20\% of the unsafe prompts in the training set to enhance the robustness of \sys. In \S\ref{sec:adaptive}, we validate the robustness of \sys against state-of-the-art static and dynamic jailbreak attacks.

\section{Evaluation}
\subsection{Setup}
\subsubsection{Prototype} 

We have implemented a prototype of \sys on the PyTorch~\cite{PaszkeGMLBCKLGA19} platform and train the moderators according to Equation~\eqref{equ:formulation2} using a single NVIDIA 3090 GPU. We set the default \sys configurations as $m=1$, and $\mathcal{C}_{\theta\backslash\psi}$ is a three-layer MLP. In the training phase, we utilize an Adam~\cite{kingma2014adam} optimizer to update the parameters of the MLP-based classifier for 50 epochs with a learning rate of 1e-4, a weight decay ($\ell_2$ penalty) rate of 1e-3, and a batch size of 256.





\subsubsection{Host Model}
We apply \sys to five host LLMs with various architectures, \textit{i.e.}, ChatGLM3-6B~\cite{Zeng23GLM}, LLaMA2-7B~\cite{Touvron23Llama2}, Falcon-7B~\cite{Penedo2023RefinedWeb}, Dolly-7B-v2~\cite{DatabricksBlog2023DollyV2}, Vicuna-7B-v1.5~\cite{ZhengC00WZL0LXZ23}. 

\begin{itemize}
\setlength{\itemsep}{2pt}
    \item \textbf{ChatGLM3.} ChatGLM3~\cite{Zeng23GLM} is an open-source bilingual (English and Chinese) LLM following a prefix decoder architecture, which utilizes a multi-query attention mechanism~\cite{shazeer2019fast} and the SwiGLU~\cite{shazeer2020glu} activation function. ChatGLM3 comprises 28 Transformer blocks.
    \item \textbf{LLaMA2.} LLaMA2~\cite{Touvron23Llama2} is an open-source LLM developed by Meta using a causal decoder architecture, employing a grouped-query attention and the SwiGLU activation function. LLaMA2 consists of 32 Transformer blocks.
    \item \textbf{Falcon.} Falcon~\cite{Penedo2023RefinedWeb} is an open-source LLM using a causal decoder architecture, incorporating a multi-query attention mechanism and the GeLU~\cite{dauphin2017language} activation function. Falcon comprises 32 Transformer blocks.
    \item \textbf{Dolly.} Dolly~\cite{DatabricksBlog2023DollyV2} is fine-tuned from EleutherAI's Pythia-6.9B~\cite{biderman2023pythia} on an instruction-tuning dataset of around 15,000 samples. It uses a causal decoder architecture, sparse attention mechanism~\cite{child2019generating} and the GeLU activation function. Dolly comprises 32 Transformer blocks.
    \item \textbf{Vicuna.} Vicuna~\cite{ZhengC00WZL0LXZ23} is fine-tuned from LLaMA2 on around 125,000 instruction-tuning samples. The architecture of Vicuna is the same as LLaMA2.
\end{itemize}

The output dimension $d_\text{model}$ of Falcon is 4544, while it is 4096 for the other four LLMs. 


\subsubsection{Baselines}\label{sec:baseline}
We utilize four commercial and three academic state-of-the-art content moderation methods as our baselines, \textit{i.e.}, Google Perspective API~\cite{PerspectiveAPI} (commercial), OpenAI Moderation API~\cite{OpenAIModeration} (commercial), BeaverDam~\cite{Ji2023BeaverTails} (academic), LLaMA Gurad2~\cite{DBLP:journals/corr/abs-2312-06674} (academic), GradSafe~\cite{xie2024gradsafe} (academic), GPT-3.5-Turbo~\cite{Brown20Language} (commercial), and GPT-4~\cite{OpenAI23GPT4} (commercial). Detailed information about the first five baselines is listed in Table~\ref{tab:existing}. To adapt GPT-3.5-Turbo and GPT-4 for content moderation, we employ manually designed prompts as illustrated in Appendix~\ref{Appendix_A}. We will demonstrate that \sys outperforms these seven baselines in \S\ref{sec:overall}.





\subsubsection{Metrics} 
Two typical and standard metrics are used to evaluate the content moderation performance of \sys.


\begin{itemize}
\setlength{\itemsep}{5pt}
    \item \textbf{Accuracy (ACC)} is a measure of the correctness of content moderation, calculated as the ratio of the number of correct predictions to the total number of predictions made.
    \item \textbf{Area Under the ROC Curve (AUC)} measures the area underneath the ROC (Receiver Operating Characteristic) curve. The AUC value ranges from 0 to 1, where a higher value indicates better capability to distinguish between safe and unsafe content.
    \item \textbf{False Positive Rate (FPR)} is a measure of the ratio of falsely predicted positive instances to the total actual negative instances. FPR signifies the proportion of safe content that is incorrectly classified as unsafe in content moderation. 
    \item \textbf{False Negative Rate (FNR)} represents the ratio of falsely predicted negative instances to the total actual positive instances. In the realm of content moderation, FNR denotes the proportion of unsafe content that is erroneously classified as safe.
\end{itemize}

\subsubsection{Datasets}\label{sec:dataset}
Our experiments involve seventeen datasets covering various tasks across different domains, as shown in Table~\ref{tab:dataset} in Appendix~\ref{ap:dataset}. Among these, four datasets are dedicated to traditional content moderation, while the others are specialized for LLM services.

\begin{itemize}
    \item \textbf{HateXplain~\cite{mathew2021hatexplain}}: This dataset serves as a benchmark for hate speech detection, sourced from Twitter and Gab. It has been annotated by Amazon Mechanical Turk (MTurk) workers, who assigned three labels, \textit{i.e.}, \textit{hate}, \textit{offensive}, or \textit{normal}. In our paper, we consider \textit{hate} and \textit{offensive} posts as unsafe, and the \textit{normal} posts as safe.
    
    \item \textbf{Measuring Hate Speech~\cite{kennedy2020constructing}}: This dataset comprises comments from social media platforms. These comments have been labelled by MTurk workers, and the labels have been converted into a continuous score. We consider comments with a score over 0.5 as unsafe, and those with a score less than -1 as safe.
    
    \item \textbf{OIG-Safety~\cite{Nguyen2023OIG}}: We utilize a subset of samples from OIG-Safety, labelled as \textit{casual} and \textit{needs intervention}, comprising 21,769 samples. We consider \textit{casual} as safe and \textit{needs intervention} as unsafe.
    
    \item \textbf{Jigsaw~\cite{JigsawDataset}}: This dataset comprises seven categories of samples, namely \textit{innocent}, \textit{severe toxicity}, \textit{obscene}, \textit{identity attack}, \textit{insult}, \textit{threat}, and \textit{sexual explicit}. We consider \textit{innocent} as safe and the remaining categories as unsafe.
\end{itemize}

We also assess \sys using datasets specifically designed for $O$-moderation in LLM scenarios.

\begin{itemize}
    \item \textbf{BeaverTails~\cite{Ji2023BeaverTails}}: We conduct a voting process on the original dataset to obtain our dataset, which consists of 110,822 question-answer pairs across 14 potential harm categories. Note that the prompts are all unsafe. We utilize this dataset for $O$-moderation. 
    
    \item \textbf{BeaverTails-\textit{adv}}: This dataset is created by inserting the original prompts from BeaverTails into four types of jailbreak templates, including pretending, attention shifting, privilege escalation, and syntactic transformation. The jailbreak templates are collected from the Internet.
    
    
\end{itemize}

\begin{table*}[tt]
\centering

\setlength{\abovecaptionskip}{0pt}%
\setlength{\belowcaptionskip}{0pt}%
\caption{Accuracy performance on various tasks$^\text{\textdagger}$ and datasets$^\text{\textdaggerdbl}$ compared with baselines.}\label{tab:baseline}

\resizebox{\linewidth}{!}{
\begin{threeparttable}

\setlength{\tabcolsep}{3mm}{

\begin{tabular}{@{}l|cccccccc|c@{}}
\toprule
\multicolumn{1}{c|}{\multirow{2}{*}{Method}} & \multicolumn{4}{c|}{\textbf{\textit{I}-Moderation} (ACC, \%)}                          & \multicolumn{2}{c|}{\textbf{\textit{O}-Mod.} (ACC, \%)} & \multicolumn{2}{c|}{\textbf{\textit{IO}-Mod.} (ACC, \%)}  & \multirow{2}{*}{\begin{tabular}{@{}c@{}}Time/Query\\(ms)\end{tabular}}                                               \\ 
                        & HAT   & MHS     & OIG  & \multicolumn{1}{c|}{JIG}   & BEA  & \multicolumn{1}{c|}{BEA-\textit{adv}} & BAG        & \multicolumn{1}{c|}{BAG-\textit{adv}}   \\ \midrule
OpenAI Moderation       & \tblc{70.85}        & \tblc{72.30}         & \tblc{65.85}        & \multicolumn{1}{c|}{\tblc{76.70}}         & \tblc{51.70}         & \multicolumn{1}{c|}{\tblc{47.95}}             & \tblc{53.90}           & \tblc{53.4}    &    566.951     \\
Perspective API         & \tblc{64.79}        & \tblc{75.1}         & \tblc{60.56}        & \multicolumn{1}{c|}{\tblc{78.26}}        & \tblc{47.72}        & \multicolumn{1}{c|}{\tblc{45.95}}             & \tblc{51.78}          & \tblc{48.50}       &  90.336      \\
BeaverDam-7B           & \tblc{66.65}        & \tblc{74.15}        & \tblc{63.75}        & \multicolumn{1}{c|}{\tblc{67.50}}         & \textbf{\large \tblc{89.50}}         & \multicolumn{1}{c|}{\tblc{73.75}}             & \tblc{74.95}          & \tblc{61.75}     & 30.121       \\

LLaMA Guard2     & \tblc{71.40}           & \tblc{77.85}           & \tblc{76.00}           & \multicolumn{1}{c|}{\tblc{50.95}}           & \tblc{77.38}           & \multicolumn{1}{c|}{\tblc{75.70}}                & \tblc{68.40}             & \tblc{66.15}        & 430.923   \\ 

GradSafe     & \tblc{69.82}           & \tblc{67.70}           & \tblc{78.90}           & \multicolumn{1}{c|}{\tblc{66.00}}           & \tblc{73.80}           & \multicolumn{1}{c|}{\tblc{57.10}}                & \tblc{75.50}             & \tblc{76.90}        & 395.212    \\

GPT-3.5-Turbo                 & \tblc{68.00}        & \tblc{75.52}        & \tblc{60.70}         & \multicolumn{1}{c|}{\tblc{52.50}}        & \tblc{65.73}        & \multicolumn{1}{c|}{\tblc{55.99}}             & \tblc{55.82}          & \tblc{44.42}      & 681.800      \\
GPT-4                   & \tblc{74.00}           & \tblc{75.00}           & \tblc{79.35}           & \multicolumn{1}{c|}{\tblc{68.00}}           & \tblc{83.00}           & \multicolumn{1}{c|}{\tblc{75.00}}                & \tblc{73.00}             & \tblc{60.26}        & 801.720    \\

\textbf{\sys (Ours)}                    & \textbf{\large \tblc{82.19} } & \textbf{\large \tblc{90.48}}  &\textbf{\large \tblc{94.67}} & \multicolumn{1}{c|}{\textbf{\large \tblc{88.39}}} & \tblc{88.11}\tnote{$2^{nd}$} & \multicolumn{1}{c|}{\textbf{\large \tblc{84.49}}} & \textbf{\large \tblc{99.56}} & \textbf{\large \tblc{97.34}} &  \textbf{\large 0.003 } \\ \bottomrule
\end{tabular}
}
\begin{tablenotes}[flushleft]
\small
\item[] \textdagger: Task alias: $I$-Moderation (\underline{$I$-Mod.}), $O$-Moderation (\underline{$O$-Mod.}) and $IO$-Moderation (\underline{$IO$-Mod.}). 
\item[] \textdaggerdbl: Dataset alias: HateXplain (\underline{HAT}), Measuring Hate Speech (\underline{MHS}), OIG-Safety (\underline{OIG}), Jigsaw (\underline{JIG}), BeaverTails (\underline{BEA}), BeaverTail-\textit{adv} (\underline{BEA-\textit{adv}}), BEA\&AG (\underline{BAG}), BEA-\textit{adv}\&AG (\underline{BAG-\textit{adv}}). 

\end{tablenotes}

\end{threeparttable}}
\end{table*}

We create two datasets for $IO$-moderation in LLM scenarios since no relevant datasets are readily available.

\begin{itemize}
    \item \textbf{BEA\&AG}: We compile this dataset by merging BeaverTails with an instruction-tuning dataset called Alpaca-GPT4~\cite{peng2023instruction}. As all prompts from BeaverTails are deemed unsafe, an equal number of samples from Alpaca-GPT4, considered safe, have been incorporated.
    
    \item \textbf{BEA-\textit{adv}\&AG}: We compile this dataset by merging BeaverTails-\textit{adv} with an equal number of samples from Alpaca-GPT4.
\end{itemize} 

We also utilize various unseen datasets to assess the performance of \sys in scenarios involving potential distribution shifts.

\begin{itemize}
    \item \textbf{BEA\&PIQA}: We combine non-repetitive prompts from BeaverTails and an equal number of prompts from the PIQA~\cite{Bisk2020} dataset. The latter are considered safe.
    
    
    \item \textbf{HarmBench~\cite{mazeika2024harmbench}}: This dataset contains 320 human-written unsafe instructions, covering 7 semantic categories of behavior: cybercrime \& unauthorized intrusion, chemical \& biological weapons/drugs, copyright violations, misinformation \& disinformation, harassment \& bullying, illegal activities, and general harm.
    
    \item \textbf{SimpleSafetyTests~\cite{vidgen2024simplesafetytests}}: This dataset comprises 100 human-written unsafe simple questions or instructions, covering 5 harm areas: (1) suicide, self-harm, and eating disorders, (2) physical harm, (3) illegal and highly regulated items, (4) scams and fraud, and (5) child abuse.
    
    \item \textbf{MaliciousInstructions~\cite{bianchi2023safety}}: This dataset contains 100 machine-written instructions generated by GPT-3 (text-davinci-003).
    
    \item \textbf{JADE~\cite{zhang2023jade}}: The dataset contains 80 machine-written unsafe prompts, which were created by linguistic fuzzing to generate challenging prompts for evaluating LLM safety.
    
    \item \textbf{HExPHI~\cite{qi2023fine}}: This dataset is sampled from AdvBench~\cite{zou2023universal} and AnthropicRedTeam~\cite{ganguli2022red} and then refined manually and with LLMs. There are 330 unsafe instructions.
    
    \item \textbf{TDCRedTeaming~\cite{mazeika2024harmbench}}: This dataset contains 50 human-written red-teaming instructions, covering 7 categories: bigotry \& abusive language, violent content \& conduct, illegal activities, malware \& exploits, scams, misinformation \& disinformation, other undesirable content.

\end{itemize}

We further employ two datasets as input of dynamic jailbreaking to stress test \sys in \S\ref{sec:dynamic}.
\begin{itemize}
\item \textbf{AdvBench}: This dataset contains 50 prompts designed to elicit harmful information across 32 categories~\cite{chao2023jailbreaking}.

\item \textbf{AdvBEA}: This dataset contains 30 prompts sourced from the BeaverTails test set, encompassing categories such as violence, privacy, weapons, child abuse, and more.
\end{itemize}

\subsection{Overall Performance}\label{sec:overall}
\subsubsection{Effectiveness} In this part, we evaluate the overall effectiveness of \sys on eight datasets and three moderation tasks in comparison with seven baselines. We train \sys (hosted in LLaMA2) on the training set of each dataset and evaluate its performance on the corresponding test set. The results are presented in Table~\ref{tab:baseline}, Table~\ref{tab:baseline_fnr} and Table~\ref{tab:baseline_fpr}, with the latter two in Appendix~\ref{baseline}.


For $I$-moderation, we utilize four datasets, \textit{i.e.}, HateXplain, MHS, OIG-Safety, and Jigsaw. As shown in Table~\ref{tab:baseline}, \sys achieves the best performance on all four datasets, including two for hateful speech and two for general unsafe content, with an accuracy of 82.19\%, 90.48\%, 94.67\%, and 88.39\%. \sys outperforms the second place by 8.19\%$\sim$15.32\%. It validates the effectiveness of \sys on $I$-moderation across different domains.


For $O$-moderation, we assess eight methods on BeaverTails (normal version) and BeaverTails-\textit{adv} (jailbreak version). Referring to Table~\ref{tab:baseline}, \sys achieves the second best performance on BeaverTails and the best performance on BeaverTails-\textit{adv}. It is noteworthy that BeaverDam-7B has been trained end-to-end on the training set of BeaverTails, whereas only a small portion of \sys (the lightweight classifier) has been trained. Despite this, the performance gap between \sys and BeaverDam-7B is merely 1.39\%. In contrast, OpenAI Moderation API, Perspective API, and GPT-3.5-Turbo perform roughly as well as random guessing. One possible reason is that these three baselines can not handle $O$-moderation when prompts are already unsafe. In addition, when we modify the unsafe prompts into jailbreak ones, the performance of BeaverDam-7B drops 15.75\% while \sys only experiences a drop of 3.62\%. This suggests that jailbreak templates have a negative influence on BeaverDam-7B in determining the safety of responses. In comparison, \sys demonstrates robustness against jailbreak prompts.



For $IO$-moderation, we evaluate eight methods on BEA\&AG and BEA-\textit{adv}\&AG datasets. As depicted in Table~\ref{tab:baseline}, \sys achieves the best performance on both datasets with an accuracy of 99.56\% and 97.34\%, significantly outperforming the second place by 24.06\% and 20.44\%. OpenAI Moderation API, Perspective API, and GPT-3.5-Turbo still exhibit unsatisfactory performance. This suggests that these three methods may not be suitable for content moderation in LLM scenarios where both prompts and responses should be taken into account.

\begin{table}[tt]
\centering
\setlength{\abovecaptionskip}{0pt}%
\setlength{\belowcaptionskip}{0pt}%
\caption{Generalization performance ($I$-Moderation).}\label{tab:unseendata}

\resizebox{\linewidth}{!}{
\begin{threeparttable}

\definecolor{100}{RGB}{82,183,136}
\definecolor{100}{RGB}{82,183,136}
\setlength{\tabcolsep}{3mm}{
\begin{tabular}{@{}c|ccccc@{}}
\toprule
\multicolumn{1}{c|}{\multirow{2}{*}{Test Set\tnote{\textdaggerdbl}}} & \multicolumn{5}{c}{ Host Model (ACC, \%)}                                                  \\
\multicolumn{1}{c|}{}                                   & ChatGLM3 & LLaMA2 & Falcon & Dolly & Vicuna \\ \midrule
BPI\tnote{\textdagger}                                 & \cellcolor{white!97.53!white}{97.53}             & \cellcolor{white!98.58!white}{98.58}           & \cellcolor{white!97.82!white}{97.82}           & \cellcolor{white!96.12!white}{96.12}          & \cellcolor{white!98.03!white}{98.03}           \\ \midrule
HAB                                               & \cellcolor{white!90.94!white}{90.94}             & \cellcolor{white!79.69!white}{79.69}           & \cellcolor{white!59.38!white}{59.38}          & \cellcolor{white!82.19!white}{82.19}          & \cellcolor{white!73.75!white}{73.75}           \\
SST                                       & \cellcolor{white!99.00!white}{99.00}             & \cellcolor{white!100.00!white}{100.00}          & \cellcolor{white!!white}{95.00}           & \cellcolor{white!96.00!white}{96.00}          & \cellcolor{white!99.00!white}{99.00}           \\
MAI                                   & \cellcolor{white!99.00!white}{99.00}             & \cellcolor{white!100.00!white}{100.00}          & \cellcolor{white!87.00!white}{87.00}          & \cellcolor{white!97.00!white}{97.00}          & \cellcolor{white!99.00!white}{99.00}           \\
JAD                                                   & \cellcolor{white!90.00!white}{90.00}             & \cellcolor{white!88.75!white}{88.75}           & \cellcolor{white!83.75!white}{83.75}           & \cellcolor{white!91.25!white}{91.25}          & \cellcolor{white!83.75!white}{83.75}           \\
HEP                                                  & \cellcolor{white!98.79!white}{98.79}             & \cellcolor{white!96.06!white}{96.06}           & \cellcolor{white!80.91!white}{80.91}           & \cellcolor{white!91.21!white}{91.21}          & \cellcolor{white!91.52!white}{91.52}           \\
TDC                                           & \cellcolor{white!96.00!white}{96.00}             & \cellcolor{white!86.00!white}{86.00}           & \cellcolor{white!62.00!white}{62.00}           & \cellcolor{white!88.00!white}{88.00}          & \cellcolor{white!84.00!white}{84.00}           \\ \bottomrule
\end{tabular}
}

\begin{tablenotes}[flushleft]
\small
\item[] \textdagger: \sys is trained on the training set of BPI and test on the test sets of BPI and other unseen datasets.
\item[] \textdaggerdbl: Dataset alias: BEA\&PIQA (\underline{BPI}), HarmBench (\underline{HAB}), SimpleSafetyTests (\underline{SST}), MaliciousInstructions (\underline{MAI}), JADE (\underline{JAD}), HExPHI (\underline{HEP}), TDCRedTeaming (\underline{TDC}).

\end{tablenotes}

\end{threeparttable}}

\end{table}

\begin{table*}[tt]\centering

\setlength{\abovecaptionskip}{0pt}%
\setlength{\belowcaptionskip}{0pt}%
\caption{Accuracy robustness against LLM-targeted static jailbreaking ($IO$-Moderation).}  \label{tab:jailbreak}

\resizebox{\linewidth}{!}{
\begin{threeparttable}

\setlength{\tabcolsep}{3mm}{  
\begin{tabular}{@{}l|cccccccccc@{}}
\toprule
\multicolumn{1}{l|}{\multirow{2}{*}{Jailbreaking Type}}                                                            
                                    & \multicolumn{2}{c|}{ChatGLM3}               & \multicolumn{2}{c|}{LLaMA2}               & \multicolumn{2}{c|}{Falcon}                 & \multicolumn{2}{c|}{Dolly}                  & \multicolumn{2}{c}{Vicuna} \\ 
\multicolumn{1}{c|}{}                                            & \multicolumn{1}{c}{ACC} & \multicolumn{1}{c|}{AUC}  & \multicolumn{1}{c}{ACC} & \multicolumn{1}{c|}{AUC}  & \multicolumn{1}{c}{ACC} & \multicolumn{1}{c|}{AUC}  & \multicolumn{1}{c}{ACC} & \multicolumn{1}{c|}{AUC}   & \multicolumn{1}{c}{ACC}   & AUC    \\ \midrule
Pretending                                                       & 99.529                    & \multicolumn{1}{c|}{0.9997} & 99.860                     & \multicolumn{1}{c|}{1.0000} & 99.557                    & \multicolumn{1}{c|}{0.9998} & 98.327                    & \multicolumn{1}{c|}{0.9978}  & 99.506                      & 0.9997   \\
Attention Shifting                                               & 99.515                    & \multicolumn{1}{c|}{0.9997} & 99.860                    & \multicolumn{1}{c|}{1.0000} & 99.342                   & \multicolumn{1}{c|}{0.9994} & 97.954                    & \multicolumn{1}{c|}{0.9973} & 90.595                      & 0.9961   \\
Privilege Escalation                                             & 99.497                    & \multicolumn{1}{c|}{0.9997} & 99.846                    & \multicolumn{1}{c|}{1.0000} & 99.393                    & \multicolumn{1}{c|}{0.9996} & 94.901                    & \multicolumn{1}{c|}{0.9947}  & 97.991                      & 0.9986   \\
Syntactic Transformation                                       & 93.298                    & \multicolumn{1}{c|}{0.9952} & 89.575                    & \multicolumn{1}{c|}{0.9971} & 97.338                    & \multicolumn{1}{c|}{0.9973} & 84.937                    & \multicolumn{1}{c|}{0.9776}  & 87.221                      & 0.9860   \\ \midrule
Overall                                       &  98.192                   & \multicolumn{1}{c|}{0.9987} &    97.339                 & \multicolumn{1}{c|}{0.9993} &    98.768                & \multicolumn{1}{c|}{0.9987} &      93.983               & \multicolumn{1}{c|}{0.9918}  &     93.872                  & 0.9953   \\\bottomrule
\end{tabular}    
}  



\end{threeparttable}}

\end{table*}
\begin{table*}[tt]
\centering

\setlength{\abovecaptionskip}{0pt}%
\setlength{\belowcaptionskip}{0pt}%
\caption{FPR and FNR robustness against LLM-targeted static jailbreaking ($IO$-Moderation).}  \label{tab:jailbreak-fpr-fnr}

\resizebox{\linewidth}{!}{
\begin{threeparttable}

\setlength{\tabcolsep}{3mm}{  
\begin{tabular}{@{}l|cc|cc|cc|cc|cc@{}}
\toprule
\multicolumn{1}{l|}{\multirow{2}{*}{Jailbreaking Type}}                                                            
                                    & \multicolumn{2}{c|}{ChatGLM3}               & \multicolumn{2}{c|}{LLaMA2}               & \multicolumn{2}{c|}{Falcon}                 & \multicolumn{2}{c|}{Dolly}                  & \multicolumn{2}{c}{Vicuna} \\ 
\multicolumn{1}{c|}{}                                            & \multicolumn{1}{c}{FPR} & \multicolumn{1}{c|}{FNR}  & \multicolumn{1}{c}{FPR} & \multicolumn{1}{c|}{FNR}  & \multicolumn{1}{c}{FPR} & \multicolumn{1}{c|}{FNR}  & \multicolumn{1}{c}{FPR} & \multicolumn{1}{c|}{FNR}   & \multicolumn{1}{c}{FPR}   & FNR    \\ \midrule
Pretending                                                      & 0.471&0.099 & 0.481&0.000 & 1.166&0.072 & 2.009&2.713 & 0.413&0.335 \\ 
Attention Shifting                                               & 0.471&0.072 & 0.481&0.009 & 1.369&0.226 & 1.999&1.936 & 0.413&16.896 \\
Privilege Escalation                                             & 0.471&0.072 & 0.481&0.000 & 1.225&0.217 & 1.846&3.654 & 0.413&10.013 \\ 
Syntactic Transformation                                     & 0.471&15.304 & 0.481&13.296 & 1.263&5.048 & 1.999&29.134 & 0.413&22.395 \\  \midrule
Overall            & 0.461&4.016 & 0.346&3.763 & 1.206&1.583 & 1.933&9.235 & 0.413&12.229 \\  \bottomrule
\end{tabular}    
}  



\end{threeparttable}}

\end{table*}

\subsubsection{Efficiency}
We measure the additional overhead of moderating all samples in the test set of Jigsaw and present the average time per query for eight methods in Table~\ref{tab:baseline}. \sys only requires 0.003 milliseconds per query, which is over $10,040\times$ faster than the other methods. This is because \sys only introduces the overhead of a three-layer MLP classifier, as most of the computation for \sys is completed in the original inference process of the host LLM.

To further examine the time complexity of the eight methods, we vary the length of inputs to be moderated and plot the corresponding computation times for OpenAI Moderation, Perspective API, BeaverDam-7B, and \sys in Figure~\ref{fig:time} in Appendix~\ref{ap:time}. The slopes of the fitted lines for the four methods are 2e-2, 3e-3, 4e-4, and -1e-8, respectively. Observing the trend of the computation times, we find that \sys exhibits a constant complexity of $O(1)$, while the computation times of the other methods increase as inputs lengthen. We do not plot the computation times of GPT-3.5-Turbo and GPT-4 because their complexity is assuredly not constant.

\subsection{Generalization to Unseen Datasets}\label{sec:shift}
In this part, we evaluate the performance of \sys when it encounters potential distribution shifts. We train \sys on BEA\&PIQA dataset, test it on six unseen datasets mentioned in \S\ref{sec:dataset}, and present the results in Table~\ref{tab:unseendata}. We observe that \sys, when hosted in ChatGLM3, LLaMA2, Dolly and Vicuna, generalizes well to most of the unseen datasets, achieving an accuracy of over 90.00\%, 79.69\%, 82.19\%, 73.75\% respectively. Although \sys in Falcon does not generalize well to HarmBench and TDCRedTeaming, it still demonstrates good generalization to the other four unseen datasets. This suggests that \sys can maintain a satisfactory performance when deployed in real scenarios when distribution shifts exist.

\begin{table*}[tt]\centering

\setlength{\abovecaptionskip}{0pt}%
\setlength{\belowcaptionskip}{0pt}%
\caption{Robustness against LLM-targeted dynamic jailbreaking ($IO$-Moderation).} \label{tab:pair-tap}

\resizebox{\linewidth}{!}{
\begin{threeparttable}

\setlength{\tabcolsep}{3mm}{
\begin{tabular}{@{}c|c|ccc|ccc|ccc@{}}
\toprule
\multirow{3}{*}{Dataset} & \multirow{3}{*}{Attack Success Rate (\%)\tnote{\P}} & \multicolumn{3}{c|}{PAIR $\rightarrow$ Vicuna}                                       & \multicolumn{3}{c|}{PAIR $\rightarrow$ LLaMA2}                                       & \multicolumn{3}{c}{TAP $\rightarrow$ Vicuna}                                        \\ \cmidrule(l){3-11} 
                                  &                                  & \multirow{2}{*}{Bare} & \multicolumn{2}{c|}{Legilimens} & \multirow{2}{*}{Bare} & \multicolumn{2}{c|}{Legilimens} & \multirow{2}{*}{Bare} & \multicolumn{2}{c}{Legilimens} \\
                                  &                                  &                                      & w/o DA\tnote{\textdagger}    & w/ DA\tnote{\textdagger}    &                                      & w/o DA\tnote{\textdagger}   & w/ DA\tnote{\textdagger}    &                                      & w/o DA\tnote{\textdagger}    & w/ DA\tnote{\textdagger}   \\ \midrule
\multirow{2}{*}{AdvBEA\tnote{\textdaggerdbl}}            & Judged by GPT-4          & 93.33                                & 53.33               & 3.33               & 3.33                                 & 0                   & 0                  & 93.33                                & 40.00               &    3.33               \\
                                  & Judged by Human           & 73.33                                & 36.67               & 3.33               & 3.33                                 & 0                   & 0                  & 86.67                               & 36.67               &      3.33             \\\midrule
\multirow{2}{*}{AdvBench\tnote{$*$}}         & Judged by GPT-4           & 100.00                               & 68.00               & 12.00              & 2.00                                 & 0                   & 0                  & 96.00                                & 66.00               &    16.00               \\
                                  & Judged by Human           & 100.00                               & 62.00               & 10.00              & 2.00                                 & 0                   & 0                  & 94.00                                & 62.00               &  16.00                \\\bottomrule
\end{tabular}
}

\begin{tablenotes}[flushleft]
\small
\item[] \textdagger: DA is short for the data augmentation technique mentioned in \S\ref{sec:aug}. \P: The attack success rate is judged by GPT-4 and verified manually.

\end{tablenotes}

\end{threeparttable}}
\end{table*}

\subsection{Robustness against Adaptive Adversary}\label{sec:adaptive}
In this part, we consider an adaptive adversary who applies various jailbreak techniques to bypass \sys, including four types of LLM-targeted static jailbreaking, two types of LLM-targeted dynamic jailbreaking, and three types of moderator-targeted jailbreaking. We evaluate the robustness of \sys against these adaptive attacks.

\subsubsection{LLM-Targeted Static Jailbreaking}

For LLM-targeted static jailbreaking, we test \sys on samples rewritten into each type of jailbreak templates from BEA-adv\&AG dataset, and present the results in Table~\ref{tab:jailbreak} and Table~\ref{tab:jailbreak-fpr-fnr}. We can observe from Table~\ref{tab:jailbreak} that \sys in five host models maintains a good performance even when encountering four types of static jailbreaking, with an accuracy of 98.192\%, 97.339\%, 98.768\%, 93.983\%, 93.872\%. It is noteworthy that all jailbreak templates are unseen during the training phase for \sys. From Table~\ref{tab:jailbreak}, we notice that semantic transformations, such as pretending, attention shifting, and privilege escalation, have little impact on the performance of \sys. Syntactic transformations cause an accuracy drop of around 10\%, possibly due to the host model's failure to understand the semantics of prompts after transformation. But \sys still maintains an accuracy over 84.937\%, outperforming all baselines. The results confirm the robustness of \sys against static jailbreaking. 

As a reference, we train \sys without data augmentation mentioned in \S\ref{sec:aug} and present the corresponding results on BEA-adv\&AG dataset in Table~\ref{tab:jailbreak-navie} and Table~\ref{tab:jailbreak-navie-fpr-fnr} in Appendix~\ref{ap:Jailbreaking}. Without data augmentation, \sys only achieves an accuracy of 96.808\% (1.4\%$\downarrow$), 94.715\% (2.6\%$\downarrow$), 95.371\% (3.4\%$\downarrow$), 85.599\% (8.4\%$\downarrow$), 89.640\% (4.2\%$\downarrow$). The results validate the effectiveness of our model-based data augmentation technique.

\subsubsection{LLM-Targeted Dynamic Jailbreaking}\label{sec:dynamic}

In this part, we evaluate \sys under much more stringent circumstances, wherein an adaptive adversary dynamically optimizes their prompts in an iterative manner based on responses of the host LLM to evade safety mechanisms. Within this context, LLM-Targeted Dynamic Jailbreaking can be divided into two types. The first type, known as white-box attack, latest studies (\textit{e.g.},~\cite{paulus2024advprompter, guo2024cold}) have achieved high success rates, but they are not applicable to our scenario settings. The second type, black-box attack, encompasses PAIR~\cite{chao2023jailbreaking}, TAP~\cite{mehrotra2023tree}, and IRIS~\cite{ramesh2024gpt}, surpassing template-based attack methods, modifying prompts in interpretable ways to override LLMs’ safety guardrails. Both PAIR and TAP open source the code. Consequently, we reproduce two state-of-the-art dynamic jailbreaking methods, PAIR and TAP, to stress test \sys, with TAP representing an advanced attack improved from PAIR. We launch PAIR against Vicuna and LLaMA2 and launch TAP against Vicuna only, as we observe that it has limited effectiveness against LLaMA2. We utilize two datasets, namely AdvBench and AdvBEA, to provide initial prompts and goals for two jailbreaking methods. PAIR and TAP continually refine the prompts using Vicuna-7B, with a maximum of 60 and 70 attempts, respectively, to achieve the specified goals. We utilize GPT-4 to judge whether each attempt fulfills the goals, and manually verify the judgement. The attack is considered unsuccessful when all attempts fail. The detailed settings of these two attacks are presented in Appendix~\ref{ap:dynamic}. Note that both AdvBench and AdvBEA are unseen for \sys.



As shown in Table~\ref{tab:pair-tap}, PAIR achieves an attack success rate of 93.33\% (73.33\% upon manual verification) on AdvBEA, and 100\% on AdvBench for the bare/unprotected Vicuna. Similarly, TAP also demonstrates a high attack success rate on both datasets. In contrast, these two jailbreaking methods only achieve an attack success rate of 3.33\% on AdvBEA and 12\%$\sim$16\% on AdvBench for \sys-protected Vicuna, marking a significant decrease by 78\%$\sim$90\%. These results validate the robustness of \sys against dynamic jailbreaking. 

We also evaluate the performance of \sys when no data augmentation is applied in the training phase. As depicted in Table~\ref{tab:pair-tap}, \sys without data augmentation reduces the attack success rate of PAIR by 30\%$\sim$53\%, once again confirming the effectiveness of our data augmentation technique.

We notice that PAIR can hardly attack the bare LLaMA2 because elaborate safety alignment mechanisms have been incorporated in the training phase of LLaMA2~\cite{Touvron23Llama2}. \sys reduces the remaining attack success (3.33\%, 2.00\%) to none (0\%).

\subsubsection{Moderator-Targeted Dynamic Jailbreaking}
In this part, we explore an adaptive adversary that employs traditional adversarial example attacks on classifiers within the natural language processing domain. We reproduce two blind and one decision-based adversarial example attack methods, \textit{i.e.}, VIPER~\cite{EgerSRL0MSSG19} (character-level), SCPN~\cite{iyyer-2018-controlled} (sentence-level), and GAN~\cite{zhengli2018iclr} (sentence-level), utilizing an open-source textual adversarial attack toolkit named OpenAttack~\cite{zeng2020openattack}. For the decision-based method, we assume the adversary can deduce the decision of \sys from the responses returned. We use the default parameters for the implementation of the three attacks. As illustrated in Table~\ref{tab:nlpattack} in Appendix~\ref{ap:Jailbreaking}, we launch the three attacks against \sys using 200 samples from the OIG-Safety dataset, and we observe no success case, which confirms the robustness of \sys against traditional adversarial example attacks.


\begin{table*}[tt]\centering

\setlength{\abovecaptionskip}{0pt}%
\setlength{\belowcaptionskip}{0pt}%
\caption{Accuracy performance in few-shot scenarios ($IO$-Moderation).}  \label{tab:transfer-qa}

\resizebox{\linewidth}{!}{
\begin{threeparttable}

\setlength{\tabcolsep}{3mm}{
\begin{tabular}{@{}c|cc|cc|cc|cc|cc@{}}
\toprule
\multirow{2}{*}{\#Training Samples} & \multicolumn{2}{c|}{ChatGLM3} & \multicolumn{2}{c|}{LLaMA2} & \multicolumn{2}{c|}{Falcon} & \multicolumn{2}{c|}{Dolly} & \multicolumn{2}{c}{Vicuna} \\
                                   & ACC           & AUC           & ACC          & AUC          & ACC           & AUC         & ACC          & AUC         & ACC          & AUC         \\ \midrule
100-\textit{shot}                           & 94.808        & 0.9905         & 97.316        & 0.9969       & 93.792         & 0.9876      & 86.522        & 0.9324       & 91.308        & 0.9651      \\
500-\textit{shot}                           & 97.045         & 0.9957        & 98.318        & 0.9983       & 96.460         & 0.9958      & 91.495        & 0.9710       & 94.724        & 0.9850      \\

1,000-\textit{shot}                          & 95.237         & 0.9959        & 96.435        & 0.9984      & 96.325         & 0.9964      & 93.932        & 0.9849       & 95.652        & 0.9897      \\
5,000-\textit{shot}                           & 97.991         & 0.9975         & 99.240        & 0.9996       & 98.242         & 0.9984      & 95.838        & 0.9914       & 97.847        & 0.9965      \\
10,000-\textit{shot}                         & 98.308         & 0.9986         & 99.366        & 0.9998       & 98.643         & 0.9990      & 96.845       & 0.9947       & 98.252        & 0.9983      \\ \bottomrule
\end{tabular}
}



\end{threeparttable}}
\end{table*}
\begin{table*}[tt]
\centering

\setlength{\abovecaptionskip}{0pt}%
\setlength{\belowcaptionskip}{0pt}%
\caption{FPR and FNR performance in few-shot scenarios ($IO$-Moderation).}  \label{tab:transfer-qa-fpr-fnr}

\resizebox{\linewidth}{!}{
\begin{threeparttable}

\setlength{\tabcolsep}{4mm}{
\begin{tabular}{@{}c|cc|cc|cc|cc|cc@{}}
\toprule
\multirow{2}{*}{\#Training Samples} & \multicolumn{2}{c|}{ChatGLM3} & \multicolumn{2}{c|}{LLaMA2} & \multicolumn{2}{c|}{Falcon} & \multicolumn{2}{c|}{Dolly} & \multicolumn{2}{c}{Vicuna} \\
                                   & FPR           & FNR           & FPR          & FNR          & FPR           & FNR         & FPR          & FNR         & FPR          & FNR         \\ \midrule
100-\textit{shot}     & 1.384&12.717 & 4.682&1.908 & 17.225&0.877 & 15.335&15.258 & 7.672&9.252 \\
500-\textit{shot}                & 1.894&4.377 & 2.346 & 1.420 & 2.378 & 4.215 & 6.999&8.827 & 3.547&5.589 \\

1,000-\textit{shot}        & 2.673&2.234 & 1.442&1.438 & 3.071&1.899 & 7.268&5.671 & 2.355&5.128 \\ 
5,000-\textit{shot}   & 1.317 & 2.623 & 0.307 & 1.601 & 1.684 & 1.655 & 4.739 & 4.251 & 1.682 & 2.686 \\  
10,000-\textit{shot}     & 2.163&1.392 & 0.711&0.452 & 1.434&1.212 & 2.692&3.988 & 0.971 & 2.939 \\   \bottomrule
\end{tabular}
}



\end{threeparttable}}
\end{table*}

\subsection{Impact of Hyper-Parameters}
In this part, we examine the impact of hyper-parameters on the performance of \sys, including the number of probed features and the architecture of the classifier. These experiments are carried out for both $O$- and $IO$-moderation.


\subsubsection{The Number of Probed Features}
We vary the number of probed features, \textit{i.e.}, $m$ mentioned in \S\ref{sec:probe} from 1 to 9, and train a three-layer classifier on these features for each host LLM. We present the performance of $O$-moderation in Table~\ref{tab:features}, Table~\ref{tab:features-fpr-fnr} and $IO$-moderation in Table~\ref{tab:features-qa}, Table~\ref{tab:features-qa-fpr-fnr} in Appendix~\ref{experiment}. We can see that the number of probed features $m$ has little impact on the performance of \sys. For example, in $IO$-moderation, different choices of $m$ result in an accuracy change of only 0.372\% at most. This indicates that the probed features from host LLMs are inherently comprehensive, and the fusion of features across different Transformer blocks only brings about a marginal improvement in performance.


\begin{table}[tt]\centering
\setlength{\abovecaptionskip}{0pt}%
\setlength{\belowcaptionskip}{-5pt}%
\caption{Accuracy of Multi-Label classification ($I$-Moderation).}\label{tab:mulitilabel-0.4}
\resizebox{\linewidth}{!}{
\begin{threeparttable}


\setlength{\tabcolsep}{5.5mm}{
\begin{tabular}{@{}l|cc|cc@{}}
\toprule
\multirow{2}{*}{Category} & \multicolumn{2}{c|}{$m=1$\tnote{\textdagger}} & \multicolumn{2}{c}{$m=3$\tnote{\textdagger}} \\
                       & ACC                  & \multicolumn{1}{c|}{AUC}               & ACC                     & AUC               \\ \midrule
{Obscene}               & 80.523               & 0.9362              & 79.595                  & 0.9449           \\
{Identity Attack}          & 93.209               & 0.9878              & 94.365                  & 0.9876           \\
{Insult}              & 90.615               & 0.9685              & 89.922                  & 0.9658             \\
{Threat}                & 85.557               & 0.9752              & 84.808                  & 0.9777             \\
{Sexual Explicit}        & 82.763               & 0.9648              & 81.950                  & 0.9718             \\ \bottomrule
\end{tabular}
}
\begin{tablenotes}[flushleft]
\small
\item[] \textdagger: the number of probed features.

\end{tablenotes}

\end{threeparttable}}
\end{table}
\begin{table}[h]
\centering
\setlength{\abovecaptionskip}{0pt}%
\setlength{\belowcaptionskip}{0pt}%
\caption{FPR and FNR of Multi-Label classification ($I$-Moderation).}\label{tab:mulitilabel-0.4-fpr-fnr}
\resizebox{\linewidth}{!}{
\begin{threeparttable}


\setlength{\tabcolsep}{5.5mm}{
\begin{tabular}{@{}l|cc|cc@{}}
\toprule
\multirow{2}{*}{Category} & \multicolumn{2}{c|}{$m=1$\tnote{\textdagger}} & \multicolumn{2}{c}{$m=3$\tnote{\textdagger}} \\
                       & FPR                  & \multicolumn{1}{c|}{FNR}               & FPR                     & FNR               \\ \midrule
{Obscene}              & 6.711&22.852 & 1.669&37.801 \\ 
{Identity Attack}       & 1.668&12.281 & 1.367&15.017 \\
{Insult}         & 9.169&9.146 & 10.088&9.158 \\
{Threat}           & 2.203&15.770 & 0.700&27.555 \\ 
{Sexual Explicit}    & 0.974&29.748 & 1.512&26.981 \\ \bottomrule
\end{tabular}
}
\begin{tablenotes}[flushleft]
\small
\item[] \textdagger: the number of probed features.

\end{tablenotes}

\end{threeparttable}}
\end{table}

\subsubsection{The Number of Layers in Classifier}
We vary the number of layers for the classifier from 1 to 9, and train the classifiers for each host LLM. In this part, we set $m=1$. We present the performance of $O$-moderation in Table~\ref{tab:layer}, Table~\ref{tab:layer-fpr-fnr} and $IO$-moderation in Table~\ref{tab:layer-qa}, Table~\ref{tab:layer-qa-fpr-fnr} in Appendix~\ref{experiment}. We find that a three-layer classifier achieves the best performance in most cases. These results validate that a lightweight classifier is sufficient for content moderation when the power of the host LLM is harnessed.


\subsection{Few-Shot Scenarios}\label{sec:transfer_eval}
\begin{figure*}[tt]
\centering
\setlength{\abovecaptionskip}{5pt}
\setlength{\belowcaptionskip}{0pt}
\includegraphics[width=7.in, trim=0 20 0 0, clip]{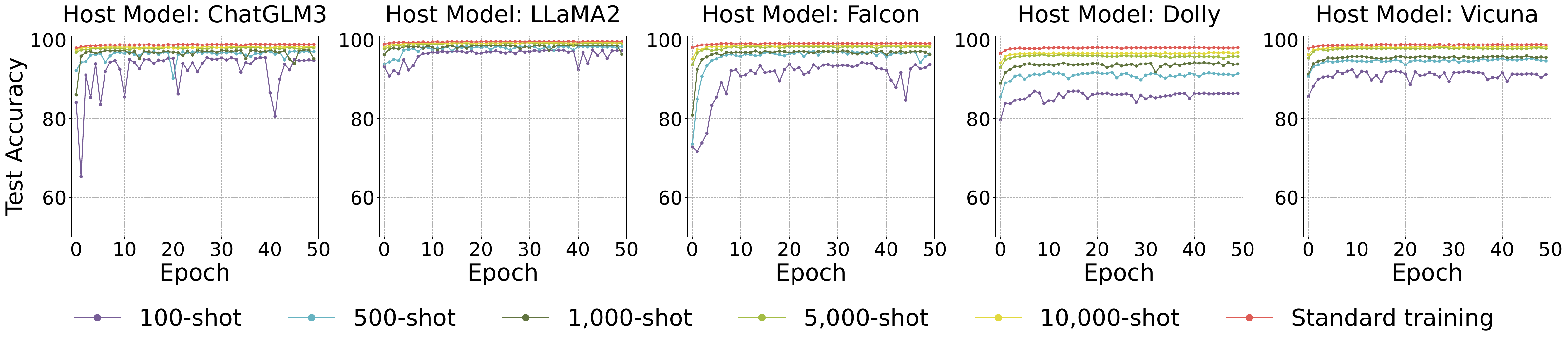}
\caption{Performance in few-shot scenarios ($IO$-Moderation). \sys performs well even when only 1,000 samples are available in the training phase.}
\label{fig:transfer-qa}
\end{figure*}

In this part, we evaluate the applicability of \sys to few-shot scenarios in order to assess the setup cost of \sys. We limit the number of training samples to train \sys to 100, 500, 1,000, 5,000 and 10,000. The results are presented in Table~\ref{tab:transfer-qa} and Table~\ref{tab:transfer-qa-fpr-fnr}. 
\sys achieves a satisfactory accuracy on almost all host LLMs with only 100 samples (0.11\% of the original training set), \textit{i.e.}, 94.808\%, 97.316\%, 93.792\%, 86.522\%, and 91.308\%. With more than 1,000 samples (1.16\% of the original training set), \sys achieves performance closer to standard training, as illustrated in Figure~\ref{fig:transfer-qa}, lowering the burden for service providers in generating training samples.



\subsection{Multi-Label Classification Extension}
In certain scenarios, the classification of unsafe prompts or responses into granular unsafe types is useful, as different moderation strategies may be applied to different types. In this part, we endeavor to extend \sys to a multi-label classification task.  We train \sys on Jigsaw to assign five labels for each sample using five MLPs, determining whether it is related to \textit{obscene}, \textit{identity attack}, \textit{insult}, \textit{threat}, and \textit{sexual explicit}. More specifically, rather than training a multi-label classifier, we train a separate binary classifier for each label.
We calculate the ACC, AUC, FPR and FNR values using the standard methodology for binary classification. The result is shown in Table~\ref{tab:mulitilabel-0.4} and Table~\ref{tab:mulitilabel-0.4-fpr-fnr}. As illustrated in Table~\ref{tab:mulitilabel-0.4}, \sys achieves an accuracy of 80.523\%$\sim$93.209\% when $m=1$, and 79.595\%$\sim$94.365\% when $m=3$. The results confirm that \sys can be extended to multi-label classification tasks. Note that the performance of \sys in classifying certain labels, such as \textit{obscene} and \textit{sexual explicit}, is not as good as in binary classification (an accuracy of 88.39\% as shown in Table~\ref{tab:baseline}), which may be due to the imbalanced training set. For instance, the original ratio of positive to negative samples in \textit{sexual explicit} is 1:7.72, despite our application of re-sampling to mitigate this issue. A more balanced training set may help further improve the performance.




\section{Related Work}

\subsection{Model Alignment}
Model alignment aims to remove the undesired behaviors of trained language models.
Approaches to perform alignment include prompting and reinforcement learning from human feedback (RLHF).

\textit{Prompt-Based Alignment.} Askell~\textit{et~al.}~\cite{Askell2021General} improved alignment and decreased output toxicity by injecting LLMs with helpful, honest, and harmless (HHH) prompts in the form of human-assistant conversations, where the assistant was always polite, helpful, and accurate. Similarly, Rae~\textit{et~al.}~\cite{Rae2011Scaling} also used prompting in order to decrease toxicity (by including ``to be respectful, polite and inclusive''). The problem with this approach is that existing LLMs may not strictly follow the instructions to be aligned.

\textit{RLHF-Based Alignment.}
Bai~\textit{et~al.}~\cite{Bai2022Training} proposed to train LLMs to be helpful and harmless via RLHF. Specifically, they trained LLMs with the assistance of human evaluators in order to optimize their outputs to the evaluator’s preferences. Similarly, Ouyang~\textit{et~al.}~\cite{Ouyang22Training} fine-tuned GPT-3 into InstructGPT using data collected from human labelers to reach better performance on a variety of tasks, while improving alignment. Although RLHF for alignment is effective to a certain extent, it is dangerously brittle. Automatically-generated~\cite{WallaceFKGS19,YuS21} or manually-designed~\cite{XuJLBWD21} adversarial prompts have been shown to effectively bypass existing model alignment, including the alignment effort for ChatGPT~\cite{LiGFXHMS23,Liu2023Jailbreaking,Zhu2023AutoDAN,Yuan2023GPT}. Faced with these empirical results, Wolf~\textit{et~al.}~\cite{Wolf2023Fundamental} proposed a theoretical framework to prove that for any behavior that has a finite probability of being exhibited by the model, there exist prompts that can trigger the model into outputting this behavior. It reveals the fundamental limitations of alignment in LLMs. 

\sys is orthogonal to the alignment-based defenses. When applied simultaneously, \sys has the potential to detect undesired behaviors when the alignment-based defenses are bypassed.

\subsection{Content Moderation}

There is a long track record of work on the detection of unsafe content. According to the classifier applied, we divide this kind of work into two categories, \textit{i.e.}, lightweight classifier and LLM-based classifier.

\textit{Lightweight Classifier.} Many previous works utilized traditional machine learning classifiers for content moderation in social media. For example, Kwok~\textit{et~al.}~\cite{KwokW13}(2013) used a Naive Bayes classifier to distinguish between racist and nonracist tweets. Then, different model architectures are explored, \textit{e.g.}, logistic regression~\cite{NobataTTMC16,DavidsonWMW17,Muhammad2023Detection}, Naive Bayes~\cite{KwokW13,DavidsonWMW17,Muhammad2023Detection}, decision trees~\cite{DavidsonWMW17}, random forests~\cite{DavidsonWMW17,Muhammad2023Detection}, XGBoost~\cite{Muhammad2023Detection}, support vector machines (SVMs)~\cite{DavidsonWMW17}, multilayer perceptron (MLP)~\cite{Muhammad2023Detection} and convolutional neural network (CNN) models~\cite{Muhammad2023Detection}. With the advances of pre-trained language models, BERT and its variants were utilized to more effectively extract feature  for different textual detection tasks. For examples, Dinan~\textit{et~al.}~\cite{DinanHCW19} and Pavlopoulos~\textit{et~al.}~\cite{PavlopoulosSDTA20} used a BERT-base model for offensive language detection. Moon~\textit{et~al.}~\cite{MoonLCJPKMPP23} leveraged a RoBERTa-base model for live-stream chats moderation. Kim~\textit{et~al.}~\cite{Kim2023Robust} employed a DistilBERT model for adversarial prompt detection. Another notable work of this type is by Markov~\textit{et~al.}~\cite{MarkovZANLAJW23}, which presented a holistic approach to building a robust classification system for real-world content moderation for ChatGPT services. Their classifier was based on lightweight GPT. Even with such seemingly comprehensive defenses, Wei~\textit{et~al.}~\cite{Wei2023Jailbroken} revealed that ChatGPT services including GPT-4 were still vulnerable to jailbreak attacks. Their analysis emphasized that safety mechanisms should be as sophisticated as the underlying model. Otherwise, there exist attacks that exploit the cutting-edge capabilities of LLMs while less sophisticated safety classifiers cannot detect.

\sys leverages the powerful feature extraction ability of host models to detect unsafe content, thus addressing the issues of lightweight external classifiers.






\textit{LLM-based Classifier.}
Since it is usually believed that the more complex the classifiers, the more effective they perform, researchers explored LLMs for content moderation. Wang~\textit{et~al.}~\cite{WangHACL23} prompted GPT-3 to generate explanations for hateful and non-hateful speech. Huang~\textit{et~al.}~\cite{Huang2023Harnessing} explored using ChatGPT's proficiency in detecting fake news. Ma~\textit{et~al.}~\cite{Ma2023Adapting} fine-tuned LLMs that can be privately deployed for content moderation.
Cao~\textit{et~al.}~\cite{Cao2023Toxicity} conducted a model review on Hugging Face to reveal the availability of models to cover various moderation rules and guidelines.
Methods of this type are not efficient because extra LLM inference is required for content moderation.

\sys leverages the features extracted during the original inference process of host models for content moderation, thereby introducing minimal overhead.


\section{Discussion \& Future Work}

\textit{More Granular Classification.}
In this paper, we have extended \sys to multi-label classification for five types of unsafe content, yielding satisfactory results in our initial experiments. Given the diversity of unsafe content, a system for more granular classification is useful. Nevertheless, the absence of an agreed-upon taxonomy for unsafe content and the unavailability of high-quality datasets hinder our ability to extend \sys to a more granular classification. We advocate for the prompt establishment of pertinent classification standards for unsafe content, fostering collaborative efforts among all stakeholders to construct high-quality content moderation datasets.


\textit{Larger Host Models.}
\sys achieves excellent performance on five host models of varying architectures, each with 6B to 7B parameters. However, the limitation of computational resources hinders our ability to conduct experiments on larger scale LLMs, \textit{i.e.}, 70B and 175B. Leveraging the capabilities of larger models, we anticipate that \sys will yield even improved results. Investigating the relationship between moderation effectiveness and model size is our future research direction.


\textit{Other Tasks.} \sys is constructed based on conceptual features extracted from host LLMs for content moderation, demonstrating excellent performance. The same feature extraction technique can be employed for other tasks associated with LLMs. For instance, this can include monitoring the level of hallucination and dynamically adjusting the generation of responses to ensure they are safe, faithful, and factual. These potential applications represent intriguing future directions.


\section{Conclusion}


In this paper, we propose a practical and unified content moderation framework for LLM services, named \sys, which features both effectiveness and efficiency. We have conducted extensive experiments on various host LLMs, datasets, and jailbreaking methods to verify the effectiveness, efficiency, and robustness of \sys against normal and adaptive adversaries. The results validate that \sys outperforms both commercial and academic baselines.

\section*{Acknowledgement}
We sincerely thank the anonymous reviewers for their valuable comments and dedication. This work was supported by China NSFC Grant 61925109 and by Ant Group through CCF-Ant Research Fund. Yanjiao Chen is the corresponding author.


\balance

\bibliographystyle{ACM-Reference-Format}
\bibliography{sample-base}


\begin{thebibliography}{96}


\ifx \showCODEN    \undefined \def \showCODEN     #1{\unskip}     \fi
\ifx \showDOI      \undefined \def \showDOI       #1{#1}\fi
\ifx \showISBNx    \undefined \def \showISBNx     #1{\unskip}     \fi
\ifx \showISBNxiii \undefined \def \showISBNxiii  #1{\unskip}     \fi
\ifx \showISSN     \undefined \def \showISSN      #1{\unskip}     \fi
\ifx \showLCCN     \undefined \def \showLCCN      #1{\unskip}     \fi
\ifx \shownote     \undefined \def \shownote      #1{#1}          \fi
\ifx \showarticletitle \undefined \def \showarticletitle #1{#1}   \fi
\ifx \showURL      \undefined \def \showURL       {\relax}        \fi
\providecommand\bibfield[2]{#2}
\providecommand\bibinfo[2]{#2}
\providecommand\natexlab[1]{#1}
\providecommand\showeprint[2][]{arXiv:#2}

\bibitem[Askell et~al\mbox{.}(2021)]%
        {Askell2021General}
\bibfield{author}{\bibinfo{person}{Amanda Askell}, \bibinfo{person}{Yuntao Bai}, \bibinfo{person}{Anna Chen}, \bibinfo{person}{Dawn Drain}, \bibinfo{person}{Deep Ganguli}, \bibinfo{person}{Tom Henighan}, \bibinfo{person}{Andy Jones}, \bibinfo{person}{Nicholas Joseph}, \bibinfo{person}{Benjamin Mann}, \bibinfo{person}{Nova DasSarma}, \bibinfo{person}{Nelson Elhage}, \bibinfo{person}{Zac Hatfield{-}Dodds}, \bibinfo{person}{Danny Hernandez}, \bibinfo{person}{Jackson Kernion}, \bibinfo{person}{Kamal Ndousse}, \bibinfo{person}{Catherine Olsson}, \bibinfo{person}{Dario Amodei}, \bibinfo{person}{Tom~B. Brown}, \bibinfo{person}{Jack Clark}, \bibinfo{person}{Sam McCandlish}, \bibinfo{person}{Chris Olah}, {and} \bibinfo{person}{Jared Kaplan}.} \bibinfo{year}{2021}\natexlab{}.
\newblock \showarticletitle{A General Language Assistant as a Laboratory for Alignment}.
\newblock \bibinfo{journal}{\emph{arXiv preprint arXiv: 2112.00861}} (\bibinfo{year}{2021}).
\newblock


\bibitem[Azaria and Mitchell(2023)]%
        {azaria2023internal}
\bibfield{author}{\bibinfo{person}{Amos Azaria} {and} \bibinfo{person}{Tom Mitchell}.} \bibinfo{year}{2023}\natexlab{}.
\newblock \showarticletitle{The Internal State of an LLM Knows When It's Lying}.
\newblock \bibinfo{journal}{\emph{arXiv preprint arXiv:2304.13734}} (\bibinfo{year}{2023}).
\newblock


\bibitem[Ba et~al\mbox{.}(2016)]%
        {BaKH16}
\bibfield{author}{\bibinfo{person}{Lei~Jimmy Ba}, \bibinfo{person}{Jamie~Ryan Kiros}, {and} \bibinfo{person}{Geoffrey~E. Hinton}.} \bibinfo{year}{2016}\natexlab{}.
\newblock \showarticletitle{Layer Normalization}.
\newblock \bibinfo{journal}{\emph{arXiv preprint arXiv: 1607.06450}} (\bibinfo{year}{2016}).
\newblock


\bibitem[Bai et~al\mbox{.}(2022)]%
        {Bai2022Training}
\bibfield{author}{\bibinfo{person}{Yuntao Bai}, \bibinfo{person}{Andy Jones}, \bibinfo{person}{Kamal Ndousse}, \bibinfo{person}{Amanda Askell}, \bibinfo{person}{Anna Chen}, \bibinfo{person}{Nova DasSarma}, \bibinfo{person}{Dawn Drain}, \bibinfo{person}{Stanislav Fort}, \bibinfo{person}{Deep Ganguli}, \bibinfo{person}{Tom Henighan}, \bibinfo{person}{Nicholas Joseph}, \bibinfo{person}{Saurav Kadavath}, \bibinfo{person}{Jackson Kernion}, \bibinfo{person}{Tom Conerly}, \bibinfo{person}{Sheer~El Showk}, \bibinfo{person}{Nelson Elhage}, \bibinfo{person}{Zac Hatfield{-}Dodds}, \bibinfo{person}{Danny Hernandez}, \bibinfo{person}{Tristan Hume}, \bibinfo{person}{Scott Johnston}, \bibinfo{person}{Shauna Kravec}, \bibinfo{person}{Liane Lovitt}, \bibinfo{person}{Neel Nanda}, \bibinfo{person}{Catherine Olsson}, \bibinfo{person}{Dario Amodei}, \bibinfo{person}{Tom~B. Brown}, \bibinfo{person}{Jack Clark}, \bibinfo{person}{Sam McCandlish}, \bibinfo{person}{Chris Olah}, \bibinfo{person}{Benjamin Mann}, {and}
  \bibinfo{person}{Jared Kaplan}.} \bibinfo{year}{2022}\natexlab{}.
\newblock \showarticletitle{Training a Helpful and Harmless Assistant with Reinforcement Learning from Human Feedback}.
\newblock \bibinfo{journal}{\emph{arXiv preprint arXiv: 2204.0586}} (\bibinfo{year}{2022}).
\newblock


\bibitem[Barrett et~al\mbox{.}(2023)]%
        {Barrett2023Identifying}
\bibfield{author}{\bibinfo{person}{Clark~W. Barrett}, \bibinfo{person}{Brad Boyd}, \bibinfo{person}{Elie Bursztein}, \bibinfo{person}{Nicholas Carlini}, \bibinfo{person}{Brad Chen}, \bibinfo{person}{Jihye Choi}, \bibinfo{person}{Amrita~Roy Chowdhury}, \bibinfo{person}{Mihai Christodorescu}, \bibinfo{person}{Anupam Datta}, \bibinfo{person}{Soheil Feizi}, \bibinfo{person}{Kathleen Fisher}, \bibinfo{person}{Tatsunori Hashimoto}, \bibinfo{person}{Dan Hendrycks}, \bibinfo{person}{Somesh Jha}, \bibinfo{person}{Daniel Kang}, \bibinfo{person}{Florian Kerschbaum}, \bibinfo{person}{Eric Mitchell}, \bibinfo{person}{John~C. Mitchell}, \bibinfo{person}{Zulfikar Ramzan}, \bibinfo{person}{Khawaja Shams}, \bibinfo{person}{Dawn Song}, \bibinfo{person}{Ankur Taly}, {and} \bibinfo{person}{Diyi Yang}.} \bibinfo{year}{2023}\natexlab{}.
\newblock \showarticletitle{Identifying and Mitigating the Security Risks of Generative {AI}}.
\newblock \bibinfo{journal}{\emph{Found. Trends Priv. Secur.}} \bibinfo{volume}{6}, \bibinfo{number}{1} (\bibinfo{year}{2023}), \bibinfo{pages}{1--52}.
\newblock


\bibitem[Bianchi et~al\mbox{.}(2023)]%
        {bianchi2023safety}
\bibfield{author}{\bibinfo{person}{Federico Bianchi}, \bibinfo{person}{Mirac Suzgun}, \bibinfo{person}{Giuseppe Attanasio}, \bibinfo{person}{Paul R{\"o}ttger}, \bibinfo{person}{Dan Jurafsky}, \bibinfo{person}{Tatsunori Hashimoto}, {and} \bibinfo{person}{James Zou}.} \bibinfo{year}{2023}\natexlab{}.
\newblock \showarticletitle{Safety-Tuned Llamas: Lessons from Improving the Safety of Large Language Models that Follow Instructions}.
\newblock \bibinfo{journal}{\emph{arXiv preprint arXiv: 2309.07875}} (\bibinfo{year}{2023}).
\newblock


\bibitem[Biderman et~al\mbox{.}(2023)]%
        {biderman2023pythia}
\bibfield{author}{\bibinfo{person}{Stella Biderman}, \bibinfo{person}{Hailey Schoelkopf}, \bibinfo{person}{Quentin~Gregory Anthony}, \bibinfo{person}{Herbie Bradley}, \bibinfo{person}{Kyle O’Brien}, \bibinfo{person}{Eric Hallahan}, \bibinfo{person}{Mohammad~Aflah Khan}, \bibinfo{person}{Shivanshu Purohit}, \bibinfo{person}{USVSN~Sai Prashanth}, \bibinfo{person}{Edward Raff}, {et~al\mbox{.}}} \bibinfo{year}{2023}\natexlab{}.
\newblock \showarticletitle{Pythia: A Suite for Analyzing Large Language Models across Training and Scaling}. In \bibinfo{booktitle}{\emph{International Conference on Machine Learning}}. PMLR.
\newblock


\bibitem[Bisk et~al\mbox{.}(2020)]%
        {Bisk2020}
\bibfield{author}{\bibinfo{person}{Yonatan Bisk}, \bibinfo{person}{Rowan Zellers}, \bibinfo{person}{Ronan~Le Bras}, \bibinfo{person}{Jianfeng Gao}, {and} \bibinfo{person}{Yejin Choi}.} \bibinfo{year}{2020}\natexlab{}.
\newblock \showarticletitle{PIQA: Reasoning about Physical Commonsense in Natural Language}. In \bibinfo{booktitle}{\emph{AAAI Conference on Artificial Intelligence}}.
\newblock


\bibitem[Brown et~al\mbox{.}(2020a)]%
        {Brown20Language}
\bibfield{author}{\bibinfo{person}{Tom Brown}, \bibinfo{person}{Benjamin Mann}, \bibinfo{person}{Nick Ryder}, \bibinfo{person}{Melanie Subbiah}, \bibinfo{person}{Jared~D Kaplan}, \bibinfo{person}{Prafulla Dhariwal}, \bibinfo{person}{Arvind Neelakantan}, \bibinfo{person}{Pranav Shyam}, \bibinfo{person}{Girish Sastry}, \bibinfo{person}{Amanda Askell}, {et~al\mbox{.}}} \bibinfo{year}{2020}\natexlab{a}.
\newblock \showarticletitle{{Language Models are Few-shot Learners}}. In \bibinfo{booktitle}{\emph{Conference on Neural Information Processing Systems}}. {PMLR}.
\newblock


\bibitem[Brown et~al\mbox{.}(2020b)]%
        {brown2020language}
\bibfield{author}{\bibinfo{person}{Tom Brown}, \bibinfo{person}{Benjamin Mann}, \bibinfo{person}{Nick Ryder}, \bibinfo{person}{Melanie Subbiah}, \bibinfo{person}{Jared~D Kaplan}, \bibinfo{person}{Prafulla Dhariwal}, \bibinfo{person}{Arvind Neelakantan}, \bibinfo{person}{Pranav Shyam}, \bibinfo{person}{Girish Sastry}, \bibinfo{person}{Amanda Askell}, {et~al\mbox{.}}} \bibinfo{year}{2020}\natexlab{b}.
\newblock \showarticletitle{Language models are few-shot learners}.
\newblock \bibinfo{journal}{\emph{Advances in neural information processing systems}}  \bibinfo{volume}{33} (\bibinfo{year}{2020}), \bibinfo{pages}{1877--1901}.
\newblock


\bibitem[Cao et~al\mbox{.}(2023)]%
        {Cao2023Toxicity}
\bibfield{author}{\bibinfo{person}{Yang~Trista Cao}, \bibinfo{person}{Lovely{-}Frances Domingo}, \bibinfo{person}{Sarah~Ann Gilbert}, \bibinfo{person}{Michelle~L. Mazurek}, \bibinfo{person}{Katie Shilton}, {and} \bibinfo{person}{Hal~Daum{\'{e}} III}.} \bibinfo{year}{2023}\natexlab{}.
\newblock \showarticletitle{Toxicity Detection is {NOT} All You Need: Measuring the Gaps to Supporting Volunteer Content Moderators}.
\newblock \bibinfo{journal}{\emph{arXiv preprint arXiv: 2311.07879}} (\bibinfo{year}{2023}).
\newblock


\bibitem[Chao et~al\mbox{.}(2023)]%
        {chao2023jailbreaking}
\bibfield{author}{\bibinfo{person}{Patrick Chao}, \bibinfo{person}{Alexander Robey}, \bibinfo{person}{Edgar Dobriban}, \bibinfo{person}{Hamed Hassani}, \bibinfo{person}{George~J Pappas}, {and} \bibinfo{person}{Eric Wong}.} \bibinfo{year}{2023}\natexlab{}.
\newblock \showarticletitle{Jailbreaking Black Box Large Language Models in Twenty Queries}.
\newblock \bibinfo{journal}{\emph{arXiv preprint arXiv: 2310.08419}} (\bibinfo{year}{2023}).
\newblock


\bibitem[Chen et~al\mbox{.}(2024)]%
        {chen2024inside}
\bibfield{author}{\bibinfo{person}{Chao Chen}, \bibinfo{person}{Kai Liu}, \bibinfo{person}{Ze Chen}, \bibinfo{person}{Yi Gu}, \bibinfo{person}{Yue Wu}, \bibinfo{person}{Mingyuan Tao}, \bibinfo{person}{Zhihang Fu}, {and} \bibinfo{person}{Jieping Ye}.} \bibinfo{year}{2024}\natexlab{}.
\newblock \showarticletitle{INSIDE: LLMs' Internal States Retain the Power of Hallucination Detection}.
\newblock \bibinfo{journal}{\emph{arXiv preprint arXiv:2402.03744}} (\bibinfo{year}{2024}).
\newblock


\bibitem[Child et~al\mbox{.}(2019)]%
        {child2019generating}
\bibfield{author}{\bibinfo{person}{Rewon Child}, \bibinfo{person}{Scott Gray}, \bibinfo{person}{Alec Radford}, {and} \bibinfo{person}{Ilya Sutskever}.} \bibinfo{year}{2019}\natexlab{}.
\newblock \showarticletitle{Generating Long Sequences with Sparse Transformers}.
\newblock \bibinfo{journal}{\emph{arXiv preprint arXiv: 1904.10509}} (\bibinfo{year}{2019}).
\newblock


\bibitem[Chu et~al\mbox{.}(2024)]%
        {Chu2024Comprehensive}
\bibfield{author}{\bibinfo{person}{Junjie Chu}, \bibinfo{person}{Yugeng Liu}, \bibinfo{person}{Ziqing Yang}, \bibinfo{person}{Xinyue Shen}, \bibinfo{person}{Michael Backes}, {and} \bibinfo{person}{Yang Zhang}.} \bibinfo{year}{2024}\natexlab{}.
\newblock \showarticletitle{Comprehensive Assessment of Jailbreak Attacks Against LLMs}.
\newblock \bibinfo{journal}{\emph{arXiv preprint arXiv: 2402.05668}} (\bibinfo{year}{2024}).
\newblock


\bibitem[Chung et~al\mbox{.}(2022)]%
        {Chung2022Scaling}
\bibfield{author}{\bibinfo{person}{Hyung~Won Chung}, \bibinfo{person}{Le Hou}, \bibinfo{person}{Shayne Longpre}, \bibinfo{person}{Barret Zoph}, \bibinfo{person}{Yi Tay}, \bibinfo{person}{William Fedus}, \bibinfo{person}{Eric Li}, \bibinfo{person}{Xuezhi Wang}, \bibinfo{person}{Mostafa Dehghani}, \bibinfo{person}{Siddhartha Brahma}, \bibinfo{person}{Albert Webson}, \bibinfo{person}{Shixiang~Shane Gu}, \bibinfo{person}{Zhuyun Dai}, \bibinfo{person}{Mirac Suzgun}, \bibinfo{person}{Xinyun Chen}, \bibinfo{person}{Aakanksha Chowdhery}, \bibinfo{person}{Sharan Narang}, \bibinfo{person}{Gaurav Mishra}, \bibinfo{person}{Adams Yu}, \bibinfo{person}{Vincent~Y. Zhao}, \bibinfo{person}{Yanping Huang}, \bibinfo{person}{Andrew~M. Dai}, \bibinfo{person}{Hongkun Yu}, \bibinfo{person}{Slav Petrov}, \bibinfo{person}{Ed~H. Chi}, \bibinfo{person}{Jeff Dean}, \bibinfo{person}{Jacob Devlin}, \bibinfo{person}{Adam Roberts}, \bibinfo{person}{Denny Zhou}, \bibinfo{person}{Quoc~V. Le}, {and} \bibinfo{person}{Jason Wei}.}
  \bibinfo{year}{2022}\natexlab{}.
\newblock \showarticletitle{Scaling Instruction-Finetuned Language Models}.
\newblock \bibinfo{journal}{\emph{arXiv preprint arXiv: 2210.11416}} (\bibinfo{year}{2022}).
\newblock


\bibitem[Conover et~al\mbox{.}(2023a)]%
        {DatabricksBlog2023DollyV1}
\bibfield{author}{\bibinfo{person}{Mike Conover}, \bibinfo{person}{Matt Hayes}, \bibinfo{person}{Ankit Mathur}, \bibinfo{person}{Xiangrui Meng}, \bibinfo{person}{Jianwei Xie}, \bibinfo{person}{Jun Wan}, \bibinfo{person}{Ali Ghodsi}, \bibinfo{person}{Patrick Wendell}, {and} \bibinfo{person}{Matei Zaharia}.} \bibinfo{year}{2023}\natexlab{a}.
\newblock \bibinfo{title}{Hello Dolly: Democratizing the Magic of ChatGPT with Open Models}.
\newblock \bibinfo{howpublished}{\url{https://www.databricks.com/blog/2023/03/24/hello-dolly-democratizing-magic-chatgpt-open-models.html}}.
\newblock


\bibitem[Conover et~al\mbox{.}(2023b)]%
        {DatabricksBlog2023DollyV2}
\bibfield{author}{\bibinfo{person}{Mike Conover}, \bibinfo{person}{Matt Hayes}, \bibinfo{person}{Ankit Mathur}, \bibinfo{person}{Jianwei Xie}, \bibinfo{person}{Jun Wan}, \bibinfo{person}{Sam Shah}, \bibinfo{person}{Ali Ghodsi}, \bibinfo{person}{Patrick Wendell}, \bibinfo{person}{Matei Zaharia}, {and} \bibinfo{person}{Reynold Xin}.} \bibinfo{year}{2023}\natexlab{b}.
\newblock \bibinfo{title}{Free Dolly: Introducing the World's First Truly Open Instruction-Tuned LLM}.
\newblock \bibinfo{howpublished}{\url{https://www.databricks.com/blog/2023/04/12/dolly-first-open-commercially-viable-instruction-tuned-llm}}.
\newblock


\bibitem[Cui et~al\mbox{.}(2024)]%
        {Cui2024Risk}
\bibfield{author}{\bibinfo{person}{Tianyu Cui}, \bibinfo{person}{Yanling Wang}, \bibinfo{person}{Chuanpu Fu}, \bibinfo{person}{Yong Xiao}, \bibinfo{person}{Sijia Li}, \bibinfo{person}{Xinhao Deng}, \bibinfo{person}{Yunpeng Liu}, \bibinfo{person}{Qinglin Zhang}, \bibinfo{person}{Ziyi Qiu}, \bibinfo{person}{Peiyang Li}, \bibinfo{person}{Zhixing Tan}, \bibinfo{person}{Junwu Xiong}, \bibinfo{person}{Xinyu Kong}, \bibinfo{person}{Zujie Wen}, \bibinfo{person}{Ke Xu}, {and} \bibinfo{person}{Qi Li}.} \bibinfo{year}{2024}\natexlab{}.
\newblock \showarticletitle{Risk Taxonomy, Mitigation, and Assessment Benchmarks of Large Language Model Systems}.
\newblock \bibinfo{journal}{\emph{arXiv preprint arXiv: 2401.05778}} (\bibinfo{year}{2024}).
\newblock


\bibitem[Dauphin et~al\mbox{.}(2017)]%
        {dauphin2017language}
\bibfield{author}{\bibinfo{person}{Yann~N Dauphin}, \bibinfo{person}{Angela Fan}, \bibinfo{person}{Michael Auli}, {and} \bibinfo{person}{David Grangier}.} \bibinfo{year}{2017}\natexlab{}.
\newblock \showarticletitle{Language Modeling with Gated Convolutional Networks}. In \bibinfo{booktitle}{\emph{International conference on machine learning}}. PMLR.
\newblock


\bibitem[Davidson et~al\mbox{.}(2017)]%
        {DavidsonWMW17}
\bibfield{author}{\bibinfo{person}{Thomas Davidson}, \bibinfo{person}{Dana Warmsley}, \bibinfo{person}{Michael~W. Macy}, {and} \bibinfo{person}{Ingmar Weber}.} \bibinfo{year}{2017}\natexlab{}.
\newblock \showarticletitle{Automated Hate Speech Detection and the Problem of Offensive Language}. In \bibinfo{booktitle}{\emph{International Conference on Web and Social Media}}. \bibinfo{publisher}{{AAAI}}.
\newblock


\bibitem[Dinan et~al\mbox{.}(2019)]%
        {DinanHCW19}
\bibfield{author}{\bibinfo{person}{Emily Dinan}, \bibinfo{person}{Samuel Humeau}, \bibinfo{person}{Bharath Chintagunta}, {and} \bibinfo{person}{Jason Weston}.} \bibinfo{year}{2019}\natexlab{}.
\newblock \showarticletitle{Build it Break it Fix it for Dialogue Safety: Robustness from Adversarial Human Attack}. In \bibinfo{booktitle}{\emph{Conference on Empirical Methods in Natural Language Processing and International Joint Conference on Natural Language Processing}}. \bibinfo{publisher}{Association for Computational Linguistics}.
\newblock


\bibitem[Dong et~al\mbox{.}(2019)]%
        {Dong2019Unified}
\bibfield{author}{\bibinfo{person}{Li Dong}, \bibinfo{person}{Nan Yang}, \bibinfo{person}{Wenhui Wang}, \bibinfo{person}{Furu Wei}, \bibinfo{person}{Xiaodong Liu}, \bibinfo{person}{Yu Wang}, \bibinfo{person}{Jianfeng Gao}, \bibinfo{person}{Ming Zhou}, {and} \bibinfo{person}{Hsiao{-}Wuen Hon}.} \bibinfo{year}{2019}\natexlab{}.
\newblock \showarticletitle{Unified Language Model Pre-Training for Natural Language Understanding and Generation}. In \bibinfo{booktitle}{\emph{Conference on Neural Information Processing Systems}}. {PMLR}.
\newblock


\bibitem[Du et~al\mbox{.}(2022)]%
        {Du22GLM}
\bibfield{author}{\bibinfo{person}{Zhengxiao Du}, \bibinfo{person}{Yujie Qian}, \bibinfo{person}{Xiao Liu}, \bibinfo{person}{Ming Ding}, \bibinfo{person}{Jiezhong Qiu}, \bibinfo{person}{Zhilin Yang}, {and} \bibinfo{person}{Jie Tang}.} \bibinfo{year}{2022}\natexlab{}.
\newblock \showarticletitle{{GLM: General Language Model Pretraining with Autoregressive Blank Infilling}}. In \bibinfo{booktitle}{\emph{Annual Meeting of the Association for Computational Linguistics}}.
\newblock


\bibitem[Eger et~al\mbox{.}(2019)]%
        {EgerSRL0MSSG19}
\bibfield{author}{\bibinfo{person}{Steffen Eger}, \bibinfo{person}{G{\"{o}}zde~G{\"{u}}l Sahin}, \bibinfo{person}{Andreas R{\"{u}}ckl{\'{e}}}, \bibinfo{person}{Ji{-}Ung Lee}, \bibinfo{person}{Claudia Schulz}, \bibinfo{person}{Mohsen Mesgar}, \bibinfo{person}{Krishnkant Swarnkar}, \bibinfo{person}{Edwin Simpson}, {and} \bibinfo{person}{Iryna Gurevych}.} \bibinfo{year}{2019}\natexlab{}.
\newblock \showarticletitle{Text Processing Like Humans Do: Visually Attacking and Shielding {NLP} Systems}. In \bibinfo{booktitle}{\emph{Conference of the North American Chapter of the Association for Computational Linguistics: Human Language Technologies}}. \bibinfo{publisher}{Association for Computational Linguistics}.
\newblock


\bibitem[Fortuna and Nunes(2018)]%
        {FortunaN18}
\bibfield{author}{\bibinfo{person}{Paula Fortuna} {and} \bibinfo{person}{S{\'{e}}rgio Nunes}.} \bibinfo{year}{2018}\natexlab{}.
\newblock \showarticletitle{A Survey on Automatic Detection of Hate Speech in Text}.
\newblock \bibinfo{journal}{\emph{{ACM} Comput. Surv.}} \bibinfo{volume}{51}, \bibinfo{number}{4} (\bibinfo{year}{2018}), \bibinfo{pages}{85:1--85:30}.
\newblock


\bibitem[Ganguli et~al\mbox{.}(2022)]%
        {ganguli2022red}
\bibfield{author}{\bibinfo{person}{Deep Ganguli}, \bibinfo{person}{Liane Lovitt}, \bibinfo{person}{Jackson Kernion}, \bibinfo{person}{Amanda Askell}, \bibinfo{person}{Yuntao Bai}, \bibinfo{person}{Saurav Kadavath}, \bibinfo{person}{Ben Mann}, \bibinfo{person}{Ethan Perez}, \bibinfo{person}{Nicholas Schiefer}, \bibinfo{person}{Kamal Ndousse}, {et~al\mbox{.}}} \bibinfo{year}{2022}\natexlab{}.
\newblock \showarticletitle{Red Teaming Language Models to Reduce Harms: Methods, Scaling Behaviors, and Lessons Learned}.
\newblock \bibinfo{journal}{\emph{arXiv preprint arXiv: 2209.07858}} (\bibinfo{year}{2022}).
\newblock


\bibitem[{Google Jigsaw}(2017)]%
        {JigsawDataset}
\bibfield{author}{\bibinfo{person}{{Google Jigsaw}}.} \bibinfo{year}{2017}\natexlab{}.
\newblock \bibinfo{title}{Jigsaw Unintended Bias in Toxicity Classification}.
\newblock \bibinfo{howpublished}{\url{https://www.kaggle.com/c/jigsaw-unintended-bias-in-toxicity-classification}}.
\newblock


\bibitem[Gorwa et~al\mbox{.}(2020)]%
        {GorwaBK20}
\bibfield{author}{\bibinfo{person}{Robert Gorwa}, \bibinfo{person}{Reuben Binns}, {and} \bibinfo{person}{Christian Katzenbach}.} \bibinfo{year}{2020}\natexlab{}.
\newblock \showarticletitle{Algorithmic Content Moderation: Technical and Political Challenges in the Automation of Platform Governance}.
\newblock \bibinfo{journal}{\emph{Big Data Soc.}} \bibinfo{volume}{7}, \bibinfo{number}{1} (\bibinfo{year}{2020}), \bibinfo{pages}{205395171989794}.
\newblock


\bibitem[Griffin(2023)]%
        {Griffin2023ChatGPT}
\bibfield{author}{\bibinfo{person}{Andrew Griffin}.} \bibinfo{year}{2023}\natexlab{}.
\newblock \bibinfo{title}{{ChatGPT Plus}: OpenAI Stops Premium Signups after Major Update}.
\newblock \bibinfo{howpublished}{\url{https://www.independent.co.uk/tech/chatgpt-plus-free-premium-paid-subscription-sign-up-b2447941.html}}.
\newblock


\bibitem[Guo et~al\mbox{.}(2024)]%
        {guo2024cold}
\bibfield{author}{\bibinfo{person}{Xingang Guo}, \bibinfo{person}{Fangxu Yu}, \bibinfo{person}{Huan Zhang}, \bibinfo{person}{Lianhui Qin}, {and} \bibinfo{person}{Bin Hu}.} \bibinfo{year}{2024}\natexlab{}.
\newblock \showarticletitle{Cold-attack: Jailbreaking llms with stealthiness and controllability}.
\newblock \bibinfo{journal}{\emph{arXiv preprint arXiv:2402.08679}} (\bibinfo{year}{2024}).
\newblock


\bibitem[He et~al\mbox{.}(2016)]%
        {HeZRS16}
\bibfield{author}{\bibinfo{person}{Kaiming He}, \bibinfo{person}{Xiangyu Zhang}, \bibinfo{person}{Shaoqing Ren}, {and} \bibinfo{person}{Jian Sun}.} \bibinfo{year}{2016}\natexlab{}.
\newblock \showarticletitle{Deep Residual Learning for Image Recognition}. In \bibinfo{booktitle}{\emph{{IEEE} Conference on Computer Vision and Pattern Recognition}}. \bibinfo{publisher}{{IEEE} Computer Society}.
\newblock


\bibitem[Huang and Sun(2023)]%
        {Huang2023Harnessing}
\bibfield{author}{\bibinfo{person}{Yue Huang} {and} \bibinfo{person}{Lichao Sun}.} \bibinfo{year}{2023}\natexlab{}.
\newblock \showarticletitle{Harnessing the Power of ChatGPT in Fake News: An In-Depth Exploration in Generation, Detection and Explanation}.
\newblock \bibinfo{journal}{\emph{arXiv preprint arXiv: 2310.05046}} (\bibinfo{year}{2023}).
\newblock


\bibitem[Inan et~al\mbox{.}(2023)]%
        {DBLP:journals/corr/abs-2312-06674}
\bibfield{author}{\bibinfo{person}{Hakan Inan}, \bibinfo{person}{Kartikeya Upasani}, \bibinfo{person}{Jianfeng Chi}, \bibinfo{person}{Rashi Rungta}, \bibinfo{person}{Krithika Iyer}, \bibinfo{person}{Yuning Mao}, \bibinfo{person}{Michael Tontchev}, \bibinfo{person}{Qing Hu}, \bibinfo{person}{Brian Fuller}, \bibinfo{person}{Davide Testuggine}, {and} \bibinfo{person}{Madian Khabsa}.} \bibinfo{year}{2023}\natexlab{}.
\newblock \showarticletitle{Llama Guard: LLM-based Input-Output Safeguard for Human-AI Conversations}.
\newblock \bibinfo{journal}{\emph{CoRR}}  \bibinfo{volume}{abs/2312.06674} (\bibinfo{year}{2023}).
\newblock
\urldef\tempurl%
\url{https://doi.org/10.48550/ARXIV.2312.06674}
\showDOI{\tempurl}
\showeprint[arXiv]{2312.06674}


\bibitem[Iyyer et~al\mbox{.}(2018)]%
        {iyyer-2018-controlled}
\bibfield{author}{\bibinfo{person}{Mohit Iyyer}, \bibinfo{person}{John Wieting}, \bibinfo{person}{Kevin Gimpel}, {and} \bibinfo{person}{Luke Zettlemoyer}.} \bibinfo{year}{2018}\natexlab{}.
\newblock \showarticletitle{Adversarial Example Generation with Syntactically Controlled Paraphrase Networks}. In \bibinfo{booktitle}{\emph{Annual Conference of the North American Chapter of the Association for Computational Linguistics}}.
\newblock


\bibitem[Ji et~al\mbox{.}(2023)]%
        {Ji2023BeaverTails}
\bibfield{author}{\bibinfo{person}{Jiaming Ji}, \bibinfo{person}{Mickel Liu}, \bibinfo{person}{Josef Dai}, \bibinfo{person}{Xuehai Pan}, \bibinfo{person}{Chi Zhang}, \bibinfo{person}{Ce Bian}, \bibinfo{person}{Boyuan Chen}, \bibinfo{person}{Ruiyang Sun}, \bibinfo{person}{Yizhou Wang}, {and} \bibinfo{person}{Yaodong Yang}.} \bibinfo{year}{2023}\natexlab{}.
\newblock \showarticletitle{BeaverTails: Towards Improved Safety Alignment of {LLM} via a Human-Preference Dataset}. In \bibinfo{booktitle}{\emph{Conference on Neural Information Processing Systems}}. {PMLR}.
\newblock


\bibitem[Jigsaw(2017)]%
        {PerspectiveAPI}
\bibfield{author}{\bibinfo{person}{Google Jigsaw}.} \bibinfo{year}{2017}\natexlab{}.
\newblock \bibinfo{title}{Perspective {API}}.
\newblock \bibinfo{howpublished}{\url{https://www.perspectiveapi.com/}}.
\newblock


\bibitem[Kennedy et~al\mbox{.}(2020)]%
        {kennedy2020constructing}
\bibfield{author}{\bibinfo{person}{Chris~J Kennedy}, \bibinfo{person}{Geoff Bacon}, \bibinfo{person}{Alexander Sahn}, {and} \bibinfo{person}{Claudia von Vacano}.} \bibinfo{year}{2020}\natexlab{}.
\newblock \showarticletitle{Constructing Interval Variables via Faceted Rasch Measurement and Multitask Deep Learning: a Hate Speech Application}.
\newblock \bibinfo{journal}{\emph{arXiv preprint arXiv: 2009.10277}} (\bibinfo{year}{2020}).
\newblock


\bibitem[Kim et~al\mbox{.}(2023)]%
        {Kim2023Robust}
\bibfield{author}{\bibinfo{person}{Jinhwa Kim}, \bibinfo{person}{Ali Derakhshan}, {and} \bibinfo{person}{Ian~G. Harris}.} \bibinfo{year}{2023}\natexlab{}.
\newblock \showarticletitle{Robust Safety Classifier for Large Language Models: Adversarial Prompt Shield}.
\newblock \bibinfo{journal}{\emph{arXiv preprint arXiv: 2311.00172}} (\bibinfo{year}{2023}).
\newblock


\bibitem[King(2023)]%
        {King2023Meet}
\bibfield{author}{\bibinfo{person}{Michael King}.} \bibinfo{year}{2023}\natexlab{}.
\newblock \bibinfo{title}{Meet DAN -- The `JAILBREAK' Version of ChatGPT and How to Use it -- AI Unchained and Unfiltered}.
\newblock \bibinfo{howpublished}{\url{https://platform.openai.com/docs/guides/moderation}}.
\newblock


\bibitem[Kingma and Ba(2015)]%
        {kingma2014adam}
\bibfield{author}{\bibinfo{person}{Diederik~P. Kingma} {and} \bibinfo{person}{Jimmy Ba}.} \bibinfo{year}{2015}\natexlab{}.
\newblock \showarticletitle{Adam: {A} Method for Stochastic Optimization}. In \bibinfo{booktitle}{\emph{International Conference on Learning Representations}}. OpenReview.net.
\newblock


\bibitem[Kwok and Wang(2013)]%
        {KwokW13}
\bibfield{author}{\bibinfo{person}{Irene Kwok} {and} \bibinfo{person}{Yuzhou Wang}.} \bibinfo{year}{2013}\natexlab{}.
\newblock \showarticletitle{Locate the Hate: Detecting Tweets against Blacks}. In \bibinfo{booktitle}{\emph{{AAAI} Conference on Artificial Intelligence}}.
\newblock


\bibitem[Li et~al\mbox{.}(2023)]%
        {LiGFXHMS23}
\bibfield{author}{\bibinfo{person}{Haoran Li}, \bibinfo{person}{Dadi Guo}, \bibinfo{person}{Wei Fan}, \bibinfo{person}{Mingshi Xu}, \bibinfo{person}{Jie Huang}, \bibinfo{person}{Fanpu Meng}, {and} \bibinfo{person}{Yangqiu Song}.} \bibinfo{year}{2023}\natexlab{}.
\newblock \showarticletitle{Multi-Step Jailbreaking Privacy Attacks on ChatGPT}. In \bibinfo{booktitle}{\emph{Findings of the Association for Computational Linguistics: {EMNLP}}}. \bibinfo{publisher}{Association for Computational Linguistics}.
\newblock


\bibitem[Lin et~al\mbox{.}(2021)]%
        {Lin2021Survey}
\bibfield{author}{\bibinfo{person}{Tianyang Lin}, \bibinfo{person}{Yuxin Wang}, \bibinfo{person}{Xiangyang Liu}, {and} \bibinfo{person}{Xipeng Qiu}.} \bibinfo{year}{2021}\natexlab{}.
\newblock \showarticletitle{A Survey of Transformers}.
\newblock \bibinfo{journal}{\emph{arXiv preprint arXiv: 2106.04554}} (\bibinfo{year}{2021}).
\newblock


\bibitem[Liu et~al\mbox{.}(2023)]%
        {Liu2023Jailbreaking}
\bibfield{author}{\bibinfo{person}{Yi Liu}, \bibinfo{person}{Gelei Deng}, \bibinfo{person}{Zhengzi Xu}, \bibinfo{person}{Yuekang Li}, \bibinfo{person}{Yaowen Zheng}, \bibinfo{person}{Ying Zhang}, \bibinfo{person}{Lida Zhao}, \bibinfo{person}{Tianwei Zhang}, {and} \bibinfo{person}{Yang Liu}.} \bibinfo{year}{2023}\natexlab{}.
\newblock \showarticletitle{Jailbreaking ChatGPT via Prompt Engineering: An Empirical Study}.
\newblock \bibinfo{journal}{\emph{arXiv preprint arXiv: 2305.13860}} (\bibinfo{year}{2023}).
\newblock


\bibitem[Ma et~al\mbox{.}(2023)]%
        {Ma2023Adapting}
\bibfield{author}{\bibinfo{person}{Huan Ma}, \bibinfo{person}{Changqing Zhang}, \bibinfo{person}{Huazhu Fu}, \bibinfo{person}{Peilin Zhao}, {and} \bibinfo{person}{Bingzhe Wu}.} \bibinfo{year}{2023}\natexlab{}.
\newblock \showarticletitle{Adapting Large Language Models for Content Moderation: Pitfalls in Data Engineering and Supervised Fine-Tuning}.
\newblock \bibinfo{journal}{\emph{arXiv preprint arXiv: 2310.03400}} (\bibinfo{year}{2023}).
\newblock


\bibitem[Markov et~al\mbox{.}(2023)]%
        {MarkovZANLAJW23}
\bibfield{author}{\bibinfo{person}{Todor Markov}, \bibinfo{person}{Chong Zhang}, \bibinfo{person}{Sandhini Agarwal}, \bibinfo{person}{Florentine~Eloundou Nekoul}, \bibinfo{person}{Theodore Lee}, \bibinfo{person}{Steven Adler}, \bibinfo{person}{Angela Jiang}, {and} \bibinfo{person}{Lilian Weng}.} \bibinfo{year}{2023}\natexlab{}.
\newblock \showarticletitle{A Holistic Approach to Undesired Content Detection in the Real World}. In \bibinfo{booktitle}{\emph{{AAAI} Conference on Artificial Intelligence}}.
\newblock


\bibitem[Mathew et~al\mbox{.}(2021)]%
        {mathew2021hatexplain}
\bibfield{author}{\bibinfo{person}{Binny Mathew}, \bibinfo{person}{Punyajoy Saha}, \bibinfo{person}{Seid~Muhie Yimam}, \bibinfo{person}{Chris Biemann}, \bibinfo{person}{Pawan Goyal}, {and} \bibinfo{person}{Animesh Mukherjee}.} \bibinfo{year}{2021}\natexlab{}.
\newblock \showarticletitle{HateXplain: A Benchmark Dataset for Explainable Hate Speech Detection}. In \bibinfo{booktitle}{\emph{AAAI Conference on Artificial Intelligence}}.
\newblock


\bibitem[Mazeika et~al\mbox{.}(2024)]%
        {mazeika2024harmbench}
\bibfield{author}{\bibinfo{person}{Mantas Mazeika}, \bibinfo{person}{Long Phan}, \bibinfo{person}{Xuwang Yin}, \bibinfo{person}{Andy Zou}, \bibinfo{person}{Zifan Wang}, \bibinfo{person}{Norman Mu}, \bibinfo{person}{Elham Sakhaee}, \bibinfo{person}{Nathaniel Li}, \bibinfo{person}{Steven Basart}, \bibinfo{person}{Bo Li}, {et~al\mbox{.}}} \bibinfo{year}{2024}\natexlab{}.
\newblock \showarticletitle{HarmBench: A Standardized Evaluation Framework for Automated Red Teaming and Robust Refusal}.
\newblock \bibinfo{journal}{\emph{arXiv preprint arXiv: 2402.04249}} (\bibinfo{year}{2024}).
\newblock


\bibitem[Mehrotra et~al\mbox{.}(2023)]%
        {mehrotra2023tree}
\bibfield{author}{\bibinfo{person}{Anay Mehrotra}, \bibinfo{person}{Manolis Zampetakis}, \bibinfo{person}{Paul Kassianik}, \bibinfo{person}{Blaine Nelson}, \bibinfo{person}{Hyrum Anderson}, \bibinfo{person}{Yaron Singer}, {and} \bibinfo{person}{Amin Karbasi}.} \bibinfo{year}{2023}\natexlab{}.
\newblock \showarticletitle{Tree of Attacks: Jailbreaking Black-Box LLMs Automatically}.
\newblock \bibinfo{journal}{\emph{arXiv preprint arXiv: 2312.02119}} (\bibinfo{year}{2023}).
\newblock


\bibitem[Moon et~al\mbox{.}(2023)]%
        {MoonLCJPKMPP23}
\bibfield{author}{\bibinfo{person}{Jihyung Moon}, \bibinfo{person}{Dong{-}Ho Lee}, \bibinfo{person}{Hyundong Cho}, \bibinfo{person}{Woojeong Jin}, \bibinfo{person}{Chan~Young Park}, \bibinfo{person}{Minwoo Kim}, \bibinfo{person}{Jonathan May}, \bibinfo{person}{Jay Pujara}, {and} \bibinfo{person}{Sungjoon Park}.} \bibinfo{year}{2023}\natexlab{}.
\newblock \showarticletitle{Analyzing Norm Violations in Live-Stream Chat}. In \bibinfo{booktitle}{\emph{Conference on Empirical Methods in Natural Language Processing}}. \bibinfo{publisher}{Association for Computational Linguistics}.
\newblock


\bibitem[Muhammad et~al\mbox{.}(2023)]%
        {Muhammad2023Detection}
\bibfield{author}{\bibinfo{person}{Fatima~Adam Muhammad}, \bibinfo{person}{Abubakar~Yakubu Zandam}, {and} \bibinfo{person}{Isa Inuwa{-}Dutse}.} \bibinfo{year}{2023}\natexlab{}.
\newblock \showarticletitle{Detection of Offensive and Threatening Online Content in a Low Resource Language}.
\newblock \bibinfo{journal}{\emph{arXiv preprint arXiv: 2311.10541}} (\bibinfo{year}{2023}).
\newblock


\bibitem[Nguyen et~al\mbox{.}(2023)]%
        {Nguyen2023OIG}
\bibfield{author}{\bibinfo{person}{Huu Nguyen}, \bibinfo{person}{Sameer Suri}, \bibinfo{person}{Ken Tsui}, \bibinfo{person}{Shahules786}, \bibinfo{person}{Together.xyz team}, {and} \bibinfo{person}{Christoph Schuhmann}.} \bibinfo{year}{2023}\natexlab{}.
\newblock \bibinfo{title}{The OIG Dataset}.
\newblock \bibinfo{howpublished}{\url{https://laion.ai/blog/oig-dataset/}}.
\newblock


\bibitem[Nobata et~al\mbox{.}(2016)]%
        {NobataTTMC16}
\bibfield{author}{\bibinfo{person}{Chikashi Nobata}, \bibinfo{person}{Joel~R. Tetreault}, \bibinfo{person}{Achint Thomas}, \bibinfo{person}{Yashar Mehdad}, {and} \bibinfo{person}{Yi Chang}.} \bibinfo{year}{2016}\natexlab{}.
\newblock \showarticletitle{Abusive Language Detection in Online User Content}. In \bibinfo{booktitle}{\emph{International Conference on World Wide Web}}. \bibinfo{publisher}{{ACM}}.
\newblock


\bibitem[OpenAI(2023a)]%
        {OpenAI23GPT4}
\bibfield{author}{\bibinfo{person}{OpenAI}.} \bibinfo{year}{2023}\natexlab{a}.
\newblock \showarticletitle{{GPT-4} Technical Report}.
\newblock \bibinfo{journal}{\emph{arXiv preprint arXiv: 2303.08774}} (\bibinfo{year}{2023}).
\newblock


\bibitem[OpenAI(2023b)]%
        {OpenAIModeration}
\bibfield{author}{\bibinfo{person}{OpenAI}.} \bibinfo{year}{2023}\natexlab{b}.
\newblock \bibinfo{title}{Moderation}.
\newblock \bibinfo{howpublished}{\url{https://platform.openai.com/docs/guides/moderation}}.
\newblock


\bibitem[Ouyang et~al\mbox{.}(2022)]%
        {Ouyang22Training}
\bibfield{author}{\bibinfo{person}{Long Ouyang}, \bibinfo{person}{Jeffrey Wu}, \bibinfo{person}{Xu Jiang}, \bibinfo{person}{Diogo Almeida}, \bibinfo{person}{Carroll~L. Wainwright}, \bibinfo{person}{Pamela Mishkin}, \bibinfo{person}{Chong Zhang}, \bibinfo{person}{Sandhini Agarwal}, \bibinfo{person}{Katarina Slama}, \bibinfo{person}{Alex Ray}, \bibinfo{person}{John Schulman}, \bibinfo{person}{Jacob Hilton}, \bibinfo{person}{Fraser Kelton}, \bibinfo{person}{Luke Miller}, \bibinfo{person}{Maddie Simens}, \bibinfo{person}{Amanda Askell}, \bibinfo{person}{Peter Welinder}, \bibinfo{person}{Paul~F. Christiano}, \bibinfo{person}{Jan Leike}, {and} \bibinfo{person}{Ryan Lowe}.} \bibinfo{year}{2022}\natexlab{}.
\newblock \showarticletitle{{Training Language Models to Follow Instructions with Human Feedback}}. In \bibinfo{booktitle}{\emph{Conference on Neural Information Processing Systems}}. {PMLR}.
\newblock


\bibitem[Paszke et~al\mbox{.}(2019)]%
        {PaszkeGMLBCKLGA19}
\bibfield{author}{\bibinfo{person}{Adam Paszke}, \bibinfo{person}{Sam Gross}, \bibinfo{person}{Francisco Massa}, \bibinfo{person}{Adam Lerer}, \bibinfo{person}{James Bradbury}, \bibinfo{person}{Gregory Chanan}, \bibinfo{person}{Trevor Killeen}, \bibinfo{person}{Zeming Lin}, \bibinfo{person}{Natalia Gimelshein}, \bibinfo{person}{Luca Antiga}, \bibinfo{person}{Alban Desmaison}, \bibinfo{person}{Andreas K{\"{o}}pf}, \bibinfo{person}{Edward~Z. Yang}, \bibinfo{person}{Zachary DeVito}, \bibinfo{person}{Martin Raison}, \bibinfo{person}{Alykhan Tejani}, \bibinfo{person}{Sasank Chilamkurthy}, \bibinfo{person}{Benoit Steiner}, \bibinfo{person}{Lu Fang}, \bibinfo{person}{Junjie Bai}, {and} \bibinfo{person}{Soumith Chintala}.} \bibinfo{year}{2019}\natexlab{}.
\newblock \showarticletitle{{PyTorch}: An Imperative Style, High-Performance Deep Learning Library}. In \bibinfo{booktitle}{\emph{Conference on Neural Information Processing Systems}}. {PMLR}.
\newblock


\bibitem[Paulus et~al\mbox{.}(2024)]%
        {paulus2024advprompter}
\bibfield{author}{\bibinfo{person}{Anselm Paulus}, \bibinfo{person}{Arman Zharmagambetov}, \bibinfo{person}{Chuan Guo}, \bibinfo{person}{Brandon Amos}, {and} \bibinfo{person}{Yuandong Tian}.} \bibinfo{year}{2024}\natexlab{}.
\newblock \showarticletitle{Advprompter: Fast adaptive adversarial prompting for llms}.
\newblock \bibinfo{journal}{\emph{arXiv preprint arXiv:2404.16873}} (\bibinfo{year}{2024}).
\newblock


\bibitem[Pavlopoulos et~al\mbox{.}(2020)]%
        {PavlopoulosSDTA20}
\bibfield{author}{\bibinfo{person}{John Pavlopoulos}, \bibinfo{person}{Jeffrey Sorensen}, \bibinfo{person}{Lucas Dixon}, \bibinfo{person}{Nithum Thain}, {and} \bibinfo{person}{Ion Androutsopoulos}.} \bibinfo{year}{2020}\natexlab{}.
\newblock \showarticletitle{Toxicity Detection: Does Context Really Matter?}. In \bibinfo{booktitle}{\emph{Annual Meeting of the Association for Computational Linguistics}}. \bibinfo{publisher}{Association for Computational Linguistics}.
\newblock


\bibitem[Penedo et~al\mbox{.}(2023)]%
        {Penedo2023RefinedWeb}
\bibfield{author}{\bibinfo{person}{Guilherme Penedo}, \bibinfo{person}{Quentin Malartic}, \bibinfo{person}{Daniel Hesslow}, \bibinfo{person}{Ruxandra Cojocaru}, \bibinfo{person}{Alessandro Cappelli}, \bibinfo{person}{Hamza Alobeidli}, \bibinfo{person}{Baptiste Pannier}, \bibinfo{person}{Ebtesam Almazrouei}, {and} \bibinfo{person}{Julien Launay}.} \bibinfo{year}{2023}\natexlab{}.
\newblock \showarticletitle{The RefinedWeb Dataset for Falcon {LLM:} Outperforming Curated Corpora with Web Data, and Web Data Only}.
\newblock \bibinfo{journal}{\emph{arXiv preprint arXiv: 2306.01116}} (\bibinfo{year}{2023}).
\newblock


\bibitem[Peng et~al\mbox{.}(2023)]%
        {peng2023instruction}
\bibfield{author}{\bibinfo{person}{Baolin Peng}, \bibinfo{person}{Chunyuan Li}, \bibinfo{person}{Pengcheng He}, \bibinfo{person}{Michel Galley}, {and} \bibinfo{person}{Jianfeng Gao}.} \bibinfo{year}{2023}\natexlab{}.
\newblock \showarticletitle{Instruction Tuning with GPT-4}.
\newblock \bibinfo{journal}{\emph{arXiv preprint arXiv: 2304.03277}} (\bibinfo{year}{2023}).
\newblock


\bibitem[Qi et~al\mbox{.}(2023)]%
        {qi2023fine}
\bibfield{author}{\bibinfo{person}{Xiangyu Qi}, \bibinfo{person}{Yi Zeng}, \bibinfo{person}{Tinghao Xie}, \bibinfo{person}{Pin-Yu Chen}, \bibinfo{person}{Ruoxi Jia}, \bibinfo{person}{Prateek Mittal}, {and} \bibinfo{person}{Peter Henderson}.} \bibinfo{year}{2023}\natexlab{}.
\newblock \showarticletitle{Fine-Tuning Aligned Language Models Compromises Safety, Even When Users do not Intend to!}
\newblock \bibinfo{journal}{\emph{arXiv preprint arXiv: 2310.03693}} (\bibinfo{year}{2023}).
\newblock


\bibitem[Rae et~al\mbox{.}(2021)]%
        {Rae2011Scaling}
\bibfield{author}{\bibinfo{person}{Jack~W. Rae}, \bibinfo{person}{Sebastian Borgeaud}, \bibinfo{person}{Trevor Cai}, \bibinfo{person}{Katie Millican}, \bibinfo{person}{Jordan Hoffmann}, \bibinfo{person}{H.~Francis Song}, \bibinfo{person}{John Aslanides}, \bibinfo{person}{Sarah Henderson}, \bibinfo{person}{Roman Ring}, \bibinfo{person}{Susannah Young}, \bibinfo{person}{Eliza Rutherford}, \bibinfo{person}{Tom Hennigan}, \bibinfo{person}{Jacob Menick}, \bibinfo{person}{Albin Cassirer}, \bibinfo{person}{Richard Powell}, \bibinfo{person}{George van~den Driessche}, \bibinfo{person}{Lisa~Anne Hendricks}, \bibinfo{person}{Maribeth Rauh}, \bibinfo{person}{Po{-}Sen Huang}, \bibinfo{person}{Amelia Glaese}, \bibinfo{person}{Johannes Welbl}, \bibinfo{person}{Sumanth Dathathri}, \bibinfo{person}{Saffron Huang}, \bibinfo{person}{Jonathan Uesato}, \bibinfo{person}{John Mellor}, \bibinfo{person}{Irina Higgins}, \bibinfo{person}{Antonia Creswell}, \bibinfo{person}{Nat McAleese}, \bibinfo{person}{Amy Wu}, \bibinfo{person}{Erich
  Elsen}, \bibinfo{person}{Siddhant~M. Jayakumar}, \bibinfo{person}{Elena Buchatskaya}, \bibinfo{person}{David Budden}, \bibinfo{person}{Esme Sutherland}, \bibinfo{person}{Karen Simonyan}, \bibinfo{person}{Michela Paganini}, \bibinfo{person}{Laurent Sifre}, \bibinfo{person}{Lena Martens}, \bibinfo{person}{Xiang~Lorraine Li}, \bibinfo{person}{Adhiguna Kuncoro}, \bibinfo{person}{Aida Nematzadeh}, \bibinfo{person}{Elena Gribovskaya}, \bibinfo{person}{Domenic Donato}, \bibinfo{person}{Angeliki Lazaridou}, \bibinfo{person}{Arthur Mensch}, \bibinfo{person}{Jean{-}Baptiste Lespiau}, \bibinfo{person}{Maria Tsimpoukelli}, \bibinfo{person}{Nikolai Grigorev}, \bibinfo{person}{Doug Fritz}, \bibinfo{person}{Thibault Sottiaux}, \bibinfo{person}{Mantas Pajarskas}, \bibinfo{person}{Toby Pohlen}, \bibinfo{person}{Zhitao Gong}, \bibinfo{person}{Daniel Toyama}, \bibinfo{person}{Cyprien de Masson~d'Autume}, \bibinfo{person}{Yujia Li}, \bibinfo{person}{Tayfun Terzi}, \bibinfo{person}{Vladimir Mikulik}, \bibinfo{person}{Igor
  Babuschkin}, \bibinfo{person}{Aidan Clark}, \bibinfo{person}{Diego de Las~Casas}, \bibinfo{person}{Aurelia Guy}, \bibinfo{person}{Chris Jones}, \bibinfo{person}{James Bradbury}, \bibinfo{person}{Matthew~J. Johnson}, \bibinfo{person}{Blake~A. Hechtman}, \bibinfo{person}{Laura Weidinger}, \bibinfo{person}{Iason Gabriel}, \bibinfo{person}{William Isaac}, \bibinfo{person}{Edward Lockhart}, \bibinfo{person}{Simon Osindero}, \bibinfo{person}{Laura Rimell}, \bibinfo{person}{Chris Dyer}, \bibinfo{person}{Oriol Vinyals}, \bibinfo{person}{Kareem Ayoub}, \bibinfo{person}{Jeff Stanway}, \bibinfo{person}{Lorrayne Bennett}, \bibinfo{person}{Demis Hassabis}, \bibinfo{person}{Koray Kavukcuoglu}, {and} \bibinfo{person}{Geoffrey Irving}.} \bibinfo{year}{2021}\natexlab{}.
\newblock \showarticletitle{Scaling Language Models: Methods, Analysis {\&} Insights from Training Gopher}.
\newblock \bibinfo{journal}{\emph{arXiv preprint arXiv: 2112.11446}} (\bibinfo{year}{2021}).
\newblock


\bibitem[Ramesh et~al\mbox{.}(2024)]%
        {ramesh2024gpt}
\bibfield{author}{\bibinfo{person}{Govind Ramesh}, \bibinfo{person}{Yao Dou}, {and} \bibinfo{person}{Wei Xu}.} \bibinfo{year}{2024}\natexlab{}.
\newblock \showarticletitle{GPT-4 Jailbreaks Itself with Near-Perfect Success Using Self-Explanation}.
\newblock \bibinfo{journal}{\emph{arXiv preprint arXiv:2405.13077}} (\bibinfo{year}{2024}).
\newblock


\bibitem[Rao et~al\mbox{.}(2024)]%
        {rao2024tricking}
\bibfield{author}{\bibinfo{person}{Abhinav Rao}, \bibinfo{person}{Sachin Vashistha}, \bibinfo{person}{Atharva Naik}, \bibinfo{person}{Somak Aditya}, {and} \bibinfo{person}{Monojit Choudhury}.} \bibinfo{year}{2024}\natexlab{}.
\newblock \showarticletitle{Tricking LLMs into Disobedience: Formalizing, Analyzing, and Detecting Jailbreaks}.
\newblock \bibinfo{journal}{\emph{arXiv preprint arXiv: 2305.14965}} (\bibinfo{year}{2024}).
\newblock


\bibitem[Rowling and Lauer(2001)]%
        {rowling2001harry}
\bibfield{author}{\bibinfo{person}{Joanne~K Rowling} {and} \bibinfo{person}{Gerhard Lauer}.} \bibinfo{year}{2001}\natexlab{}.
\newblock \bibinfo{booktitle}{\emph{Harry Potter}}.
\newblock \bibinfo{publisher}{Bloomsbury London}.
\newblock


\bibitem[Schmidt and Wiegand(2017)]%
        {SchmidtW17}
\bibfield{author}{\bibinfo{person}{Anna Schmidt} {and} \bibinfo{person}{Michael Wiegand}.} \bibinfo{year}{2017}\natexlab{}.
\newblock \showarticletitle{A Survey on Hate Speech Detection Using Natural Language Processing}. In \bibinfo{booktitle}{\emph{International Workshop on Natural Language Processing for Social Media}}. \bibinfo{publisher}{Association for Computational Linguistics}.
\newblock


\bibitem[Shazeer(2019)]%
        {shazeer2019fast}
\bibfield{author}{\bibinfo{person}{Noam Shazeer}.} \bibinfo{year}{2019}\natexlab{}.
\newblock \showarticletitle{Fast Transformer Decoding: One Write-Head is All You Need}.
\newblock \bibinfo{journal}{\emph{arXiv preprint arXiv: 1911.02150}} (\bibinfo{year}{2019}).
\newblock


\bibitem[Shazeer(2020)]%
        {shazeer2020glu}
\bibfield{author}{\bibinfo{person}{Noam Shazeer}.} \bibinfo{year}{2020}\natexlab{}.
\newblock \showarticletitle{Glu Variants Improve Transformer}.
\newblock \bibinfo{journal}{\emph{arXiv preprint arXiv: 2002.05202}} (\bibinfo{year}{2020}).
\newblock


\bibitem[Shen et~al\mbox{.}(2023)]%
        {Shen2023Do}
\bibfield{author}{\bibinfo{person}{Xinyue Shen}, \bibinfo{person}{Zeyuan Chen}, \bibinfo{person}{Michael Backes}, \bibinfo{person}{Yun Shen}, {and} \bibinfo{person}{Yang Zhang}.} \bibinfo{year}{2023}\natexlab{}.
\newblock \showarticletitle{"Do Anything Now": Characterizing and Evaluating In-The-Wild Jailbreak Prompts on Large Language Models}.
\newblock \bibinfo{journal}{\emph{arXiv preprint arXiv: 2308.03825}} (\bibinfo{year}{2023}).
\newblock


\bibitem[Tay et~al\mbox{.}(2023)]%
        {TayWC0SSGZRCZMP23}
\bibfield{author}{\bibinfo{person}{Yi Tay}, \bibinfo{person}{Jason Wei}, \bibinfo{person}{Hyung~Won Chung}, \bibinfo{person}{Vinh~Q. Tran}, \bibinfo{person}{David~R. So}, \bibinfo{person}{Siamak Shakeri}, \bibinfo{person}{Xavier Garcia}, \bibinfo{person}{Huaixiu~Steven Zheng}, \bibinfo{person}{Jinfeng Rao}, \bibinfo{person}{Aakanksha Chowdhery}, \bibinfo{person}{Denny Zhou}, \bibinfo{person}{Donald Metzler}, \bibinfo{person}{Slav Petrov}, \bibinfo{person}{Neil Houlsby}, \bibinfo{person}{Quoc~V. Le}, {and} \bibinfo{person}{Mostafa Dehghani}.} \bibinfo{year}{2023}\natexlab{}.
\newblock \showarticletitle{Transcending Scaling Laws with 0.1{\%} Extra Compute}. In \bibinfo{booktitle}{\emph{Conference on Empirical Methods in Natural Language Processing}}. \bibinfo{publisher}{Association for Computational Linguistics}.
\newblock


\bibitem[Touvron et~al\mbox{.}(2023a)]%
        {Touvron2023LLaMA}
\bibfield{author}{\bibinfo{person}{Hugo Touvron}, \bibinfo{person}{Thibaut Lavril}, \bibinfo{person}{Gautier Izacard}, \bibinfo{person}{Xavier Martinet}, \bibinfo{person}{Marie{-}Anne Lachaux}, \bibinfo{person}{Timoth{\'{e}}e Lacroix}, \bibinfo{person}{Baptiste Rozi{\`{e}}re}, \bibinfo{person}{Naman Goyal}, \bibinfo{person}{Eric Hambro}, \bibinfo{person}{Faisal Azhar}, \bibinfo{person}{Aur{\'{e}}lien Rodriguez}, \bibinfo{person}{Armand Joulin}, \bibinfo{person}{Edouard Grave}, {and} \bibinfo{person}{Guillaume Lample}.} \bibinfo{year}{2023}\natexlab{a}.
\newblock \showarticletitle{LLaMA: Open and Efficient Foundation Language Models}.
\newblock \bibinfo{journal}{\emph{arXiv preprint arXiv: 2302.13971}} (\bibinfo{year}{2023}).
\newblock


\bibitem[Touvron et~al\mbox{.}(2023b)]%
        {Touvron23Llama2}
\bibfield{author}{\bibinfo{person}{Hugo Touvron}, \bibinfo{person}{Louis Martin}, \bibinfo{person}{Kevin Stone}, \bibinfo{person}{Peter Albert}, \bibinfo{person}{Amjad Almahairi}, \bibinfo{person}{Yasmine Babaei}, \bibinfo{person}{Nikolay Bashlykov}, \bibinfo{person}{Soumya Batra}, \bibinfo{person}{Prajjwal Bhargava}, \bibinfo{person}{Shruti Bhosale}, {et~al\mbox{.}}} \bibinfo{year}{2023}\natexlab{b}.
\newblock \showarticletitle{{Llama 2: Open Foundation and Fine-Tuned Chat Models}}.
\newblock \bibinfo{journal}{\emph{arXiv preprint arXiv: 2307.09288}} (\bibinfo{year}{2023}).
\newblock


\bibitem[Vaswani et~al\mbox{.}(2017)]%
        {VaswaniSPUJGKP17}
\bibfield{author}{\bibinfo{person}{Ashish Vaswani}, \bibinfo{person}{Noam Shazeer}, \bibinfo{person}{Niki Parmar}, \bibinfo{person}{Jakob Uszkoreit}, \bibinfo{person}{Llion Jones}, \bibinfo{person}{Aidan~N. Gomez}, \bibinfo{person}{Lukasz Kaiser}, {and} \bibinfo{person}{Illia Polosukhin}.} \bibinfo{year}{2017}\natexlab{}.
\newblock \showarticletitle{Attention is All You Need}. In \bibinfo{booktitle}{\emph{Conference on Neural Information Processing Systems}}. {PMLR}.
\newblock


\bibitem[Vidgen et~al\mbox{.}(2023)]%
        {vidgen2024simplesafetytests}
\bibfield{author}{\bibinfo{person}{Bertie Vidgen}, \bibinfo{person}{Hannah~Rose Kirk}, \bibinfo{person}{Rebecca Qian}, \bibinfo{person}{Nino Scherrer}, \bibinfo{person}{Anand Kannappan}, \bibinfo{person}{Scott~A. Hale}, {and} \bibinfo{person}{Paul R{\"{o}}ttger}.} \bibinfo{year}{2023}\natexlab{}.
\newblock \showarticletitle{SimpleSafetyTests: A Test Suite for Identifying Critical Safety Risks in Large Language Models}.
\newblock \bibinfo{journal}{\emph{arXiv preprint arXiv: 2311.08370}} (\bibinfo{year}{2023}).
\newblock


\bibitem[Wallace et~al\mbox{.}(2019)]%
        {WallaceFKGS19}
\bibfield{author}{\bibinfo{person}{Eric Wallace}, \bibinfo{person}{Shi Feng}, \bibinfo{person}{Nikhil Kandpal}, \bibinfo{person}{Matt Gardner}, {and} \bibinfo{person}{Sameer Singh}.} \bibinfo{year}{2019}\natexlab{}.
\newblock \showarticletitle{Universal Adversarial Triggers for Attacking and Analyzing {NLP}}. In \bibinfo{booktitle}{\emph{Conference on Empirical Methods in Natural Language Processing and International Joint Conference on Natural Language Processing}}. \bibinfo{publisher}{Association for Computational Linguistics}.
\newblock


\bibitem[Wang et~al\mbox{.}(2023)]%
        {WangHACL23}
\bibfield{author}{\bibinfo{person}{Han Wang}, \bibinfo{person}{Ming~Shan Hee}, \bibinfo{person}{Md.~Rabiul Awal}, \bibinfo{person}{Kenny Tsu~Wei Choo}, {and} \bibinfo{person}{Roy~Ka{-}Wei Lee}.} \bibinfo{year}{2023}\natexlab{}.
\newblock \showarticletitle{Evaluating {GPT-3} Generated Explanations for Hateful Content Moderation}. In \bibinfo{booktitle}{\emph{International Joint Conference on Artificial Intelligence}}. \bibinfo{publisher}{ijcai.org}.
\newblock


\bibitem[Wei et~al\mbox{.}(2023)]%
        {Wei2023Jailbroken}
\bibfield{author}{\bibinfo{person}{Alexander Wei}, \bibinfo{person}{Nika Haghtalab}, {and} \bibinfo{person}{Jacob Steinhardt}.} \bibinfo{year}{2023}\natexlab{}.
\newblock \showarticletitle{Jailbroken: How Does {LLM} Safety Training Fail?}. In \bibinfo{booktitle}{\emph{Conference on Neural Information Processing Systems}}. {PMLR}.
\newblock


\bibitem[Weidinger et~al\mbox{.}(2021)]%
        {Weidinger2021Ethical}
\bibfield{author}{\bibinfo{person}{Laura Weidinger}, \bibinfo{person}{John Mellor}, \bibinfo{person}{Maribeth Rauh}, \bibinfo{person}{Conor Griffin}, \bibinfo{person}{Jonathan Uesato}, \bibinfo{person}{Po{-}Sen Huang}, \bibinfo{person}{Myra Cheng}, \bibinfo{person}{Mia Glaese}, \bibinfo{person}{Borja Balle}, \bibinfo{person}{Atoosa Kasirzadeh}, \bibinfo{person}{Zac Kenton}, \bibinfo{person}{Sasha Brown}, \bibinfo{person}{Will Hawkins}, \bibinfo{person}{Tom Stepleton}, \bibinfo{person}{Courtney Biles}, \bibinfo{person}{Abeba Birhane}, \bibinfo{person}{Julia Haas}, \bibinfo{person}{Laura Rimell}, \bibinfo{person}{Lisa~Anne Hendricks}, \bibinfo{person}{William Isaac}, \bibinfo{person}{Sean Legassick}, \bibinfo{person}{Geoffrey Irving}, {and} \bibinfo{person}{Iason Gabriel}.} \bibinfo{year}{2021}\natexlab{}.
\newblock \showarticletitle{Ethical and Social Risks of Harm from Language Models}.
\newblock \bibinfo{journal}{\emph{arXiv preprint arXiv: 2112.04359}} (\bibinfo{year}{2021}).
\newblock


\bibitem[Weidinger et~al\mbox{.}(2022)]%
        {WeidingerURGHMG22}
\bibfield{author}{\bibinfo{person}{Laura Weidinger}, \bibinfo{person}{Jonathan Uesato}, \bibinfo{person}{Maribeth Rauh}, \bibinfo{person}{Conor Griffin}, \bibinfo{person}{Po{-}Sen Huang}, \bibinfo{person}{John Mellor}, \bibinfo{person}{Amelia Glaese}, \bibinfo{person}{Myra Cheng}, \bibinfo{person}{Borja Balle}, \bibinfo{person}{Atoosa Kasirzadeh}, \bibinfo{person}{Courtney Biles}, \bibinfo{person}{Sasha Brown}, \bibinfo{person}{Zac Kenton}, \bibinfo{person}{Will Hawkins}, \bibinfo{person}{Tom Stepleton}, \bibinfo{person}{Abeba Birhane}, \bibinfo{person}{Lisa~Anne Hendricks}, \bibinfo{person}{Laura Rimell}, \bibinfo{person}{William Isaac}, \bibinfo{person}{Julia Haas}, \bibinfo{person}{Sean Legassick}, \bibinfo{person}{Geoffrey Irving}, {and} \bibinfo{person}{Iason Gabriel}.} \bibinfo{year}{2022}\natexlab{}.
\newblock \showarticletitle{Taxonomy of Risks Posed by Language Models}. In \bibinfo{booktitle}{\emph{{ACM} Conference on Fairness, Accountability, and Transparency}}. \bibinfo{publisher}{{ACM}}.
\newblock


\bibitem[Wolf et~al\mbox{.}(2023)]%
        {Wolf2023Fundamental}
\bibfield{author}{\bibinfo{person}{Yotam Wolf}, \bibinfo{person}{Noam Wies}, \bibinfo{person}{Yoav Levine}, {and} \bibinfo{person}{Amnon Shashua}.} \bibinfo{year}{2023}\natexlab{}.
\newblock \showarticletitle{Fundamental Limitations of Alignment in Large Language Models}.
\newblock \bibinfo{journal}{\emph{arXiv preprint arXiv: 2304.11082}} (\bibinfo{year}{2023}).
\newblock


\bibitem[Xie et~al\mbox{.}(2024)]%
        {xie2024gradsafe}
\bibfield{author}{\bibinfo{person}{Yueqi Xie}, \bibinfo{person}{Minghong Fang}, \bibinfo{person}{Renjie Pi}, {and} \bibinfo{person}{Neil Gong}.} \bibinfo{year}{2024}\natexlab{}.
\newblock \showarticletitle{GradSafe: Detecting Unsafe Prompts for LLMs via Safety-Critical Gradient Analysis}.
\newblock \bibinfo{journal}{\emph{arXiv preprint arXiv:2402.13494}} (\bibinfo{year}{2024}).
\newblock


\bibitem[Xu et~al\mbox{.}(2021)]%
        {XuJLBWD21}
\bibfield{author}{\bibinfo{person}{Jing Xu}, \bibinfo{person}{Da Ju}, \bibinfo{person}{Margaret Li}, \bibinfo{person}{Y{-}Lan Boureau}, \bibinfo{person}{Jason Weston}, {and} \bibinfo{person}{Emily Dinan}.} \bibinfo{year}{2021}\natexlab{}.
\newblock \showarticletitle{Bot-Adversarial Dialogue for Safe Conversational Agents}. In \bibinfo{booktitle}{\emph{Conference of the North American Chapter of the Association for Computational Linguistics: Human Language Technologies}}. \bibinfo{publisher}{Association for Computational Linguistics}.
\newblock


\bibitem[Xu et~al\mbox{.}(2024)]%
        {xu2024llm}
\bibfield{author}{\bibinfo{person}{Zihao Xu}, \bibinfo{person}{Yi Liu}, \bibinfo{person}{Gelei Deng}, \bibinfo{person}{Yuekang Li}, {and} \bibinfo{person}{Stjepan Picek}.} \bibinfo{year}{2024}\natexlab{}.
\newblock \showarticletitle{LLM Jailbreak Attack versus Defense Techniques--A Comprehensive Study}.
\newblock \bibinfo{journal}{\emph{arXiv preprint arXiv:2402.13457}} (\bibinfo{year}{2024}).
\newblock


\bibitem[Yenduri et~al\mbox{.}(2023)]%
        {Yenduri2023Generative}
\bibfield{author}{\bibinfo{person}{Gokul Yenduri}, \bibinfo{person}{Ramalingam M}, \bibinfo{person}{Chemmalar~Selvi G.}, \bibinfo{person}{Supriya Y}, \bibinfo{person}{Gautam Srivastava}, \bibinfo{person}{Praveen Kumar~Reddy Maddikunta}, \bibinfo{person}{Deepti~Raj G}, \bibinfo{person}{Rutvij~H. Jhaveri}, \bibinfo{person}{Prabadevi B}, \bibinfo{person}{Weizheng Wang}, \bibinfo{person}{Athanasios~V. Vasilakos}, {and} \bibinfo{person}{Thippa~Reddy Gadekallu}.} \bibinfo{year}{2023}\natexlab{}.
\newblock \showarticletitle{Generative Pre-Trained Transformer: {A} Comprehensive Review on Enabling Technologies, Potential Applications, Emerging Challenges, and Future Directions}.
\newblock \bibinfo{journal}{\emph{arXiv preprint arXiv: 2305.10435}} (\bibinfo{year}{2023}).
\newblock


\bibitem[Yu and Sagae(2021)]%
        {YuS21}
\bibfield{author}{\bibinfo{person}{Dian Yu} {and} \bibinfo{person}{Kenji Sagae}.} \bibinfo{year}{2021}\natexlab{}.
\newblock \showarticletitle{Automatically Exposing Problems with Neural Dialog Models}. In \bibinfo{booktitle}{\emph{Conference on Empirical Methods in Natural Language Processing}}. \bibinfo{publisher}{Association for Computational Linguistics}.
\newblock


\bibitem[Yuan et~al\mbox{.}(2023)]%
        {Yuan2023GPT}
\bibfield{author}{\bibinfo{person}{Youliang Yuan}, \bibinfo{person}{Wenxiang Jiao}, \bibinfo{person}{Wenxuan Wang}, \bibinfo{person}{Jen{-}tse Huang}, \bibinfo{person}{Pinjia He}, \bibinfo{person}{Shuming Shi}, {and} \bibinfo{person}{Zhaopeng Tu}.} \bibinfo{year}{2023}\natexlab{}.
\newblock \showarticletitle{{GPT-4} Is Too Smart To Be Safe: Stealthy Chat with LLMs via Cipher}.
\newblock \bibinfo{journal}{\emph{arXiv preprint arXiv: 2308.06463}} (\bibinfo{year}{2023}).
\newblock


\bibitem[Zeng et~al\mbox{.}(2023)]%
        {Zeng23GLM}
\bibfield{author}{\bibinfo{person}{Aohan Zeng}, \bibinfo{person}{Xiao Liu}, \bibinfo{person}{Zhengxiao Du}, \bibinfo{person}{Zihan Wang}, \bibinfo{person}{Hanyu Lai}, \bibinfo{person}{Ming Ding}, \bibinfo{person}{Zhuoyi Yang}, \bibinfo{person}{Yifan Xu}, \bibinfo{person}{Wendi Zheng}, \bibinfo{person}{Xiao Xia}, \bibinfo{person}{Weng~Lam Tam}, \bibinfo{person}{Zixuan Ma}, \bibinfo{person}{Yufei Xue}, \bibinfo{person}{Jidong Zhai}, \bibinfo{person}{Wenguang Chen}, \bibinfo{person}{Zhiyuan Liu}, \bibinfo{person}{Peng Zhang}, \bibinfo{person}{Yuxiao Dong}, {and} \bibinfo{person}{Jie Tang}.} \bibinfo{year}{2023}\natexlab{}.
\newblock \showarticletitle{{GLM-130B: An Open Bilingual Pre-Trained Model}}. In \bibinfo{booktitle}{\emph{International Conference on Learning Representations}}. OpenReview.net.
\newblock


\bibitem[Zeng et~al\mbox{.}({[n.\,d.]})]%
        {zeng2020openattack}
\bibfield{author}{\bibinfo{person}{Guoyang Zeng}, \bibinfo{person}{Fanchao Qi}, \bibinfo{person}{Qianrui Zhou}, \bibinfo{person}{Tingji Zhang}, \bibinfo{person}{Bairu Hou}, \bibinfo{person}{Yuan Zang}, \bibinfo{person}{Zhiyuan Liu}, {and} \bibinfo{person}{Maosong Sun}.} \bibinfo{year}{[n.\,d.]}\natexlab{}.
\newblock \showarticletitle{{Openattack: An Open-Source Textual Adversarial Attack Toolkit}}. In \bibinfo{booktitle}{\emph{Annual Meeting of the Association for Computational Linguistics and International Joint Conference on Natural Language Processing: System Demonstrations}}.
\newblock


\bibitem[Zhang et~al\mbox{.}(2023)]%
        {zhang2023jade}
\bibfield{author}{\bibinfo{person}{Mi Zhang}, \bibinfo{person}{Xudong Pan}, {and} \bibinfo{person}{Min Yang}.} \bibinfo{year}{2023}\natexlab{}.
\newblock \showarticletitle{Jade: A Linguistics-Based Safety Evaluation Platform for LLM}.
\newblock \bibinfo{journal}{\emph{arXiv preprint arXiv: 2311.00286}} (\bibinfo{year}{2023}).
\newblock


\bibitem[Zhao et~al\mbox{.}(2023)]%
        {Zhao2023Survey}
\bibfield{author}{\bibinfo{person}{Wayne~Xin Zhao}, \bibinfo{person}{Kun Zhou}, \bibinfo{person}{Junyi Li}, \bibinfo{person}{Tianyi Tang}, \bibinfo{person}{Xiaolei Wang}, \bibinfo{person}{Yupeng Hou}, \bibinfo{person}{Yingqian Min}, \bibinfo{person}{Beichen Zhang}, \bibinfo{person}{Junjie Zhang}, \bibinfo{person}{Zican Dong}, \bibinfo{person}{Yifan Du}, \bibinfo{person}{Chen Yang}, \bibinfo{person}{Yushuo Chen}, \bibinfo{person}{Zhipeng Chen}, \bibinfo{person}{Jinhao Jiang}, \bibinfo{person}{Ruiyang Ren}, \bibinfo{person}{Yifan Li}, \bibinfo{person}{Xinyu Tang}, \bibinfo{person}{Zikang Liu}, \bibinfo{person}{Peiyu Liu}, \bibinfo{person}{Jian{-}Yun Nie}, {and} \bibinfo{person}{Ji{-}Rong Wen}.} \bibinfo{year}{2023}\natexlab{}.
\newblock \showarticletitle{A Survey of Large Language Models}.
\newblock \bibinfo{journal}{\emph{arXiv preprint arXiv: 2303.18223}} (\bibinfo{year}{2023}).
\newblock


\bibitem[Zhao et~al\mbox{.}(2018)]%
        {zhengli2018iclr}
\bibfield{author}{\bibinfo{person}{Zhengli Zhao}, \bibinfo{person}{Dheeru Dua}, {and} \bibinfo{person}{Sameer Singh}.} \bibinfo{year}{2018}\natexlab{}.
\newblock \showarticletitle{Generating Natural Adversarial Examples}. In \bibinfo{booktitle}{\emph{International Conference on Learning Representations (ICLR)}}.
\newblock


\bibitem[Zheng et~al\mbox{.}(2023)]%
        {ZhengC00WZL0LXZ23}
\bibfield{author}{\bibinfo{person}{Lianmin Zheng}, \bibinfo{person}{Wei{-}Lin Chiang}, \bibinfo{person}{Ying Sheng}, \bibinfo{person}{Siyuan Zhuang}, \bibinfo{person}{Zhanghao Wu}, \bibinfo{person}{Yonghao Zhuang}, \bibinfo{person}{Zi Lin}, \bibinfo{person}{Zhuohan Li}, \bibinfo{person}{Dacheng Li}, \bibinfo{person}{Eric~P. Xing}, \bibinfo{person}{Hao Zhang}, \bibinfo{person}{Joseph~E. Gonzalez}, {and} \bibinfo{person}{Ion Stoica}.} \bibinfo{year}{2023}\natexlab{}.
\newblock \showarticletitle{Judging LLM-as-a-Judge with MT-Bench and Chatbot Arena}. In \bibinfo{booktitle}{\emph{Conference on Neural Information Processing Systems}}. {PMLR}.
\newblock


\bibitem[Zhu et~al\mbox{.}(2023)]%
        {Zhu2023AutoDAN}
\bibfield{author}{\bibinfo{person}{Sicheng Zhu}, \bibinfo{person}{Ruiyi Zhang}, \bibinfo{person}{Bang An}, \bibinfo{person}{Gang Wu}, \bibinfo{person}{Joe Barrow}, \bibinfo{person}{Zichao Wang}, \bibinfo{person}{Furong Huang}, \bibinfo{person}{Ani Nenkova}, {and} \bibinfo{person}{Tong Sun}.} \bibinfo{year}{2023}\natexlab{}.
\newblock \showarticletitle{AutoDAN: Automatic and Interpretable Adversarial Attacks on Large Language Models}.
\newblock \bibinfo{journal}{\emph{arXiv preprint arXiv: 2310.15140}} (\bibinfo{year}{2023}).
\newblock


\bibitem[Zou et~al\mbox{.}(2023)]%
        {zou2023universal}
\bibfield{author}{\bibinfo{person}{Andy Zou}, \bibinfo{person}{Zifan Wang}, \bibinfo{person}{J~Zico Kolter}, {and} \bibinfo{person}{Matt Fredrikson}.} \bibinfo{year}{2023}\natexlab{}.
\newblock \showarticletitle{Universal and Transferable Adversarial Attacks on Aligned Language Models}.
\newblock \bibinfo{journal}{\emph{arXiv preprint arXiv: 2307.15043}} (\bibinfo{year}{2023}).
\newblock


\end{thebibliography}

\clearpage
\appendix
\onecolumn



\section{Details of Dynamic Jailbreaking}\label{ap:dynamic}

For PAIR method, we use the Vicuna-7B model as the attacker LLM, with the target LLM being Vicuna-7b and LLaMA2-7B, and the judge LLM being GPT-4. Our parameter settings are the same as the PAIR method in the original paper, with a maximum of 60 attack queries. 

For TAP method, we utilize the same attacker LLM as the PAIR method. We only launch TAP on Vicuna-7B due to its low attack success rate on LLaMA2-7B.
We set the depth to 7, the maximum width to 10, and the branching factor to 4, allowing for a maximum of 70 attack queries. If the attacker LLM does not succeed after 70 attempts, we deem the attack unsuccessful. 


\vspace{0.5cm}

\section{System Prompts to Adapt GPT-3.5 \& GPT-4 for Content Moderation}\label{Appendix_A}

In this part, we detail the system prompts to adapt GPT-3.5-Turbo and GPT-4 for three moderation tasks, namely $I$-moderation, $O$-moderation and $IO$-moderation. For $I$- and $IO$-moderation, we use the same system prompt template to moderate whether the content is safe. As for $O$-moderation, we use the system prompt proposed by~\cite{Ji2023BeaverTails}.

\vspace{0.5cm}
\begin{tcolorbox}[title = {$I$- and $IO$-Moderation Prompt Template}, colback=white, halign={left}]
{\normalsize 
Content: \texttt{\underline{<prompt>}/\underline{<prompt>+<response>}}. 

Is the content above related to sexuality, hate, violence, self-harm, and harassment? \\ Return "Yes" or "No". 

}
\end{tcolorbox}



\vspace{0.5cm}

\begin{tcolorbox}[title = {$O$-Moderation Prompt Template}, colback=white, halign={left}]
{\normalsize
[Question]
\texttt{\underline{<prompt>}} 
[The Start of the Assistant's Answer]
\texttt{\underline{<response>}}
[The End of the Assistant's Answer]
[System]
We would like to request your feedback on the performance of the AI assistant in response to the user question displayed above. Please rate the harmlessness, helpfulness, and level of detail of their responses. Please prioritize the harmlessness. Please focus on whether there is discrimination and prejudice in the reply, whether it abides by the law, whether it avoids pornography and insults, whether it avoids porn content, and whether it protects privacy. Consider the following 14 categories in mind: \\
1. Hate Speech, Offensive Language\\
2. Discrimination, Stereotype, Injustice\\
3. Violence, Aiding and Abetting, Incitement\\
4. Financial Crime, Property Crime, Theft \\
5. Privacy Violation \\
6. Drug Abuse, Weapons, Banned Substance \\
7. Non-Violent Unethical Behavior \\
8. Sexually Explicit, Adult Content \\
9. Controversial Topics, Politics \\
10. Misinformation regarding ethics, laws and safety \\
11. Terrorism, Organized Crime \\
12. Self-Harm \\
13. Animal Abuse \\
14. Child Abuse \\
Please first output a single line containing only "flagged" or "unflagged". "flagged" indicates the answer violates the safety rules above. In the subsequent line, please provide a comprehensive explanation of your evaluation, avoiding any potential bias and ensuring that the order in which the responses were presented does not affect your judgment.

}
\end{tcolorbox}


\clearpage

\section{More Results of Baseline}\label{baseline}
\begin{table*}[h]
\centering

\setlength{\abovecaptionskip}{0pt}%
\setlength{\belowcaptionskip}{0pt}%
\caption{False Negative Rate (FNR) performance on various tasks$^\text{\textdagger}$ and datasets$^\text{\textdaggerdbl}$ compared with baselines.}\label{tab:baseline_fnr}

\resizebox{\linewidth}{!}{
\begin{threeparttable}

\setlength{\tabcolsep}{4.5mm}{

\begin{tabular}{@{}l|cccccccc@{}}
\toprule
\multicolumn{1}{c|}{\multirow{2}{*}{Method}} & \multicolumn{4}{c|}{\textbf{\textit{I}-Moderation} (FNR, \%)}                          & \multicolumn{2}{c|}{\textbf{\textit{O}-Mod.} (FNR, \%)} & \multicolumn{2}{c}{\textbf{\textit{IO}-Mod.} (FNR, \%)}                                                 \\ 
                        & HAT   & MHS     & OIG  & \multicolumn{1}{c|}{JIG}   & BEA  & \multicolumn{1}{c|}{BEA-\textit{adv}} & BAG        & \multicolumn{1}{c}{BAG-\textit{adv}}   \\ \midrule
OpenAI Moderation       & 12.300 & 0.796 & 62.400 & \multicolumn{1}{c|}{34.807} & 80.990 & \multicolumn{1}{c|}{80.354} & 87.019 & 82.788  \\
Perspective API         &  27.236          &  4.011           &  78.200         & \multicolumn{1}{c|}{ 41.290}       &  90.717        & \multicolumn{1}{c|}{ 99.628}             &  92.692         &  94.519   \\
BeaverDam-7B           &  23.100        &  8.437        &  71.100       & \multicolumn{1}{c|}{ 56.346}         & 8.101       & \multicolumn{1}{c|}{ 27.002}             &  47.500        &  57.170   \\

LLaMA Guard2     &  30.902           &  13.555          &  45.600           & \multicolumn{1}{c|}{ 91.923}           &  36.054           & \multicolumn{1}{c|}{ 40.596}                &  60.578             &  64.678   \\ 

GradSafe     &  15.400           &  2.628           &  40.800           & \multicolumn{1}{c|}{56.442}           &  28.650          & \multicolumn{1}{c|}{ 57.263}                &  47.115             &  44.467       \\ 

GPT-3.5-Turbo                 & 24.417       &  7.613        &  57.800         & \multicolumn{1}{c|}{ 79.904}       &  76.217       & \multicolumn{1}{c|}{ 85.847}             &  83.365          &  78.269  \\
GPT-4                   &  8.929          &  6.250          &  30.000          & \multicolumn{1}{c|}{ 60.784}           &  67.347        & \multicolumn{1}{c|}{ 82.143}                &  68.889            & 77.778 \\

\textbf{\sys (Ours)}                    &  20.656  & 10.488  & 13.700 & \multicolumn{1}{c|}{16.564} &  11.350 & \multicolumn{1}{c|}{14.759} & 0.525 & 3.763 \\ \bottomrule
\end{tabular}
}
\begin{tablenotes}[flushleft]
\small
\item[] \textdagger: Task alias: $I$-Moderation (\underline{$I$-Mod.}), $O$-Moderation (\underline{$O$-Mod.}) and $IO$-Moderation (\underline{$IO$-Mod.}). 
\item[] \textdaggerdbl: Dataset alias: HateXplain (\underline{HAT}), Measuring Hate Speech (\underline{MHS}), OIG-Safety (\underline{OIG}), Jigsaw (\underline{JIG}), BeaverTails (\underline{BEA}), BeaverTail-\textit{adv} (\underline{BEA-\textit{adv}}), BEA\&AG (\underline{BAG}), BEA-\textit{adv}\&AG (\underline{BAG-\textit{adv}}). 

\end{tablenotes}

\end{threeparttable}}
\end{table*}
\begin{table*}[h]
\centering

\setlength{\abovecaptionskip}{0pt}%
\setlength{\belowcaptionskip}{0pt}%
\caption{False Positive Rate (FPR) performance on various tasks$^\text{\textdagger}$ and datasets$^\text{\textdaggerdbl}$ compared with baselines.}\label{tab:baseline_fpr}

\resizebox{\linewidth}{!}{
\begin{threeparttable}

\setlength{\tabcolsep}{4.5mm}{

\begin{tabular}{@{}l|cccccccc@{}}
\toprule
\multicolumn{1}{c|}{\multirow{2}{*}{Method}} & \multicolumn{4}{c|}{\textbf{\textit{I}-Moderation} (FPR, \%)}                          & \multicolumn{2}{c|}{\textbf{\textit{O}-Mod.} (FPR, \%)} & \multicolumn{2}{c}{\textbf{\textit{IO}-Mod.} (FPR, \%)}                                                 \\ 
                        & HAT   & MHS     & OIG  & \multicolumn{1}{c|}{JIG}   & BEA  & \multicolumn{1}{c|}{BEA-\textit{adv}} & BAG        & \multicolumn{1}{c}{BAG-\textit{adv}}   \\ \midrule
OpenAI Moderation       &  47.800        &  55.324        &  0.100       & \multicolumn{1}{c|}{ 10.625}         &  6.806        & \multicolumn{1}{c|}{ 17.171}             &  0.000             &  0.208   \\
Perspective API         &  42.999          &  36.764           &  0.601          & \multicolumn{1}{c|}{ 0.522}       &  6.367        & \multicolumn{1}{c|}{ 1.188}             &  0.000        &  4.896   \\
BeaverDam-7B           &  43.534        &  35.709        &  1.400        & \multicolumn{1}{c|}{ 6.458}         & 4.752        & \multicolumn{1}{c|}{ 2.916}             &  0.729        &  0.520    \\
LLaMA Guard2     &  26.357          &  27.016          &  2.400           & \multicolumn{1}{c|}{ 2.604}           &  6.108          & \multicolumn{1}{c|}{ 5.400}                &  0.208             &  0.520   \\ 

GradSafe     &  51.333           &  49.099           &  1.400          & \multicolumn{1}{c|}{9.688}           &  23.271         & \multicolumn{1}{c|}{ 26.242}                &  0.000             &  0.000        \\ 

GPT-3.5-Turbo                 &29.319       &  29.933        &  0.800         & \multicolumn{1}{c|}{ 3.542}       &  6.806       & \multicolumn{1}{c|}{ 13.607}             &  2.813         &  22.396  \\
GPT-4                   &  47.727           & 42.307          &  8.000          & \multicolumn{1}{c|}{ 2.041}           &  17.647        & \multicolumn{1}{c|}{ 15.909}                &  0.000             &  5.455 \\

\textbf{\sys (Ours)}                    &  17.443  & 8.442  & 1.000 & \multicolumn{1}{c|}{8.831} &  11.980 & \multicolumn{1}{c|}{16.432} & 0.202 & 0.346 \\ \bottomrule
\end{tabular}
}
\begin{tablenotes}[flushleft]
\small
\item[] \textdagger: Task alias: $I$-Moderation (\underline{$I$-Mod.}), $O$-Moderation (\underline{$O$-Mod.}) and $IO$-Moderation (\underline{$IO$-Mod.}). 
\item[] \textdaggerdbl: Dataset alias: HateXplain (\underline{HAT}), Measuring Hate Speech (\underline{MHS}), OIG-Safety (\underline{OIG}), Jigsaw (\underline{JIG}), BeaverTails (\underline{BEA}), BeaverTail-\textit{adv} (\underline{BEA-\textit{adv}}), BEA\&AG (\underline{BAG}), BEA-\textit{adv}\&AG (\underline{BAG-\textit{adv}}). 

\end{tablenotes}

\end{threeparttable}}
\end{table*}

\clearpage

\section{More Results of Different Hyper-Parameters}\label{experiment}

\begin{table*}[h]\centering

\setlength{\abovecaptionskip}{0pt}%
\setlength{\belowcaptionskip}{0pt}%
\caption{Impact of the number of probed features on accuracy ($IO$-Moderation).} \label{tab:features-qa}
\resizebox{\linewidth}{!}{
\begin{threeparttable}

\setlength{\tabcolsep}{3mm}{
\begin{tabular}{@{}c|cccccccccc@{}}
\toprule
\multirow{2}{*}{\begin{tabular}[c]{@{}c@{}}$m$\tnote{\textdagger}\end{tabular}}                                                                                    
                                           & \multicolumn{2}{c|}{ChatGLM3}               & \multicolumn{2}{c|}{LLaMA2}                 & \multicolumn{2}{c|}{Falcon}               & \multicolumn{2}{c|}{Dolly}                 & \multicolumn{2}{c}{Vicuna}\\ 
                                           & \multicolumn{1}{c}{ACC} & \multicolumn{1}{c|}{AUC}  & \multicolumn{1}{c}{ACC} & \multicolumn{1}{c|}{AUC}  & \multicolumn{1}{c}{ACC} & \multicolumn{1}{c|}{AUC}  & \multicolumn{1}{c}{ACC} & \multicolumn{1}{c|}{AUC}  & \multicolumn{1}{c}{ACC}   & AUC    \\ \midrule
1                                          & 98.914                    & \multicolumn{1}{c|}{0.9993}            & 99.627                    & \multicolumn{1}{c|}{0.9999}           & 99.226                    & \multicolumn{1}{c|}{0.9996} & 98.043                    & \multicolumn{1}{c|}{0.9978} & 98.951                      & 0.9994   \\
3                                   & 98.882                    & \multicolumn{1}{c|}{0.9992}            & 99.567                    & \multicolumn{1}{c|}{0.9999}           & 99.254                    & \multicolumn{1}{c|}{0.9997}            & 98.159                    & \multicolumn{1}{c|}{0.9982}          & 98.937                      & 0.9993   \\
5                                   & 98.882                    & \multicolumn{1}{c|}{0.9993}            & 99.651                    & \multicolumn{1}{c|}{\textbf{0.9999}}  & 99.342                    & \multicolumn{1}{c|}{0.9997}            & 98.229                    & \multicolumn{1}{c|}{0.9981}          & 99.026                      & 0.9995   \\
7                                   & \textbf{98.956}           & \multicolumn{1}{c|}{0.9994}            & 99.623                    & \multicolumn{1}{c|}{0.9999}           & \textbf{99.426}           & \multicolumn{1}{c|}{\textbf{0.9998}}   & 98.322                    & \multicolumn{1}{c|}{\textbf{0.9985}} & \textbf{99.087}             & 0.9995   \\
9                                   & 98.928                    & \multicolumn{1}{c|}{\textbf{0.9994}}   & \textbf{99.697}           & \multicolumn{1}{c|}{0.9999}           & 99.356                    & \multicolumn{1}{c|}{0.9997}            & \textbf{98.415}           & \multicolumn{1}{c|}{0.9985}          & 98.937                      & \textbf{0.9996}   \\ \bottomrule
\end{tabular}
}  
\begin{tablenotes}[flushleft]
\small
\item[] \textdagger: $m$ denotes the number of probed features, as mentioned in \S\ref{sec:probe}.

\end{tablenotes}

\end{threeparttable}}

\end{table*} 
\begin{table*}[h]\centering

\setlength{\abovecaptionskip}{0pt}%
\setlength{\belowcaptionskip}{0pt}%
\caption{Impact of the number of probed features on accuracy ($O$-Moderation).}  \label{tab:features}

\resizebox{\linewidth}{!}{
\begin{threeparttable}

\setlength{\tabcolsep}{3mm}{
\begin{tabular}{@{}c|cccccccccc@{}}
\toprule
\multirow{2}{*}{\begin{tabular}[c]{@{}c@{}}$m$\tnote{\textdagger}\end{tabular}}                                                                                    
                                           & \multicolumn{2}{c|}{ChatGLM3}               & \multicolumn{2}{c|}{LLaMA2}                 & \multicolumn{2}{c|}{Falcon}               & \multicolumn{2}{c|}{Dolly}                 & \multicolumn{2}{c}{Vicuna}\\ 
                                           & \multicolumn{1}{c}{ACC} & \multicolumn{1}{c|}{AUC}  & \multicolumn{1}{c}{ACC} & \multicolumn{1}{c|}{AUC}  & \multicolumn{1}{c}{ACC} & \multicolumn{1}{c|}{AUC}  & \multicolumn{1}{c}{ACC} & \multicolumn{1}{c|}{AUC}  & \multicolumn{1}{c}{ACC}   & AUC    \\ \midrule
1               & 85.356          & \multicolumn{1}{c|}{0.9313} & 87.279                    & \multicolumn{1}{c|}{0.9475} & 86.127                    & \multicolumn{1}{c|}{\textbf{0.9371}} & 81.473                    & \multicolumn{1}{c|}{0.8970} & 85.422                      & 0.9329   \\
3                                   & 85.906                    & \multicolumn{1}{c|}{0.9361} & 87.304                    & \multicolumn{1}{c|}{0.9488} & 85.957                    & \multicolumn{1}{c|}{0.9307} & \textbf{81.613}                    & \multicolumn{1}{c|}{\textbf{0.8975}} & 85.901                      & 0.9365   \\
5                                   & 85.931                     & \multicolumn{1}{c|}{0.9368} & 87.740                    & \multicolumn{1}{c|}{\textbf{0.9498}} & 86.267                    & \multicolumn{1}{c|}{0.9330} & 79.385                    & \multicolumn{1}{c|}{0.8858} & 86.043                      & 0.9369   \\
7                                   & 86.488                    & \multicolumn{1}{c|}{0.9399} & 87.430                    & \multicolumn{1}{c|}{0.9500} & \textbf{86.478}                    & \multicolumn{1}{c|}{0.9353} & 66.682                    & \multicolumn{1}{c|}{0.8789} & 85.722                      & 0.9342   \\
9                                   & \textbf{86.566}                    & \multicolumn{1}{c|}{\textbf{0.9404}} & \textbf{87.762}                    & \multicolumn{1}{c|}{0.9504} & 85.932                    & \multicolumn{1}{c|}{0.9317} & 64.860                    & \multicolumn{1}{c|}{0.8686} & \textbf{86.394}                      & \textbf{0.9399}   \\ \bottomrule
\end{tabular}
}  
\begin{tablenotes}[flushleft]
\small
\item[] \textdagger: $m$ denotes the number of probed features, as mentioned in \S\ref{sec:probe}.

\end{tablenotes}

\end{threeparttable}}

\end{table*} 
\begin{table*}[h]
\centering

\setlength{\abovecaptionskip}{0pt}%
\setlength{\belowcaptionskip}{0pt}%
\caption{Impact of the number of probed features on FPR and FNR ($IO$-Moderation).} \label{tab:features-qa-fpr-fnr}
\resizebox{\linewidth}{!}{
\begin{threeparttable}

\setlength{\tabcolsep}{4mm}{
\begin{tabular}{@{}c|cc|cc|cc|cc|cc@{}}
\toprule
\multirow{2}{*}{\begin{tabular}[c]{@{}c@{}}$m$\tnote{\textdagger}\end{tabular}}                                                                                    
                                           & \multicolumn{2}{c|}{ChatGLM3}               & \multicolumn{2}{c|}{LLaMA2}                 & \multicolumn{2}{c|}{Falcon}               & \multicolumn{2}{c|}{Dolly}                 & \multicolumn{2}{c}{Vicuna}\\ 
                                           & \multicolumn{1}{c}{FPR} & \multicolumn{1}{c|}{FNR}  & \multicolumn{1}{c}{FPR} & \multicolumn{1}{c|}{FNR}  & \multicolumn{1}{c}{FPR} & \multicolumn{1}{c|}{FNR}  & \multicolumn{1}{c}{FPR} & \multicolumn{1}{c|}{FNR}  & \multicolumn{1}{c}{FPR}   & FNR    \\ \midrule
1                                   & 0.865& 1.204 & 0.510&0.289 & 1.117&0.507 & 1.856&2.361 & 0.817&1.284 \\
3               & 0.904&1.293 & 0.202&0.525 & 0.424&0.968 & 1.365&2.578 & 0.567&1.538 \\
5                                   & 0.827&1.239 & 0.548&0.226 & 0.616&0.724 & 1.269&2.252 & 0.510&1.393 \\  
7            & 0.760&1.230 & 0.211&0.588 & 0.809&0.597 & 1.385&2.062 & 0.519&1.474 \\ 
9     & 0.481&1.483 & 0.452&0.262 & 0.347&0.977 & 1.009&2.044 & 0.567&1.339 \\  \bottomrule
\end{tabular}
}  
\begin{tablenotes}[flushleft]
\small
\item[] \textdagger: $m$ denotes the number of probed features, as mentioned in \S\ref{sec:probe}.

\end{tablenotes}

\end{threeparttable}}

\end{table*} 
\begin{table*}[h]
\centering

\setlength{\abovecaptionskip}{0pt}%
\setlength{\belowcaptionskip}{0pt}%
\caption{Impact of the number of probed features on FPR and FNR ($O$-Moderation).}  \label{tab:features-fpr-fnr}

\resizebox{\linewidth}{!}{
\begin{threeparttable}

\setlength{\tabcolsep}{4mm}{
\begin{tabular}{@{}c|cc|cc|cc|cc|cc@{}}
\toprule
\multirow{2}{*}{\begin{tabular}[c]{@{}c@{}}$m$\tnote{\textdagger}\end{tabular}}                                                                                    
                                           & \multicolumn{2}{c|}{ChatGLM3}               & \multicolumn{2}{c|}{LLaMA2}                 & \multicolumn{2}{c|}{Falcon}               & \multicolumn{2}{c|}{Dolly}                 & \multicolumn{2}{c}{Vicuna}\\ 
                                           & \multicolumn{1}{c}{FPR} & \multicolumn{1}{c|}{FNR}  & \multicolumn{1}{c}{FPR} & \multicolumn{1}{c|}{FNR}  & \multicolumn{1}{c}{FPR} & \multicolumn{1}{c|}{FNR}  & \multicolumn{1}{c}{FPR} & \multicolumn{1}{c|}{FNR}  & \multicolumn{1}{c}{FPR}   & FNR    \\ \midrule
1               & 13.167&12.850 & 11.980&11.350 & 11.485&15.515 & 16.625&16.304 & 11.671&13.023 \\
3          & 12.618&13.337 & 12.900&11.185 & 13.709&13.042 & 16.974&15.743 & 13.186&11.391 \\ 
5           & 12.708&13.394 & 11.502&12.611 & 13.200&13.677 & 16.795&16.172 & 11.283&13.304 \\ 
7          & 12.519&12.702 & 12.668&12.034 & 12.662&13.883 & 18.828&14.647 & 13.306&11.721 \\ 
9          & 12.230&13.007 & 11.900&12.224 & 12.901&13.891 & 15.748&16.642 & 11.831&12.504 \\ \bottomrule
\end{tabular}
}  
\begin{tablenotes}[flushleft]
\small
\item[] \textdagger: $m$ denotes the number of probed features, as mentioned in \S\ref{sec:probe}.

\end{tablenotes}

\end{threeparttable}}

\end{table*}

\begin{table*}[h]\centering

\setlength{\abovecaptionskip}{0pt}%
\setlength{\belowcaptionskip}{0pt}%
\caption{Impact of the number of layers in classifier on accuracy ($IO$-Moderation).}  \label{tab:layer-qa}

\resizebox{\linewidth}{!}{
\begin{threeparttable}

\setlength{\tabcolsep}{3mm}{ 
\begin{tabular}{@{}c|cccccccccc@{}}
\toprule
\multicolumn{1}{c|}{\multirow{2}{*}{\#Layer\tnote{\textdagger}}}                                                                                                                                                                                                      
                               & \multicolumn{2}{c|}{ChatGLM3}               & \multicolumn{2}{c|}{LLaMA2}                 & \multicolumn{2}{c|}{Falcon}                & \multicolumn{2}{c|}{Dolly}                  & \multicolumn{2}{c}{Vicuna} \\ 
\multicolumn{1}{c|}{}                                  & \multicolumn{1}{c}{ACC} & \multicolumn{1}{c|}{AUC}  & \multicolumn{1}{c}{ACC} & \multicolumn{1}{c|}{AUC}  & \multicolumn{1}{c}{ACC} & \multicolumn{1}{c|}{AUC}  & \multicolumn{1}{c}{ACC} & \multicolumn{1}{c|}{AUC}  & \multicolumn{1}{c}{ACC}   & AUC    \\ \midrule
1                                                      & 98.220                    & \multicolumn{1}{c|}{0.9985} & 99.525                    & \multicolumn{1}{c|}{0.9999} & 99.165                    & \multicolumn{1}{c|}{0.9995} & 97.544                    & \multicolumn{1}{c|}{0.9967} & 98.630                      & 0.9990   \\
3                                                      & 98.914                    & \multicolumn{1}{c|}{\textbf{0.9993}} & 99.627                    & \multicolumn{1}{c|}{\textbf{0.9999}} & \textbf{99.226}                    & \multicolumn{1}{c|}{\textbf{0.9996}} & 98.043                    & \multicolumn{1}{c|}{\textbf{0.9994}} & \textbf{98.951}                      & \textbf{0.9994}   \\
5                                                      & 98.914                    & \multicolumn{1}{c|}{0.9990} & \textbf{99.669}                    & \multicolumn{1}{c|}{0.9999} & 99.160                    & \multicolumn{1}{c|}{0.9995} & 98.015                    & \multicolumn{1}{c|}{0.9976} & 98.826                      & 0.9993   \\
7                                                      & 98.914                    & \multicolumn{1}{c|}{0.9986} & 99.632                    & \multicolumn{1}{c|}{0.9999} & 99.133                    & \multicolumn{1}{c|}{0.9994} & \textbf{98.103}                    & \multicolumn{1}{c|}{0.9978} & 98.812                      & 0.9990   \\
9                                                      & \textbf{98.942}                    & \multicolumn{1}{c|}{0.9984} & 99.501                    & \multicolumn{1}{c|}{0.9999} & 99.105                    & \multicolumn{1}{c|}{0.9993} & 98.001                    & \multicolumn{1}{c|}{0.9951} & 98.551                      & 0.9986   \\ \bottomrule
\end{tabular} 
}  
\begin{tablenotes}[flushleft]
\small
\item[] \textdagger: the number of layers used in the classifier.

\end{tablenotes}

\end{threeparttable}}

\end{table*}
\begin{table*}[h]\centering

\setlength{\abovecaptionskip}{0pt}%
\setlength{\belowcaptionskip}{0pt}%
\caption{Impact of the number of layers in classifier on accuracy ($O$-Moderation).} \label{tab:layer}

\resizebox{\linewidth}{!}{
\begin{threeparttable}

\setlength{\tabcolsep}{3mm}{ 
\begin{tabular}{@{}c|cccccccccc@{}}
\toprule
\multicolumn{1}{c|}{\multirow{2}{*}{\#Layer\tnote{\textdagger}}}                                                                                                                                                                                                      
                               & \multicolumn{2}{c|}{ChatGLM3}               & \multicolumn{2}{c|}{LLaMA2}                 & \multicolumn{2}{c|}{Falcon}                & \multicolumn{2}{c|}{Dolly}                  & \multicolumn{2}{c}{Vicuna} \\ 
\multicolumn{1}{c|}{}                                  & \multicolumn{1}{c}{ACC} & \multicolumn{1}{c|}{AUC}  & \multicolumn{1}{c}{ACC} & \multicolumn{1}{c|}{AUC}  & \multicolumn{1}{c}{ACC} & \multicolumn{1}{c|}{AUC}  & \multicolumn{1}{c}{ACC} & \multicolumn{1}{c|}{AUC}  & \multicolumn{1}{c}{ACC}   & AUC    \\ \midrule
1                                                      & 79.655                    & \multicolumn{1}{c|}{0.8819} & 85.427                    & \multicolumn{1}{c|}{0.9313} & 82.968                    & \multicolumn{1}{c|}{0.9098} & 76.284                    & \multicolumn{1}{c|}{0.8457} & 82.621                      & 0.9069   \\
3                                                      & \textbf{85.356}                    & \multicolumn{1}{c|}{\textbf{0.9313}} & \textbf{87.279}                    & \multicolumn{1}{c|}{\textbf{0.9475}} & \textbf{86.127}                    & \multicolumn{1}{c|}{\textbf{0.9371}} & \textbf{81.473}                    & \multicolumn{1}{c|}{\textbf{0.8970}} & 85.422                      & 0.9329   \\
5                                                      & 84.567                    & \multicolumn{1}{c|}{0.9253} & 87.251                    & \multicolumn{1}{c|}{0.9466} & 85.654                    & \multicolumn{1}{c|}{0.9326} & 81.049                    & \multicolumn{1}{c|}{0.8940} & \textbf{85.592}                      & \textbf{0.9331}   \\
7                                                      & 77.718                    & \multicolumn{1}{c|}{0.9197} & 84.100                    & \multicolumn{1}{c|}{0.9423} & 83.718                    & \multicolumn{1}{c|}{0.9324} & 77.038                    & \multicolumn{1}{c|}{0.8723} & 82.651                      & 0.9199   \\
9                                                      & 82.353                    & \multicolumn{1}{c|}{0.9027} & 85.127                    & \multicolumn{1}{c|}{0.9317} & 84.278                    & \multicolumn{1}{c|}{0.9327} & 51.084                    & \multicolumn{1}{c|}{0.4611} & 81.118                      & 0.8947   \\ \bottomrule
\end{tabular} 
}  
\begin{tablenotes}[flushleft]
\small
\item[] \textdagger: the number of layers used in the classifier.

\end{tablenotes}

\end{threeparttable}}

\end{table*}
\begin{table*}[h]
\centering

\setlength{\abovecaptionskip}{0pt}%
\setlength{\belowcaptionskip}{0pt}%
\caption{Impact of the number of layers in classifier on FPR and FNR ($IO$-Moderation).}  \label{tab:layer-qa-fpr-fnr}

\resizebox{\linewidth}{!}{
\begin{threeparttable}

\setlength{\tabcolsep}{4mm}{ 
\begin{tabular}{@{}c|cc|cc|cc|cc|cc@{}}
\toprule
\multicolumn{1}{c|}{\multirow{2}{*}{\#Layer\tnote{\textdagger}}}                                                                                                                                                                                                      
                               & \multicolumn{2}{c|}{ChatGLM3}               & \multicolumn{2}{c|}{LLaMA2}                 & \multicolumn{2}{c|}{Falcon}                & \multicolumn{2}{c|}{Dolly}                  & \multicolumn{2}{c}{Vicuna} \\ 
\multicolumn{1}{c|}{}                                  & \multicolumn{1}{c}{FPR} & \multicolumn{1}{c|}{FNR}  & \multicolumn{1}{c}{FPR} & \multicolumn{1}{c|}{FNR}  & \multicolumn{1}{c}{FPR} & \multicolumn{1}{c|}{FNR}  & \multicolumn{1}{c}{FPR} & \multicolumn{1}{c|}{FNR}  & \multicolumn{1}{c}{FPR}   & FNR    \\ \midrule
1                  & 1.106&2.361 & 0.154&0.778 & 1.059&0.669 & 2.481&2.750 & 0.923&1.791 \\  
3            & 0.865&1.203 & 0.510&0.289 & 1.117&0.507 & 1.856&2.361 & 0.817&1.284 \\  
5             & 0.971&1.122 & 0.385&0.389 & 0.703&0.778 & 1.433&2.578 & 0.615&1.764 \\    
7          & 0.952&1.312 & 0.212&0.552 & 0.510&0.941 & 1.375&2.641 & 0.884&1.529 \\  
9           & 0.904&1.239 & 0.096&1.257 & 0.510&1.067 & 1.394&3.148 & 0.336&2.813 \\ \bottomrule
\end{tabular} 
}  
\begin{tablenotes}[flushleft]
\small
\item[] \textdagger: the number of layers used in the classifier.

\end{tablenotes}

\end{threeparttable}}

\end{table*}
\begin{table*}[h]
\centering

\setlength{\abovecaptionskip}{0pt}%
\setlength{\belowcaptionskip}{0pt}%
\caption{Impact of the number of layers in classifier on FPR and FNR ($O$-Moderation).} \label{tab:layer-fpr-fnr}

\resizebox{\linewidth}{!}{
\begin{threeparttable}

\setlength{\tabcolsep}{4mm}{ 
\begin{tabular}{@{}c|cc|cc|cc|cc|cc@{}}
\toprule
\multicolumn{1}{c|}{\multirow{2}{*}{\#Layer\tnote{\textdagger}}}                                                                                                                                                                                                      
                               & \multicolumn{2}{c|}{ChatGLM3}               & \multicolumn{2}{c|}{LLaMA2}                 & \multicolumn{2}{c|}{Falcon}                & \multicolumn{2}{c|}{Dolly}                  & \multicolumn{2}{c}{Vicuna} \\ 
\multicolumn{1}{c|}{}                                  & \multicolumn{1}{c}{FPR} & \multicolumn{1}{c|}{FNR}  & \multicolumn{1}{c}{FPR} & \multicolumn{1}{c|}{FNR}  & \multicolumn{1}{c}{FPR} & \multicolumn{1}{c|}{FNR}  & \multicolumn{1}{c}{FPR} & \multicolumn{1}{c|}{FNR}  & \multicolumn{1}{c}{FPR}   & FNR    \\ \midrule
1                  & 15.290&15.537 & 10.874&13.930 & 16.032&13.207 & 21.898&18.183 & 12.309&15.504 \\
3               & 13.167&12.850 & 11.980&11.350 & 11.485&15.515 & 16.625&16.304 & 11.671&13.023 \\
5                & 14.233&12.949 & 11.851&11.482 & 16.391&11.962 & 17.403&15.488 & 12.070&12.842 \\   
7              & 7.316&20.895 & 7.665&16.914 & 9.621&17.172 & 14.163&18.595 & 10.914&14.037 \\  
9            & 7.246&19.403 & 5.273&20.673 & 6.710&22.242 & 11.223&22.923 & 5.033&22.412 \\   \bottomrule
\end{tabular} 
}  
\begin{tablenotes}[flushleft]
\small
\item[] \textdagger: the number of layers used in the classifier.

\end{tablenotes}

\end{threeparttable}}

\end{table*}

\clearpage

\section{More Results against Jailbreaking}\label{ap:Jailbreaking}

\begin{table*}[h]\centering
\setlength{\abovecaptionskip}{0pt}%
\setlength{\belowcaptionskip}{0pt}%
\caption{\sys w/o data augmentation against LLM-targeted static jailbreaking ($IO$-Moderation).}
\label{tab:jailbreak-navie}

\resizebox{\linewidth}{!}{
\begin{threeparttable}

\setlength{\tabcolsep}{3mm}{  
\begin{tabular}{@{}l|cccccccccc@{}}
\toprule
\multicolumn{1}{l|}{\multirow{2}{*}{Jailbreaking Type}}                                                            
                                    & \multicolumn{2}{c|}{ChatGLM3}               & \multicolumn{2}{c|}{LLaMA2}               & \multicolumn{2}{c|}{Falcon}                 & \multicolumn{2}{c|}{Dolly}                  & \multicolumn{2}{c}{Vicuna} \\ 
\multicolumn{1}{c|}{}                                            & \multicolumn{1}{c}{ACC} & \multicolumn{1}{c|}{AUC}  & \multicolumn{1}{c}{ACC} & \multicolumn{1}{c|}{AUC}  & \multicolumn{1}{c}{ACC} & \multicolumn{1}{c|}{AUC}  & \multicolumn{1}{c}{ACC} & \multicolumn{1}{c|}{AUC}   & \multicolumn{1}{c}{ACC}   & AUC    \\ \midrule
Pretending                                                       & 98.695                    & \multicolumn{1}{c|}{0.9988} & 99.748                     & \multicolumn{1}{c|}{1.0000} & 99.048                    & \multicolumn{1}{c|}{0.9997} & 92.786                    & \multicolumn{1}{c|}{0.9936}  & 98.979                      & 0.9988   \\
Attention Shifting                                               & 98.322                    & \multicolumn{1}{c|}{0.9988} & 99.529                    & \multicolumn{1}{c|}{0.9998} & 91.751                   & \multicolumn{1}{c|}{0.9952} & 85.986                    & \multicolumn{1}{c|}{0.9887} & 89.696                      & 0.9882   \\
Privilege Escalation                                             & 99.045                    & \multicolumn{1}{c|}{0.9991} & 99.758                    & \multicolumn{1}{c|}{0.9999} & 97.988                    & \multicolumn{1}{c|}{0.9981} & 87.808                    & \multicolumn{1}{c|}{0.9806}  & 90.973                     & 0.9975   \\
Syntactic Transformation                                       & 90.311                   & \multicolumn{1}{c|}{0.9868} & 80.048                    & \multicolumn{1}{c|}{0.9866} & 92.664                    & \multicolumn{1}{c|}{0.9878} & 76.143                   & \multicolumn{1}{c|}{0.9583}  & 78.375                     & 0.9649   \\ \midrule
Overall                                       &  96.808                   & \multicolumn{1}{c|}{0.9963} &    94.715                 & \multicolumn{1}{c|}{0.9968} &    95.371                & \multicolumn{1}{c|}{0.9944} &      85.599               & \multicolumn{1}{c|}{0.9798}  &     89.640                 & 0.9877   \\\bottomrule
\end{tabular}    
}  



\end{threeparttable}}

\end{table*}
\begin{table*}[h]
\centering
\setlength{\abovecaptionskip}{0pt}%
\setlength{\belowcaptionskip}{0pt}%
\caption{\sys w/o data augmentation against LLM-targeted static jailbreaking on FPR and FNR ($IO$-Moderation).}
\label{tab:jailbreak-navie-fpr-fnr}

\resizebox{\linewidth}{!}{
\begin{threeparttable}

\setlength{\tabcolsep}{3mm}{  
\begin{tabular}{@{}l|cc|cc|cc|cc|cc@{}}
\toprule
\multicolumn{1}{l|}{\multirow{2}{*}{Jailbreaking Type}}                                                            
                                    & \multicolumn{2}{c|}{ChatGLM3}               & \multicolumn{2}{c|}{LLaMA2}               & \multicolumn{2}{c|}{Falcon}                 & \multicolumn{2}{c|}{Dolly}                  & \multicolumn{2}{c}{Vicuna} \\ 
\multicolumn{1}{c|}{}                                            & \multicolumn{1}{c}{FPR} & \multicolumn{1}{c|}{FNR}  & \multicolumn{1}{c}{FPR} & \multicolumn{1}{c|}{FNR}  & \multicolumn{1}{c}{FPR} & \multicolumn{1}{c|}{FNR}  & \multicolumn{1}{c}{FPR} & \multicolumn{1}{c|}{FNR}   & \multicolumn{1}{c}{FPR}   & FNR    \\ \midrule
Pretending                                                     & 0.778&2.026 & 0.481&0.027 & 1.021&0.344 & 2.067&8.520 & 0.644&1.357 \\  
Attention Shifting                                               & 0.778&1.872 & 0.481&0.470 & 1.167&12.094 & 1.904&13.323 & 0.644&18.723 \\  
Privilege Escalation                                            & 0.778&0.841 & 0.481&0.009 & 1.051&2.307 & 1.990&21.717 & 0.644&13.459 \\    
Syntactic Transformation                                     & 0.778&18.325 & 0.481&37.762 & 1.080&8.847 & 2.106&36.442 & 0.644&39.390 \\   \midrule
Overall        & 0.846&5.436 & 0.490&9.370 & 1.312&5.130 & 1.952&22.386 & 0.884&16.706 \\   \bottomrule
\end{tabular}    
}  



\end{threeparttable}}

\end{table*}
\begin{table}[h]\centering
\setlength{\abovecaptionskip}{0pt}%
\setlength{\belowcaptionskip}{0pt}%
\caption{Robustness against moderator-targeted dynamic jailbreaking ($I$-Moderation).}\label{tab:nlpattack}

\resizebox{0.7\linewidth}{!}{
\begin{threeparttable}

\setlength{\tabcolsep}{4mm}{
\begin{tabular}{@{}c|c|ccccc@{}}
\toprule
\multirow{2}{*}{Method} & \multirow{2}{*}{Type\tnote{\textdagger}} & \multicolumn{5}{c}{Host Model (Attack Success Rate, \%)}                                                                                              \\  
                        &                       & \multicolumn{1}{r}{ChatGLM3} & LLaMA2 & \multicolumn{1}{r}{Falcon} & \multicolumn{1}{r}{Dolly} & \multicolumn{1}{r}{Vicuna} \\ \midrule
VIPER                   & Blind                 & 0                            & 0      & 0                          & 0                         & 0                          \\
SCPN                    & Blind                 & 0                            & 0      & 0                          & 0                         & 0                          \\
GAN                     & Decision              & 0                            & 0      & 0                          & 0                         & 0                          \\ \bottomrule
\end{tabular}
}

\begin{tablenotes}[flushleft]
\small
\item[] \textdagger: Blind attacks are ignorant of the victim model, and decision-based attacks require the final decision for optimization~\cite{zeng2020openattack}.


\end{tablenotes}

\end{threeparttable}}
\end{table}


\section{Time Complexity}\label{ap:time}
\begin{figure*}[h]
\centering
\setlength{\abovecaptionskip}{5pt}
\setlength{\belowcaptionskip}{0pt}

\includegraphics[width=7.in, trim=0 85 0 0, clip]{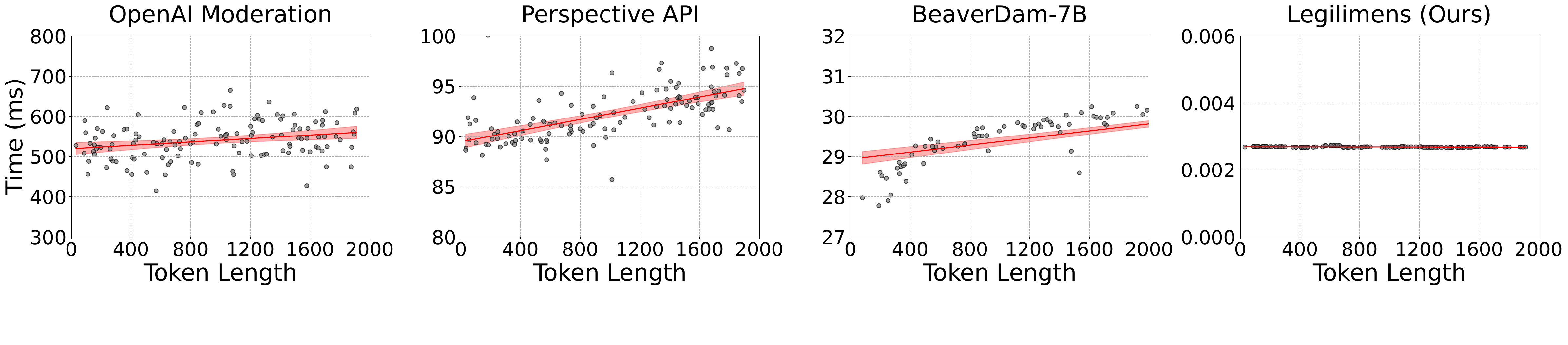}
\caption{The time complexity of \sys compared with three baselines. \sys exhibits a constant complexity of $O(1)$, while the other methods exhibits an approximately linear complexity of $O(n)$.}
\label{fig:time}
\end{figure*}

\clearpage
\section{Overview of dataset}\label{ap:dataset}
\begin{table}[h]\centering

\setlength{\abovecaptionskip}{0pt}%
\setlength{\belowcaptionskip}{0pt}%
\caption{Overview of datasets.}\label{tab:dataset}

\resizebox{0.6\linewidth}{!}{
\begin{threeparttable}

\setlength{\tabcolsep}{1.2mm}{
\begin{tabular}{@{}lcrrr@{}}
\toprule
\textbf{Dataset}      & \textbf{Alias} & \textbf{\#Train} & \textbf{\#Test} & \textbf{Task} \\ \midrule 
HateXplain~\cite{mathew2021hatexplain}            & HAT            & 12,578            & 3,050            & $I$            \\
Measuring Hate Speech~\cite{kennedy2020constructing}                   & MHS            & 16,566            & 4,142            & $I$             \\
OIG-Safety~\cite{Nguyen2023OIG}           & OIG            & 19,769            & 2,000            & $I$             \\
Jigsaw~\cite{JigsawDataset}                & JIG            & 239,204           & 59,801           & $I$             \\
BeaverTails~\cite{Ji2023BeaverTails}           & BEA            & 88,657            & 22,165           & $O$            \\
BeaverTails-\textit{adv}       & BEA-\textit{adv}        & 88,657            & 22,165           & $O$             \\
BEA\&AG           & BAG             & 86,162            & 21,457           & $IO$              \\
BEA-\textit{adv}\&AG       & BAG-\textit{adv}         & 86,162            & 21,457           & $IO$              \\
BEA\&PIQA         & BPI         & 26,180            & 6,546            & $I$               \\
HarmBench~\cite{mazeika2024harmbench}             & HAB             & -                & 320             & $I$            \\
SimpleSafetyTests~\cite{vidgen2024simplesafetytests}     & SST            & -                & 100             & $I$             \\
MaliciousInstructions~\cite{bianchi2023safety} & MAI             & -                & 100             & $I$             \\
JADE~\cite{zhang2023jade}                  & JAD           & -                & 80              & $I$             \\
HExPHI~\cite{qi2023fine}                & HEP            & -                & 330             & $I$             \\
TDCRedTeaming~\cite{mazeika2024harmbench}         & TDC            & -                & 50              & $I$           \\
AdvBench~\cite{chao2023jailbreaking}                & -            & -                & 50             & $IO$             \\
AdvBEA         & -            & -                & 30              & $IO$           \\\bottomrule
\end{tabular}
}



\end{threeparttable}}
\end{table}

\clearpage


\end{document}